%% file: main.tex
\PassOptionsToPackage{table,dvipsnames}{xcolor}
\RequirePackage{fix-cm}
\documentclass{article} %
\usepackage{iclr2027_conference,times}

\usepackage{booktabs}
\input{math_commands.tex}

\usepackage{hyperref}
\usepackage{url}
\usepackage{tabularx}
\usepackage{longtable}
\usepackage{array}
\usepackage{caption}

\usepackage{xcolor}

\usepackage[edges]{forest}
\usepackage{adjustbox}
\usepackage{xcolor}
\usepackage{tikz}
\usetikzlibrary{shadings,positioning}
\usepackage{wrapfig}
\usepackage{booktabs}
\usepackage{enumitem}
\usepackage{multirow}
\usepackage{graphicx}
\usepackage{float}
\input{iclr2026/paper_layout}

\newtcolorbox[auto counter]{promptbox}[2]{
    enhanced,
    breakable,
    colback=gray!4,
    colframe=gray!55,
    boxrule=0.6pt,
    arc=1mm,
    fonttitle=\bfseries\small,
    fontupper=\small\ttfamily,
    title={Prompt~\thetcbcounter: #1},
    label={#2},
}
\input{iclr2026/sections/appendix/7_prompt_styles}

\definecolor{i2iblue}{HTML}{DCEBFA}
\definecolor{i2torange}{HTML}{FCE5D2}
\definecolor{rootfill}{HTML}{D9E4F2}
\definecolor{familyfill}{HTML}{E9EEF5}

\pgfdeclarehorizontalshading{sharedcap}{100bp}{
  color(0bp)=(i2iblue);
  color(49.99bp)=(i2iblue);
  color(50bp)=(i2torange);
  color(100bp)=(i2torange)
}

\forestset{
  root/.style={
    fill=rootfill,
    font=\bfseries
  },
  family/.style={
    fill=familyfill,
    font=\bfseries
  },
  group/.style={
    fill=gray!8,
    font=\itshape
  },
  i2ionly/.style={
    fill=i2iblue
  },
  i2tonly/.style={
    fill=i2torange
  },
  shared/.style={
    shade,
    shading=sharedcap
  },
}
\title{\methodname: When Does Visual\\Generation Improve Visual Understanding\par\vspace{0.55em}}

\author{\hspace*{-\tabcolsep}\begin{minipage}{5.2in}
\raggedright\normalfont\normalsize
Jiaxin Ge\textsuperscript{1*}\enspace
Yiming Qin\textsuperscript{2,6*\textdaggerdbl}\enspace
Ji Xie\textsuperscript{3}\enspace
Haozhe Jiang\textsuperscript{1}\enspace
Xiaochuang Han\textsuperscript{4}\\[0.1em]
Junyi Zhang\textsuperscript{1}\enspace
Andrew Dai\textsuperscript{5}\enspace
Yinfei Yang\textsuperscript{5}\enspace
Jitendra Malik\textsuperscript{1}\enspace
Ranjay Krishna\textsuperscript{4}\enspace
Sewon Min\textsuperscript{1}\\[0.1em]
Haiwen Feng\textsuperscript{1,6\textdagger}\enspace
Le Xue\textsuperscript{5\textdagger}\enspace
Baifeng Shi\textsuperscript{1\textdagger}\enspace
Trevor Darrell\textsuperscript{1\textdagger}\enspace
XuDong Wang\textsuperscript{1,2\textdagger}\\[0.8em]
{\small
\textsuperscript{1}University of California, Berkeley\quad
\textsuperscript{2}Duke University\quad
\textsuperscript{3}Carnegie Mellon University\\
\textsuperscript{4}University of Washington\quad
\textsuperscript{5}Elorian\quad
\textsuperscript{6}Impossible Research}
\end{minipage}}

\iclrfinalcopy %
\begin{document}

\maketitle
\begingroup
\renewcommand{\thefootnote}{\fnsymbol{footnote}}
\footnotetext[1]{Equal contribution. Correspondence: Jiaxin Ge (\href{mailto:gejiaxin01@gmail.com}{\texttt{gejiaxin01@gmail.com}}) and Yiming Qin (\href{mailto:ymk4474@gmail.com}{\texttt{ymk4474@gmail.com}}).}
\footnotetext[2]{Equal contribution.}
\footnotetext[3]{Work done during a summer internship at Impossible Research.}
\endgroup
\lhead{} %
\hypersetup{
  pdftitle={OmniTaskonomy: When Does Visual Generation Improve Visual Understanding},
  pdfauthor={Jiaxin Ge, Yiming Qin, Ji Xie, Haozhe Jiang, Xiaochuang Han, Junyi Zhang, Andrew Dai, Yinfei Yang, Jitendra Malik, Ranjay Krishna, Sewon Min, Haiwen Feng, Le Xue, Baifeng Shi, Trevor Darrell, XuDong Wang},
  pdfkeywords={visual generation, visual understanding, transfer learning, unified multimodal models}
}
\addtocontents{toc}{\protect\setcounter{tocdepth}{-1}} %

\input{iclr2026/sections/0_abstract}

\begingroup
\setlength{\parskip}{5pt plus 2pt minus 1pt}
\setlength{\intextsep}{5pt plus 1pt minus 1pt}
\setlength{\textfloatsep}{5pt plus 1pt minus 1pt}
\setlength{\abovecaptionskip}{2pt}
\input{iclr2026/sections/1_introduction_clean}

\par\endgroup
\input{iclr2026/sections/2_related_work}
\input{iclr2026/sections/3_poc}
\input{iclr2026/sections/4_taxonomy}
\input{iclr2026/sections/5_analysis}
\input{iclr2026/sections/6_gradient}

\input{iclr2026/sections/7_conclusion}

\subsubsection*{Acknowledgments}
We thank Amil Dravid and Yushi Hu for helpful discussions.
\subsubsection*{AI Use Statement}
During manuscript preparation, we used ChatGPT to assist with writing. We also used Codex to assist with data processing and visualization.
The authors designed and conducted the study and take full responsibility for the paper.

\subsubsection*{Reproducibility Statement}
Appendices~\ref{app:controlled}--\ref{app:prompts} provide task definitions, training recipes, taxonomy annotations, complete transfer results, gradient computation details, and prompts. The accompanying numerical exports include evaluation counts, gradient estimates, taxonomy mappings, and figure-generation code.

\bibliography{iclr2026_conference}
\bibliographystyle{iclr2027_conference}

\appendix
\addtocontents{toc}{\protect\setcounter{tocdepth}{2}}  %
\input{iclr2026/sections/appendix/0_overview}   %
\input{iclr2026/sections/appendix/3_controlled} %
\input{iclr2026/sections/appendix/4_taskonomy_construction}
\input{iclr2026/sections/appendix/4_taskonomy}  %
\input{iclr2026/sections/appendix/5_transfer}   %
\input{iclr2026/sections/appendix/6_gradient}   %
\input{iclr2026/sections/appendix/7_prompts}

\end{document}

%% file: math_commands.tex
\usepackage{amsmath,amsfonts,bm}

\def\eqref#1{equation~\ref{#1}}

\def\1{\bm{1}}

\DeclareMathAlphabet{\mathsfit}{\encodingdefault}{\sfdefault}{m}{sl}
\SetMathAlphabet{\mathsfit}{bold}{\encodingdefault}{\sfdefault}{bx}{n}

%% file: iclr2026/paper_layout.tex
\usepackage{xspace}
\usepackage[most]{tcolorbox}
\newcommand{\methodname}{OmniTaskonomy\xspace}

\definecolor{findingaccent}{HTML}{617C96}
\definecolor{findingink}{HTML}{1C2B33}
\definecolor{paperlink}{HTML}{426780}
\hypersetup{colorlinks=true,linkcolor=paperlink,citecolor=paperlink,urlcolor=paperlink}
\newtcolorbox[auto counter]{findingbox}{
    enhanced,
    breakable,
    colframe=findingaccent,
    colback=findingaccent!5!white,
    coltext=findingink,
    boxrule=0.65pt,
    sharp corners,
    boxsep=0pt,
    left=7pt,
    right=7pt,
    top=4pt,
    bottom=4pt,
    before skip=5pt,
    after skip=5pt,
    before upper={\textsc{Finding~\thetcbcounter.}\enspace},
}

%% file: iclr2026/sections/appendix/7_prompt_styles.tex
\definecolor{promptpaper}{HTML}{F7FBFF}
\definecolor{promptheader}{HTML}{EDF5FC}
\definecolor{promptborder}{HTML}{C5DAEA}
\definecolor{promptink}{HTML}{284B63}
\tcbset{
    promptheading/.style={toptitle=4pt,bottomtitle=4pt,
        fonttitle=\bfseries\small\strut},
    promptfigure/.style={
        enhanced,colback=promptpaper,colbacktitle=promptheader,
        colframe=promptborder,coltitle=promptink,
        boxrule=0.5pt,arc=1mm,boxsep=0pt,
        left=8pt,right=8pt,top=7pt,bottom=7pt,
        before skip=0pt,after skip=0pt,
        fontupper=\small,promptheading,
    },
}
\renewtcolorbox[use counter from=promptbox]{promptbox}[2]{
    enhanced,
    breakable,
    colback=promptpaper,
    colbacktitle=promptheader,
    colframe=promptborder,
    coltitle=promptink,
    boxrule=0.5pt,
    arc=1mm,
    boxsep=0pt,
    left=8pt,right=8pt,top=7pt,bottom=7pt,
    before skip=8pt,after skip=10pt,
    promptheading,
    fontupper=\small\ttfamily,
    title={Prompt~\thetcbcounter: #1},
    title after break={Prompt~\thetcbcounter: #1 (continued)},
    label={#2},
}
\newtcolorbox{examplebox}[2][]{
    enhanced,colback=white,colbacktitle=promptheader,
    colframe=promptborder,coltitle=promptink,boxrule=0.5pt,arc=1mm,
    boxsep=0pt,left=7pt,right=7pt,top=6pt,bottom=6pt,
    promptheading,fontupper=\small\raggedright,title={#2},#1,
}
\lstdefinestyle{paperprompt}{
    basicstyle=\fontsize{8}{10}\selectfont\ttfamily,
    columns=fullflexible,breaklines=true,breakatwhitespace=true,
    breakindent=0pt,showstringspaces=false,keepspaces=true,
    aboveskip=2pt,belowskip=2pt,
    literate={á}{{\'{a}}}1 {–}{{-}}1 {—}{{---}}3
      {…}{{...}}3 {→}{{$\rightarrow$}}1 {↔}{{$\leftrightarrow$}}1,
}
\makeatletter
\newcommand{\promptcaption}{\def\@captype{figure}\caption}
\makeatother
\newenvironment{promptfigurepages}[3][]{%
    \def\promptfiguretitle{#2}%
    \def\promptfigurelabel{#3}%
    \begin{tcolorbox}[promptfigure,breakable,#1,title={#2},
        title after break={#2 (continued)}]%
}{%
    \end{tcolorbox}%
    \par\nopagebreak[4]
    \begin{minipage}{\linewidth}
    \promptcaption{\textbf{\promptfiguretitle.}}
    \label{\promptfigurelabel}
    \end{minipage}\par
}

%% file: iclr2026/sections/0_abstract.tex
\begin{abstract}
Training a model to \textit{generate} visual content
can encourage it to learn rich perceptual capabilities related to
geometry, spatial relationships, and objectness;
yet, its benefits for visual \textit{understanding} remain unclear.
We ask: when and how does visual generation supervision improve visual understanding?
We study controlled pairs of image-to-image (I2I) generation and
image-to-text (I2T) understanding tasks that express the same underlying problem in different output modalities. 
We find that under the correct recipe, I2I training improves downstream I2T performance, with \textit{larger gains as the amount of I2I training data increases}.
We next ask which generation tasks benefit which understanding capabilities. To study transfer beyond paired tasks, we introduce \methodname, a unified taxonomy spanning 19 I2I generation tasks and 25 I2T understanding capabilities. The resulting transfer map reveals selective, task-dependent benefits. Some follow intuitive correspondences, \textit{e.g.,} depth prediction improving metric 3D reasoning, object pointing improving counting, and jigsaw reconstruction improving 2D ordering. 
Interestingly, we also uncover surprising connections: 2.5D segmentation improving category recognition and Z-depth prediction improving localization. 
To probe these patterns, we analyze gradient alignment between generation and understanding tasks and find that stronger alignment is associated with larger downstream transfer gains.
Together, our results highlight visual generation as a rich source of supervision for visual understanding and provide a roadmap for unlocking its benefits through the right training curriculum and task selection. Project page: \url{https://omni-taskonomy.github.io/}.
\end{abstract}

%% file: iclr2026/sections/1_introduction_clean.tex
\section{Introduction}

\begin{figure}[htbp]
    \centering
    \includegraphics[width=\linewidth]{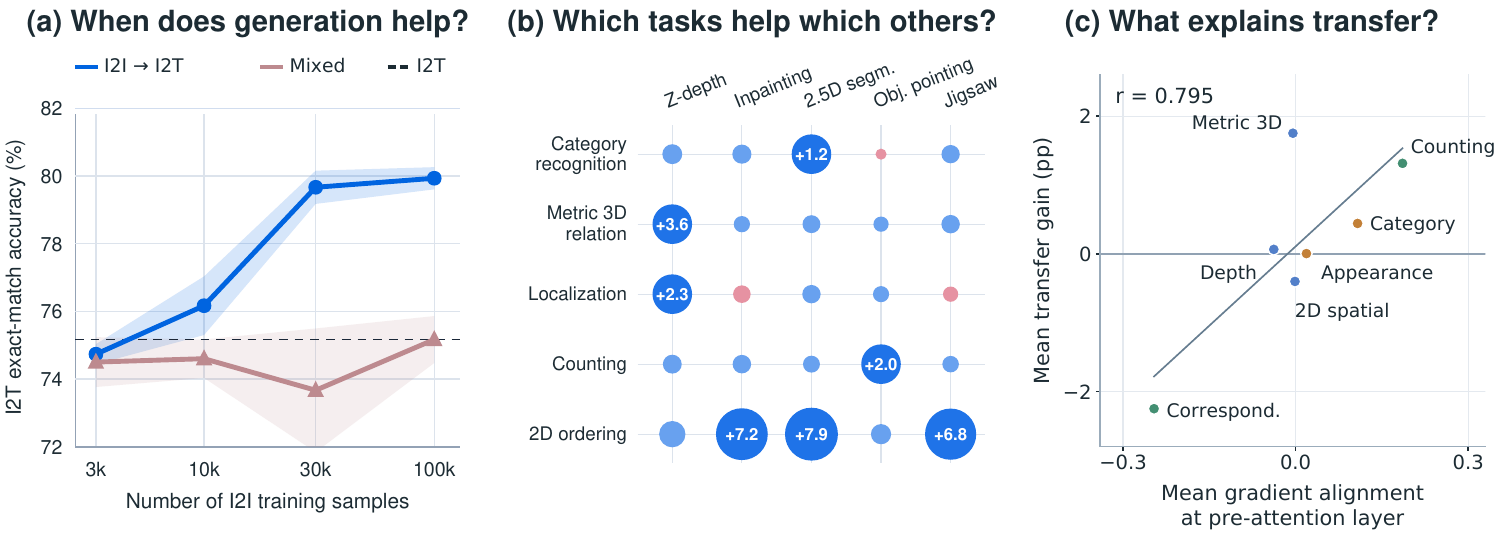}
\caption{
\textbf{When and how does visual generation improve visual understanding?}
(a) I2I training followed by I2T fine-tuning yields increasing gains
as more I2I data is added, whereas mixed training shows no consistent
scaling benefit.
(b) Different generation tasks (I2I) benefit different understanding capabilities.
Blue and red indicate accuracy gains and drops relative to I2T-only
training (percentage points).
(c) Higher average gradient alignment in the understanding branch's
pre-attention RMSNorm parameters is associated with larger mean transfer
gains across capabilities.
}
    \label{fig:teaser}
\end{figure}
When does visual generation help visual understanding?
Prior work has shown that visual understanding can improve visual generation, yet the reverse remains unclear~\citep{tong2024metamorph,kang2026transferability,xie2025reconstruction}.
This has led to a standpoint that generation provides substantially less benefit to understanding~\citep{tong2024metamorph, ICLR2026_2aadf45d, NEURIPS2025_59d4e18a, li2026manzano}.
However, this asymmetry is counterintuitive: visual generation provides dense pixel-level supervision over object appearance, spatial relationships, and geometry, which are also essential for visual understanding tasks such as recognition, counting, spatial reasoning, and 3D perception.
At the same time, recent evidence suggests that generative pretraining can learn representations useful across a broad range of downstream vision tasks~\citep{visionbanana2026}.
These findings leave an important question open: \textit{when and how does visual generation supervision improve visual understanding?}

We investigate this question systematically with unified multimodal models (``omni models''). 
Omni models provide a natural setting for studying this transfer, as visual generation and understanding are learned within the same model~\citep{metaqueries,wu2025harmonizing,bagel}.
Our study addresses three questions:
which training curricula enable generation-to-understanding transfer,
which generation tasks benefit which understanding capabilities,
and whether gradient alignment offers a signal of successful transfer.

We first ask: \textbf{\textit{what training curriculum enables visual generation to improve visual understanding?}}
To study transfer in a controlled setting, we construct paired image-to-image (I2I) generation and image-to-text (I2T) understanding tasks that express the same underlying problem in different output modalities.
For example, given a shuffled image, the I2I task reconstructs the correctly ordered image, while the corresponding I2T task predicts the patch order \textit{of the same image} in text (Figure~\ref{fig:scaling}(a)).
This design holds the input and underlying visual problem fixed
while varying the form of supervision.
We compare six training strategies for combining I2I and I2T data, including mixed I2I--I2T training and two-stage training with I2I training followed by I2T 
fine-tuning. 
We find that the training pipeline beginning with I2I training that updates shared parameters yields gains that grow consistently with additional I2I data.
Directly mixing the two objectives from the outset does not produce the same scaling benefits (Figure~\ref{fig:teaser}(a)).
Moreover, I2I training reduces the amount of I2T supervision needed to reach a given level of understanding performance, with the largest benefits when I2T data is limited.

We next ask: \textbf{\textit{which visual generation tasks help which understanding tasks?}}
Prior work has studied transfer relationships among visual tasks~\citep{zamir2018taskonomy},
but answering this question across generation and understanding requires a common taxonomy that spans both output modalities.
We therefore introduce \methodname, a unified taxonomy that organizes I2I and I2T tasks sharing the same visual capability space.
\methodname is organized around the three foundational problems of computer vision (\textbf{3Rs})~\citep{malik2016three}: 
\textit{\textbf{R}ecognition}, \textit{\textbf{R}econstruction}, and \textit{\textbf{R}eorganization}.
Within each family, we derive fine-grained understanding capabilities bottom-up from existing benchmarks and place both understanding capabilities and I2I tasks within the same taxonomy.
This shared taxonomy allows us to systematically measure transfer from each I2I task to each understanding capability.
Applying this training strategy across \methodname reveals a generation-to-understanding transfer map (Figure~\ref{fig:teaser}(b)).
Some gains follow intuitive capability correspondences, such as Z-depth improving metric 3D reasoning and object pointing improving counting.
Interestingly, we also uncover surprising connections:
2.5D segmentation improves category recognition, and Z-depth prediction
improves localization. These findings show that useful transfer extends beyond closely matched tasks, revealing opportunities to support understanding through a broader range of generation objectives.

This motivates us to ask: \textbf{\textit{What explains the success or failure of transfer between these generation and understanding tasks?}} 
Our hypothesis is that generation and understanding objectives are more likely to support each other when they favor compatible parameter updates.
This hypothesis is motivated by prior work showing that conflicting task gradients can impair multi-task optimization and that gradients can be used to estimate task affinity~\citep{yu2020gradient,fifty2021efficient}.
In controlled task pairs, we find that I2I and I2T gradients align most strongly in the understanding branch's pre-attention normalization parameters, particularly in early layers (Figure~\ref{fig:grad_analysis}(a,b)).
We then examine these parameters across \methodname. Understanding capabilities with higher mean alignment across generation tasks also exhibit larger mean transfer gains (Figure~\ref{fig:teaser}(c)). 
This positive association also holds across individual source-target pairs (Figure~\ref{fig:grad_analysis}(d)), suggesting that gradient alignment offers a promising signal for identifying beneficial generation-understanding task pairings.

Together, our findings highlight visual generation as a rich source of supervision for visual understanding. Realizing its benefits depends on both how the objectives are trained and which visual capabilities they share.
Our study provides a roadmap for exploring this potential: use generation training to establish a useful initialization, select generation tasks for the understanding capabilities they support, and investigate gradient alignment as a signal for identifying promising task pairings.
\methodname makes these relationships explicit, opening up opportunities to design omni models in which learning to generate strengthens learning to understand.

%% file: iclr2026/sections/2_related_work.tex
\begingroup
\setlength{\parskip}{2pt}
\makeatletter
\def\section{\@startsection{section}{1}{\z@}{-1.5ex plus -.3ex minus -.2ex}{1ex plus .2ex minus .2ex}{\large\sc\raggedright}}
\def\paragraph{\@startsection{paragraph}{4}{\z@}{3pt plus 1pt minus .5pt}{-1em}{\normalsize\bf}}
\makeatother
\section{Related work}
\paragraph{Omni models.}
Omni models unify multimodal understanding and generation within a single architecture. Early works explored fully autoregressive modeling over discrete text and image tokens~\citep{team2024chameleon,janus}, as well as hybrid architectures that combine autoregressive language modeling with diffusion for image generation~\citep{zhou2025transfusion,xie2025show,blip3o,blip3onext}. More recently, Mixture-of-Transformers~\citep{liang2024mixture} has become a well-adopted architecture, using specialized Transformer parameters for different modalities while retaining shared attention for cross-modal interaction~\citep{diao2026sensenova,agarwal2026cosmos,bagel}. We study visual generation-to-understanding transfer under this architecture and adopt BAGEL~\citep{bagel} as our baseline.
\paragraph{Generation and understanding.}
Previous work studies how generation and understanding help each other during inference and training. At inference time, models generate sketches, transformed images, or interleaved visual thoughts for visual reasoning~\citep{hu2024visual,gu2025thinkmorph,qin2025chain,li2025zebra}. During training, previous studies mix understanding and generation data during pretraining~\citep{tong2024metamorph,liquid,chen2025unihetero,zhang2025unified,tong2026beyond,han2026towards} or introduce auxiliary generation tasks during post-training~\citep{yu2026semantic,su2026generation,zheng2026uni}. A separate line of work runs in the opposite direction, using visual understanding signals to supervise visual generation~\citep{xie2025reconstruction,jin2025srum,niu2025does}. We study training-time transfer by measuring how different I2I tasks affect individual visual understanding tasks.

\paragraph{Visual task taxonomies and evaluation.}
Visual generation and understanding benchmarks use different task categories.
For visual generation, existing benchmarks mainly evaluate the instruction-following capability and quality of the generated images by testing basic generation and editing capabilities~\citep{ghosh2023geneval,huang2023t2i,wisebench,ye2026imgedit,ge2026constantly}.
Understanding benchmarks, instead, organize evaluation according to benchmark-specific task types or capability categories ~\citep{fu2024blink,ying2024mmt,fu2025mme}.
More recent omni-model benchmarks place understanding and generation within a common evaluation framework or construct interleaved tasks in which the two modalities interact~\citep{xie2025mme,li2025unieval,shi2025realunify,zou2026uni,wang2026quantifying,wang2026beyond,wen2026unig2u,liu2026unison}.
However, simply sharing an evaluation suite or requiring both output modalities does not by itself establish which I2I and I2T tasks depend on the same underlying visual capability.
In contrast, \methodname organizes I2I tasks and understanding capabilities within a single capability-based taxonomy, enabling cross-modal transfer to be analyzed at the level of individual visual capabilities.

\par\endgroup

%% file: iclr2026/sections/3_poc.tex
\begingroup
\setlength{\parskip}{3pt}
\setlength{\textfloatsep}{7pt plus 1pt minus 1pt}
\setlength{\abovecaptionskip}{3pt}
\makeatletter
\def\section{\@startsection{section}{1}{\z@}{-1.5ex plus -.3ex minus -.2ex}{1ex plus .2ex minus .2ex}{\large\sc\raggedright}}
\def\paragraph{\@startsection{paragraph}{4}{\z@}{5pt plus 1pt minus .5pt}{-1em}{\normalsize\bf}}
\makeatother
\section{Does visual generation help visual understanding?}
\label{sec:controlled_settings}
We first ask whether visual generation can improve visual understanding.
To isolate the effect of confounding factors, we construct paired I2I and I2T tasks that require solving the same task from the same input, differing only in whether the answer is produced as an image or as text.
This allows us to directly test whether supervision from an I2I objective transfers to the corresponding I2T task. This design holds the input and underlying visual problem fixed while varying the output modality.
We study how the training recipe affects transfer, how I2I supervision reduces the need for I2T data, and whether the two objectives correspond at the instance level.
\begin{figure}[t]
    \centering
    \includegraphics[width=\linewidth]{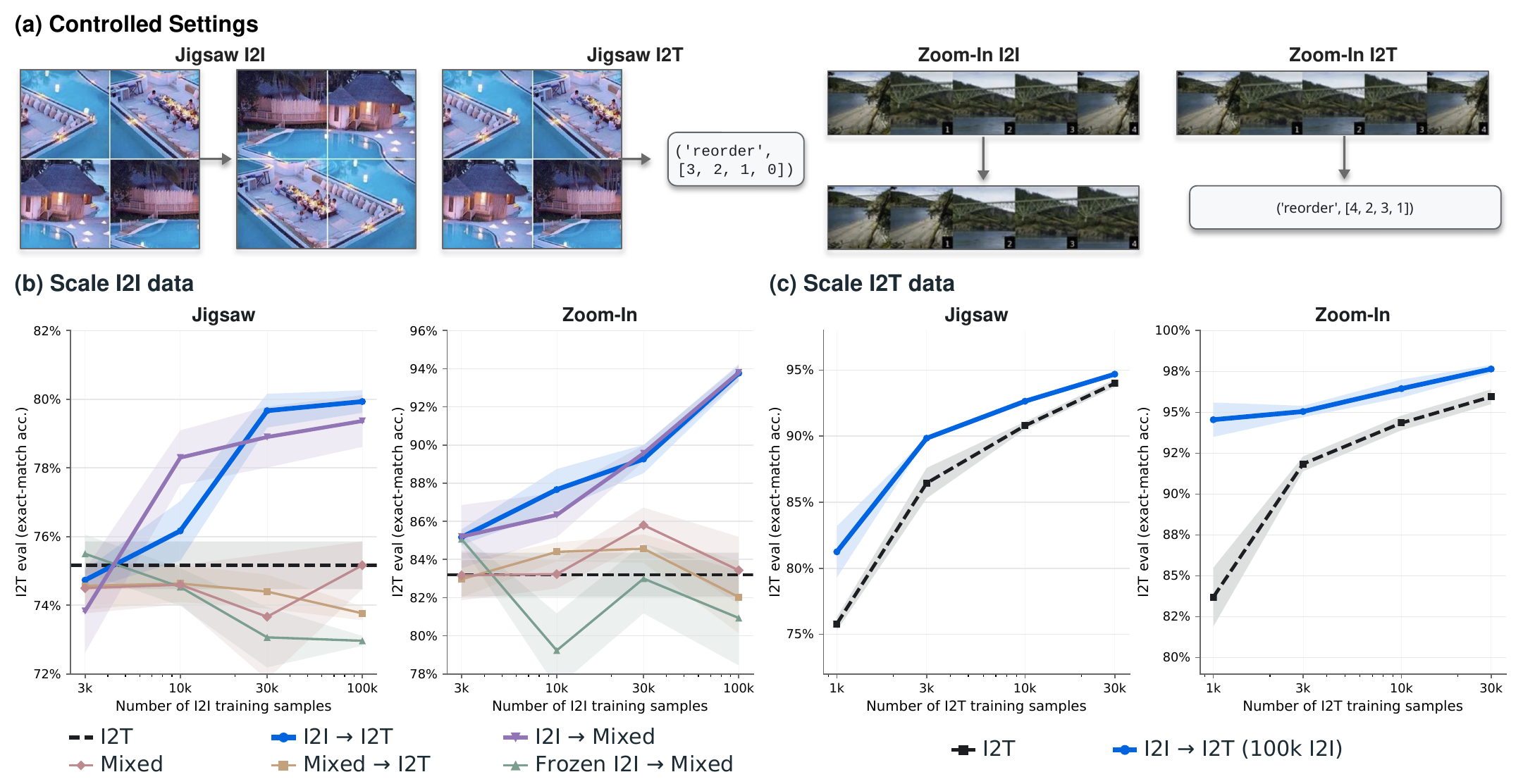}
    \caption{\textbf{Scaling I2I and I2T supervision.}
    (a) Controlled Jigsaw and Zoom-In tasks with image and text outputs.
    (b) I2T accuracy versus I2I training examples across recipes, with 1k I2T examples.
    (c) I2T accuracy versus I2T training examples for I2T-only and I2I $\rightarrow$ I2T training (100k I2I examples).
    Curves show means over three seeds. Shading indicates $\pm1$ SEM.}
    \label{fig:scaling}
\end{figure}
\paragraph{Controlled paired tasks.}
We adapt two tasks from VisGym~\citep{wang2026visgym}: Jigsaw and Zoom-In. 
In Jigsaw, patches of an image are shuffled, and the model must recover their spatial arrangement.
In Zoom-In, views of the same image at different zoom levels are shuffled, and the model must recover their correct order. 
For each input, the I2I objective produces the correctly ordered image, while the I2T objective predicts the same ordering as a permutation in text. 
The objectives require the same visual operation on the same input and differ only in output modality.
We instantiate this on BAGEL, a unified multimodal model based on a Mixture-of-Transformers (MoT) architecture, and then evaluate transfer exclusively on the paired I2T task.
\paragraph{Which training recipe transfers best?}
Under this setup, we study how I2I and I2T data should be combined during training. 
We compare six training recipes: \textbf{I2T-only}, \textbf{I2I $\rightarrow$ I2T}, \textbf{Mixed $\rightarrow$ I2T}, \textbf{Frozen I2I $\rightarrow$ Mixed}, \textbf{Mixed}, and \textbf{I2I $\rightarrow$ Mixed}.
We use $\rightarrow$ to denote sequential training stages.
\textbf{Mixed} denotes joint training with both I2I and I2T data in each batch.
\textbf{I2I} denotes training on visual generation data while updating parameters shared with the I2T objective.
\textbf{Frozen I2I} denotes visual generation training in which these shared parameters are frozen, and only visual generation-specific parameters are updated.
Architecture and training details are provided in Appendix~\ref{app:controlled-method}.
We fix the I2T budget at 1k examples with identical I2T training epochs across recipes and vary the amount of I2I data.
As shown in Figure~\ref{fig:scaling}(b), I2I $\rightarrow$ I2T and I2I $\rightarrow$ Mixed scale consistently as the I2I data increases.
Both recipes begin with an I2I-only training stage that updates parameters shared with I2T.
In contrast, freezing these weights during the I2I training stage or directly mixing the two objectives from the beginning leads to weaker or less stable gains.
We therefore use I2I $\rightarrow$ I2T as the default recipe in subsequent experiments.
\begin{findingbox}
An initial I2I training stage that updates parameters shared with the I2T objective provides a useful initialization for subsequent I2T learning.
\end{findingbox}
\paragraph{How much does I2I supervision reduce the need for I2T data?}
We next quantify how much I2I training can reduce the amount of I2T data needed to reach a given level of understanding performance.
We fix the I2I budget at 100k examples and compare I2I $\rightarrow$ I2T with I2T-only training at I2T budgets of 1k, 3k, 10k, and 30k examples.
As shown in Figure~\ref{fig:scaling}(c), I2I training improves performance across all four budgets, with larger gains at smaller I2T budgets.
On Zoom-In, 100k I2I examples followed by only 1k I2T examples achieve performance comparable to training on 10k I2T examples alone. 
On Jigsaw, 100k I2I examples followed by only 3k I2T examples achieve performance comparable to training on 10k I2T examples alone.
Thus, evidence shows that I2I supervision can partially substitute for direct I2T supervision and may be particularly useful when understanding data is limited.
Training details are in Appendix~\ref{app:controlled-recipes}.
\par\addvspace{3pt}
\begingroup
\setlength{\intextsep}{2pt}
\setlength{\columnsep}{5pt}
\begin{wraptable}[7]{r}{0.38\linewidth}
    \centering
    \small
    \setlength{\tabcolsep}{3pt}
    \setlength{\abovecaptionskip}{0pt}
    \makeatletter
    \def\@captype{table}
    \long\def\@makecaption#1#2{\vskip\abovecaptionskip{\small\raggedright #1: #2\par}\vskip\belowcaptionskip}
    \makeatother
    \caption{\textbf{I2I--I2T instance alignment.}}   \label{tab:controlled_instance_alignment}
    \begin{tabular}{@{}lcc@{}}
        \toprule
        & Jigsaw & Zoom-In \\
        \midrule
        I2I Acc. & 70.9 & 91.8 \\
        I2T Acc. & 81.3 & 90.6 \\
        $P(\mathrm{I2T}\checkmark\mid\mathrm{I2I}\checkmark)$ & 88.2 & 90.9 \\
        $P(\mathrm{I2T}\checkmark\mid\mathrm{I2I}\times)$ & 64.3 & 86.7 \\
        \bottomrule
    \end{tabular}
\end{wraptable}
\rightskip=0pt plus 1em
\noindent
\mbox{\textbf{Instance-level correspondence.}}\quad
Beyond aggregated transfer, we ask whether success on paired I2I and I2T tasks also corresponds at the level of individual examples.
We evaluate both I2I and I2T outputs from the same final I2I $\rightarrow$ I2T checkpoint for each of three training seeds, trained on 30k I2I and 1k I2T examples, using our 1k evaluation paired inputs.
We then compare I2T accuracy conditioned on whether the I2I prediction for the same input is correct.
As shown in Table~\ref{tab:controlled_instance_alignment}, I2T performance is higher on examples for which the corresponding I2I prediction is correct and lower on those that were wrong.
This indicates that the correspondence between paired I2I and I2T tasks extends to individual examples. More evaluation details are in Appendix~\ref{app:controlled-eval}.
\par
\endgroup
\par\endgroup

%% file: iclr2026/sections/4_taxonomy.tex
\section{\methodname: A unified taxonomy of visual capabilities}
Now we want to answer the question of which tasks can transfer, and to do that we need an organized benchmark that organize I2I and I2T tasks sharing the same visual capabilities.
Existing benchmarks typically organize visual tasks by task formulation or output modality, leaving visual generation and understanding tasks separated.
This makes it difficult to compare and study tasks that rely on similar visual capabilities but produce different outputs.
We introduce \methodname{}, a unified taxonomy of I2I tasks and understanding capabilities.
For a visual understanding sample, we consider the primary visual capability required to solve it; for an I2I task, we consider the visual capability directly supervised by its training objective.
At the top level, we adopt the three Rs of computer vision~\citep{malik2016three}: \textit{Recognition}, \textit{Reconstruction}, and \textit{Reorganization}.
I2I training objectives and understanding capabilities remain separate leaves but are placed in the same hierarchy according to these shared visual capabilities.

\begin{figure}[t]
\centering
\includegraphics[width=\linewidth]{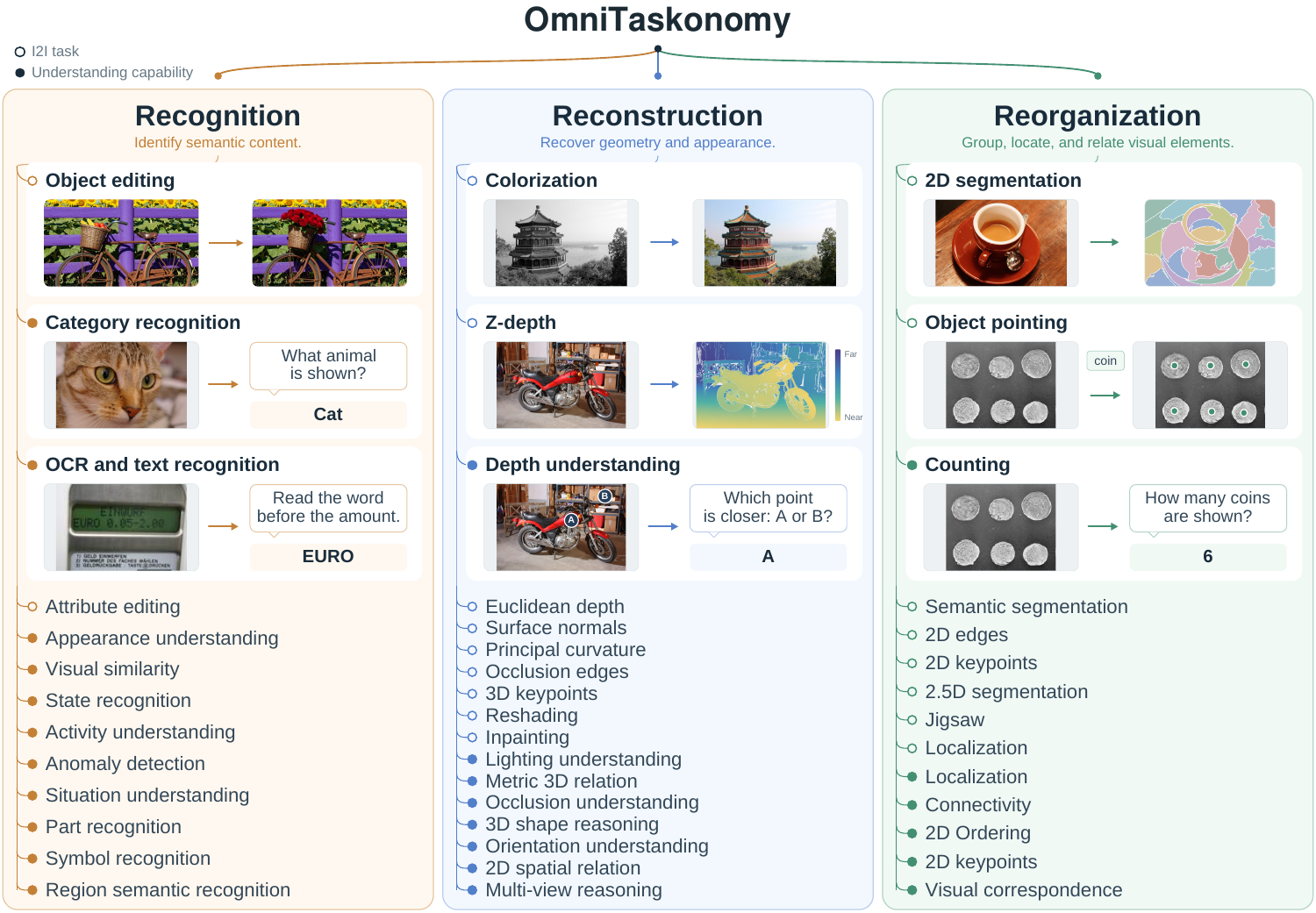}
\caption{\textbf{\methodname.}
A unified taxonomy of I2I tasks and understanding capabilities under the three Rs: \textit{Recognition}, \textit{Reconstruction}, and \textit{Reorganization}.
I2I tasks and understanding capabilities occupy separate leaves in a shared hierarchy.}
\label{fig:taskonomy}
\end{figure}

\paragraph{Organizing visual capabilities with the 3Rs.}
\textbf{Recognition} covers 
semantic content, including object and part identity, appearance, state, activity, text, and symbols.
\textbf{Reconstruction} covers geometric and photometric properties of the scene, including spatial relations, depth, metric distance, orientation, 3D shape, lighting, and occlusion.
\textbf{Reorganization} covers how visual elements are grouped, localized, and related, including segmentation, correspondence, ordering, counting, keypoints, and connectivity.
The 3R families define the top level of \methodname{}; fine-grained capabilities are derived from existing visual tasks rather than specified by the three families.
Appendix~\ref{app:unitaskonomy-definitions} lists all understanding capabilities and their definitions.

\paragraph{Deriving fine-grained visual capabilities.}
We derive the fine-grained visual understanding capabilities bottom-up from seven vision-language benchmarks: BLINK~\citep{fu2024blink}, MMStar~\citep{chen2024wemmstar}, MMT-Bench~\citep{ying2024mmt}, CV-Bench~\citep{tong2024cambrian}, RealWorldQA~\citep{xai2024realworldqa}, MMVP~\citep{tong2024eyes}, and VStarBench~\citep{wu2024vstarbench}.
We first sample benchmark questions and use \texttt{gemini-3-flash-preview}~\citep{gemini3flash} to describe the visual attributes and relations involved in solving each question, such as \emph{relative depth}, \emph{size comparison}, and \emph{occlusion}.
Then we review these descriptions and consolidate them into a fixed attribute schema, a list of key--value pairs describing image attributes, which is then used to annotate the full benchmark collection so that tasks from different benchmarks are described consistently.
We next divide the annotated samples into ten benchmark-stratified batches.
For each batch, a VLM organizes the samples into a local task tree using their annotated attributes and assigns each sample to the leaf that best describes its primary visual requirement.
We then merge the local trees, combining synonymous leaves across batches to form an initial set of fine-grained capabilities.
We place the resulting capabilities under \textit{Recognition}, \textit{Reconstruction}, or \textit{Reorganization}.
We manually review the resulting hierarchy, splitting leaves that mix distinct capabilities, merging redundant leaves, and refining definitions until the leaf categories have clear boundaries.
Figure~\ref{fig:taskonomy} shows the resulting taxonomy.
Appendix~\ref{app:construction-schema} describes the attribute schema, and Appendix~\ref{app:construction-trees} provides further details on tree construction and refinement.

\paragraph{Integrating I2I tasks.}
We next place I2I training objectives in the same 3R hierarchy.
We collect a broad set of dense prediction, reconstruction, and image-editing tasks from existing vision and multimodal learning settings, including tasks studied in Taskonomy~\citep{zamir2018taskonomy} as well as objectives such as inpainting and object editing.
Each I2I task is placed according to the main visual capability directly supervised by its objective.
For example, Z-depth and Euclidean depth prediction are placed under \textit{Reconstruction}, alongside the understanding capability \textit{depth understanding}.
2D keypoint prediction is placed under \textit{Reorganization}.
Object editing and attribute editing are placed under \textit{Recognition}, as they directly supervise object identity and visual attributes such as color or material.
The final taxonomy contains 19 I2I tasks and 25 understanding capabilities.
Appendix~\ref{app:unitaskonomy-i2i} lists all I2I tasks and their assignments, and Appendix~\ref{app:i2i-training-data} describes their training data.

\paragraph{Annotating benchmark samples.}
Once the taxonomy is fixed, we reassign each eligible benchmark sample to one of the finalized understanding leaves.
Three VLM judges independently annotate each sample using the image, question, answer choices, reference answer, and the same capability definitions.
We retain an example when at least two of the three judges assign it to the same capability and exclude examples without a majority, yielding 9,444 samples.
Appendix~\ref{app:unitaskonomy-annotation} reports unanimous agreement, majority agreement, and exclusion statistics.
Appendix~\ref{app:prompt-annotation} provides the annotation prompts and model inputs.
We further validate the resulting assignments through human reviews.
Four human annotators review a stratified subset of 250 samples covering all 25 capabilities.
Each annotator reviews 50 examples shared across all four annotators and 50 additional non-overlapping examples, yielding 400 human reviews.
For each example, the annotator can accept the proposed capability, reassign it to another capability, or mark the assignment as unsure.
Among definite judgments, 97.4\% of human labels match the VLM-assigned capability.
Human-VLM agreement remains high after correcting for chance, with Cohen's $\kappa$ ranging from 0.956 to 0.989 across annotators (mean $0.972$).
On the 50 examples reviewed by all four annotators, the resulting human capability labels also show high inter-annotator agreement (Krippendorff's $\alpha=0.943$).
These results provide human validation of the automated sample assignments.
Appendix~\ref{app:human-validation} gives the full review protocol and agreement analysis.

%% file: iclr2026/sections/5_analysis.tex
\begin{figure}[t]
\centering
\includegraphics[width=\linewidth]{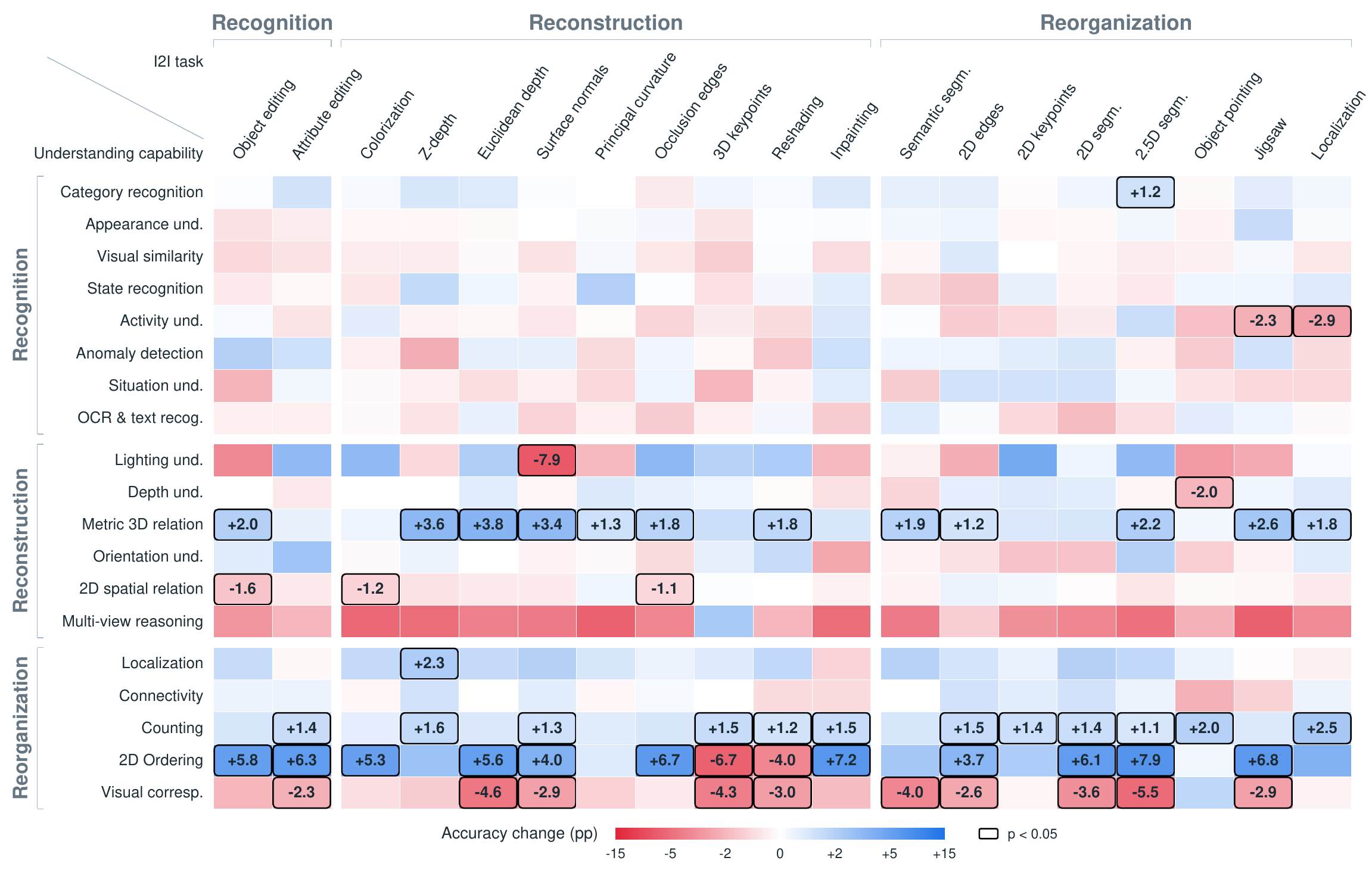}
\caption{\textbf{Generation-to-understanding transfer across visual capabilities.}
Each column corresponds to an I2I source task and each row to an understanding capability in \methodname.
Cells report the change in I2T accuracy, in percentage points, relative to the I2T-only baseline.
Results are averaged over three seeds.
Blue indicates positive transfer and red indicates negative transfer.
Cells with $p<0.05$ under an exact paired permutation test against the I2T-only baseline are outlined and labeled with their gains. The detailed numbers in the heat map are in Appendix~\ref{app:transfer-full}.}
\label{fig:transfer}
\end{figure}

\section{Which generation tasks help which capabilities?}
\label{sec:transfer}
The controlled study shows that I2I training can improve the corresponding I2T task.
We next ask whether this transfer extends beyond closely paired tasks, and which generation objectives benefit which understanding capabilities.
To study this, we evaluate all 19 I2I tasks in \methodname{} against their 25 understanding capabilities.

\subsection{Experimental setup}
We follow the I2I $\rightarrow$ I2T recipe: starting from the same BAGEL checkpoint, we use approximately 50k I2I examples followed by 50k LLaVA-Instruct examples~\citep{liu2024llava}
. The sources span Recognition (2), Reconstruction (9), and Reorganization (8); Appendix~\ref{app:transfer-setup} specifies the checkpoint set. The baseline is trained only on the I2T data. We evaluate all checkpoints on the same 9,444 evaluation examples and define transfer from source $s$ to capability $t$ as
$\Delta_{s,t}=\operatorname{Acc}(M_s,\mathcal D_t)-\operatorname{Acc}(M_{\mathrm{I2T}},\mathcal D_t)$,
reported in percentage points. We assess significance using exact paired permutation tests on per-example correctness against the I2T-only baseline ($p<0.05$; Appendix~\ref{app:transfer-tests}). Figure~\ref{fig:transfer} shows the 19 capabilities with more than 100 evaluation examples and marks cells with $p<0.05$.
Appendix~\ref{app:transfer} provides all 25 capabilities and baseline accuracies. 

\subsection{Transfer is selective across source and target tasks}
\paragraph{Transfer is concentrated in a subset of understanding capabilities.}
The transfer map in Figure~\ref{fig:transfer} is highly non-uniform.
Five of the 19 displayed capabilities improve significantly with at least one I2I source; several others show little or no positive transfer under the same recipe.
Metric 3D relation and counting benefit significantly from the broadest set of sources: 12 I2I tasks.
2D ordering follows with eleven sources.
In contrast, capabilities such as OCR, text recognition, or appearance understanding show no such improvement from any tested source.
\paragraph{Shared visual operations predict several of the strongest gains.}
Localization and object pointing produce the largest improvements in counting ($+2.5$ and $+2.0$ pp), consistent with all three tasks requiring individual objects to be identified and spatially localized.
Similarly, Z-depth, Euclidean depth, and surface normals improve metric 3D relation by $3.6$, $3.8$, and $3.4$ pp, respectively, connecting dense geometric reconstruction to relational 3D judgments.
Jigsaw improves 2D ordering by $6.8$ pp, consistent with both tasks requiring the relative spatial arrangement of image regions.
These cases suggest that transfer often follows shared visual capabilities, even when the source and target use different output modalities.
\paragraph{Useful transfer is not confined to closely matched capabilities.}
Several significant gains are less direct.
Inpainting improves both counting ($+1.5$ pp) and 2D ordering ($+7.2$ pp), despite neither target explicitly requiring missing-region reconstruction.
Solving inpainting tasks may encourage the model to infer object quantity and global spatial structure from incomplete local evidence.
2.5D segmentation likewise improves category recognition ($+1.2$ pp), potentially because capturing object boundaries can shape cues that support category recognition.
These examples show that an I2I objective can also benefit targets across tasks.

\begin{findingbox}
Visual generation supervision yields significant gains for specific understanding capabilities, both within and across task families.
For example, localization and object pointing yield the largest counting gains, while depth and surface-normal prediction improve metric 3D relation.
\end{findingbox}

%% file: iclr2026/sections/6_gradient.tex
\section{What explains visual generation-to-understanding transfer?}
\label{sec:gradient}
Our transfer map shows that gains from visual generation supervision depend on the source task and target capability.
We next ask whether these differences are associated
with the compatibility of their optimization signals.
We hypothesize that transfer is stronger when the two objectives favor similar parameter updates.
We test this by measuring where I2I and I2T gradients align in controlled paired tasks, then examining whether alignment tracks transfer across tasks in \methodname.
\begin{figure}[t]
\centering
\includegraphics[width=\linewidth]{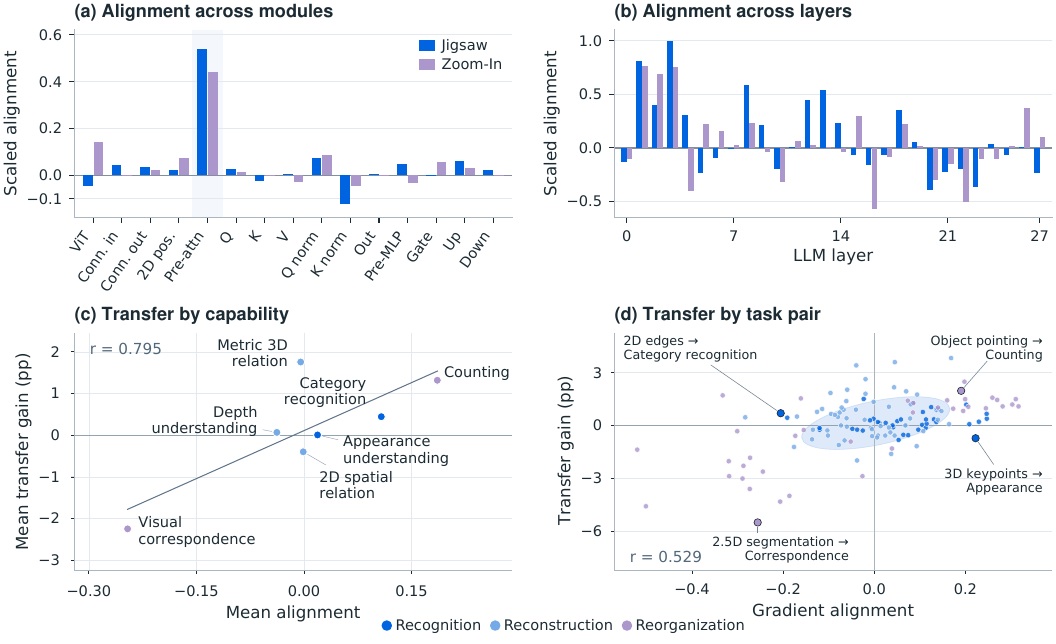}
\caption{\textbf{Gradient alignment and downstream transfer.}
(a,b) Gradient alignment for the controlled Jigsaw and Zoom-In pairs across model components and individual pre-attention RMSNorm layers.
(c) Mean gradient alignment versus mean transfer for seven understanding capabilities, averaging over 19 I2I sources.
(d) Alignment and transfer for all 133 source--target pairs.
}
\label{fig:grad_analysis}
\end{figure}

\subsection{Measuring gradient alignment}
We measure whether I2I and I2T objectives induce aligned updates to the same parameters.
For a paired example \(i\), objective \(o\in\{\mathrm{I2I},\mathrm{I2T}\}\), and parameter block \(b\) (e.g., an input RMSNorm weight or an attention query-projection weight), we compute the gradients at the pretrained checkpoint \(\boldsymbol{\theta}_0\):
\[
\mathbf{g}^{o}_{b,i}
=\left.\nabla_{\boldsymbol{\theta}_b}\mathcal L_o(x^{o}_i;\boldsymbol{\theta})\right|_{\boldsymbol{\theta}=\boldsymbol{\theta}_0},
\qquad
\mathbf{u}^{o}_{b,i}=\mathbf{g}^{o}_{b,i}/\|\mathbf{g}^{o}_{b,i}\|_2.
\]
Here \(\boldsymbol{\theta}_b\) denotes the parameters in block \(b\), \(x_i^o\) denotes the example for the objective \(o\), and \(\mathbf{u}^{o}_{b,i}\) denotes its normalized gradient direction.
Gradients are high-dimensional, and their dimensions vary across parameter blocks. We therefore fit a shared uncentered PCA basis to each block's normalized I2I and I2T gradients, retaining the fewest components whose eigenvalues sum to at least \(99\%\) of the total.
We compute cosine similarity in this shared subspace and scale it by the square root of its estimated effective dimension to compare blocks with different effective dimensionalities. The alignment score is:
\[
s_{b,i}
=
\sqrt{d_{\mathrm{eff},b}}\;
\cos\!\left(
\mathbf{P}_b^\top \mathbf{u}^{\mathrm{I2I}}_{b,i},
\mathbf{P}_b^\top \mathbf{u}^{\mathrm{I2T}}_{b,i}
\right),
\]
where \(\mathbf{P}_b\) contains the retained PCA directions and
\(d_{\mathrm{eff},b}\) denotes the estimated effective dimensionality of the projected gradient distribution.
Larger positive scores indicate stronger directional alignment, while negative scores indicate opposing directions.
Full details are provided in Appendix~\ref{app:gradient-method}.

\begingroup
\setlength{\parskip}{4pt}
\setlength{\abovedisplayskip}{4pt plus 1pt minus 1pt}
\setlength{\belowdisplayskip}{4pt plus 1pt minus 1pt}
\subsection{Gradient structure in the controlled setting}
\label{sec:gradient-controlled}
We first study the controlled Jigsaw and Zoom-In pairs from Section~\ref{sec:controlled_settings}.
For each task, we sample 500 matched examples and compute I2I and I2T gradients at the pretrained checkpoint.
Because the two objectives solve the same underlying visual problem from the same input, this setting lets us localize where their optimization signals agree.
Figure~\ref{fig:grad_analysis}(a,b) shows gradient alignment across model components.
For both tasks, alignment is strongest in the understanding branch's pre-attention RMSNorm parameters.
Examining these parameters layer by layer further shows that the strongest alignment occurs in the earlier transformer layers.
\begin{findingbox}
In controlled I2I--I2T pairs, gradient alignment is strongest in early pre-attention normalization layers of the visual understanding branch.
\end{findingbox}
\subsection{Does gradient alignment track transfer across tasks?}
Motivated by the strong alignment observed in Section~\ref{sec:gradient-controlled}, we focus on the understanding branch's pre-attention RMSNorm~\citep{zhang2019rootrmsnorm} parameters, concatenated across all 28 transformer layers.
We analyze all 19 I2I tasks and the seven understanding capabilities with more than 500 evaluation examples (Appendix~\ref{app:gradient-minibatch}).
For each capability, we sample 500 examples and compute gradients using minibatches of size 64.
For an I2I source task $s$ and target visual understanding capability $t$, we compute the mean post-PCA gradient cosine ($G_{s,t}$) at the pretrained checkpoint using these parameters.
We compare this quantity with the downstream transfer gain
($\Delta_{s,t}$), defined as the target accuracy improvement over the I2T-only baseline after I2I training followed by I2T finetuning, averaged over three seeds.
\paragraph{Which visual understanding capabilities benefit more from visual generation?}
To compare the overall benefit to each visual understanding capability \(t\), we average alignment and transfer over all 19 sources:
\[
\bar{G}_t=\frac{1}{19}\sum_{s=1}^{19}G_{s,t},
\qquad
\bar{\Delta}_t=\frac{1}{19}\sum_{s=1}^{19}\Delta_{s,t}.
\]
As shown in Figure~\ref{fig:grad_analysis}(c), average alignment and average transfer are strongly positively correlated across the seven capabilities ($r=0.795$).
Capabilities whose gradients are more compatible with generation objectives, on average, therefore, also tend to receive larger gains from I2I training.
\paragraph{Does alignment track transfer across task pairs?}
We next examine all $19\times7=133$ source–target pairs individually.
Figure~\ref{fig:grad_analysis}(d) plots each pair according to its gradient alignment ($G_{s,t}$) and downstream transfer gain ($\Delta_{s,t}$).
Alignment and transfer are positively correlated across these 133 pairs ($r=0.529$).
Together, these results identify gradient alignment as an optimization-level signal associated with generation-to-understanding transfer.
\begin{findingbox}
Gradient alignment is concentrated in early pre-attention normalization layers and is positively associated with downstream transfer across both understanding capabilities and individual source-target pairs.
\end{findingbox}
\par\endgroup

%% file: iclr2026/sections/7_conclusion.tex
\vspace{-4pt}
\section{Conclusion}
\vspace{-3pt}
We show that visual generation can improve visual understanding with an appropriate training recipe.
In controlled task pairs, I2I training followed by I2T finetuning yields gains that grow with I2I data, especially when I2T data is limited.
With \methodname, we show that transfer is capability-specific, with benefits both within and across task families.
Gradient alignment is strongest in early pre-attention normalization layers in the controlled setting and is positively associated with transfer across the studied capabilities and task pairs.
Together, these results suggest that generation objectives should be chosen for the visual capabilities they support, alongside a training recipe that enables this supervision to benefit the target understanding task.

%% file: iclr2026/sections/appendix/0_overview.tex
\clearpage
\renewcommand{\contentsname}{Appendix Contents}
{\hypersetup{linkcolor=paperlink}\color{paperlink}\tableofcontents}

%% file: iclr2026/sections/appendix/3_controlled.tex
\section{Controlled generation-to-understanding study}
\label{app:controlled}

This appendix gives the training details, model architecture, and instance-level evaluation for the controlled study in Section~\ref{sec:controlled_settings}. Both tasks are adapted from VisGym. For each Jigsaw or Zoom-In puzzle, the I2I and I2T versions share the same input and differ only in whether the answer is an image or text.

\subsection{Training setting}
\label{app:controlled-model}
\label{app:controlled-recipes}

We finetune BAGEL-7B-MoT for the Jigsaw and Zoom-In experiments in Figure~\ref{fig:scaling}. Each reported result is the mean evaluation score of three checkpoints trained with different random seeds. Shading in Figure~\ref{fig:scaling} indicates one standard error of the mean (SEM).

\paragraph{Optimization.}
Table~\ref{tab:controlled-optimizer} lists the optimization settings shared by all recipes. In multi-stage recipes, each stage loads the model weights of the previous stage and starts with a fresh optimizer and learning-rate schedule.

\begin{table}[h]
\centering\small
\caption{\textbf{Optimization settings.}}
\label{tab:controlled-optimizer}
\begin{tabularx}{0.8\linewidth}{@{}c >{\centering\arraybackslash}X@{}}
\toprule
Setting & Value \\
\midrule
Optimizer & AdamW with $\beta_1=0.9$, $\beta_2=0.95$, $\epsilon=10^{-15}$ \\
Peak learning rate & $2\times10^{-5}$ \\
Weight decay & 0 \\
Schedule & Linear warm-up, then constant learning rate \\
Warm-up & 50 updates for initial I2I training; 8 for I2T or Mixed training \\
Gradient clipping & Global gradient $\ell_2$ norm capped at 1.0 \\
Parallelism & 4 GPUs with FSDP \\
Numerical precision & BF16 mixed precision \\
\bottomrule
\end{tabularx}
\end{table}

\paragraph{Conditioning dropout.}
For I2I examples, conditioning dropout removes the instruction text, the source-image ViT tokens, or the source-image VAE latents. Each of the three conditions is dropped independently with the same probability, and the target image is always kept. I2T examples use no conditioning dropout, including those in mixed batches, and the dropout probability for each recipe is set to 0.1.

\begin{table}[h]
\centering
\small
\setlength{\tabcolsep}{4pt}
\caption{\textbf{I2I pool sizes and mixed batches.}}
\label{tab:controlled-batches}
\begin{tabular}{@{}ccc@{}}
\toprule
Plot label & I2I examples & Mixed batch (I2I:I2T) \\
\midrule
3k   & 1,876   & 4:64 \\
10k  & 9,380   & 20:64 \\
30k  & 30,016  & 64:64 \\
100k & 100,000 & 212:64 \\
\bottomrule
\end{tabular}
\end{table}

\paragraph{Batch composition.}
I2I-only and I2T-only stages use 16 examples per GPU, or 64 globally. In Mixed, I2I $\rightarrow$ Mixed, and Frozen I2I $\rightarrow$ Mixed, each mixed batch contains 16 I2T examples per GPU, as in I2T-only training, plus a fixed number of I2I examples. All settings use the same I2T batch size, and use 1k I2T data for 15 epochs, including I2T training in mixed stages. Table~\ref{tab:controlled-batches} lists the I2I pool size and the I2I:I2T ratio of the mixed batch for each plot label.

\clearpage
\subsection{Model architecture and trainable modules}
\label{app:controlled-method}
\begin{wrapfigure}{r}{0.45\textwidth}
    \centering
    \includegraphics[width=\linewidth]{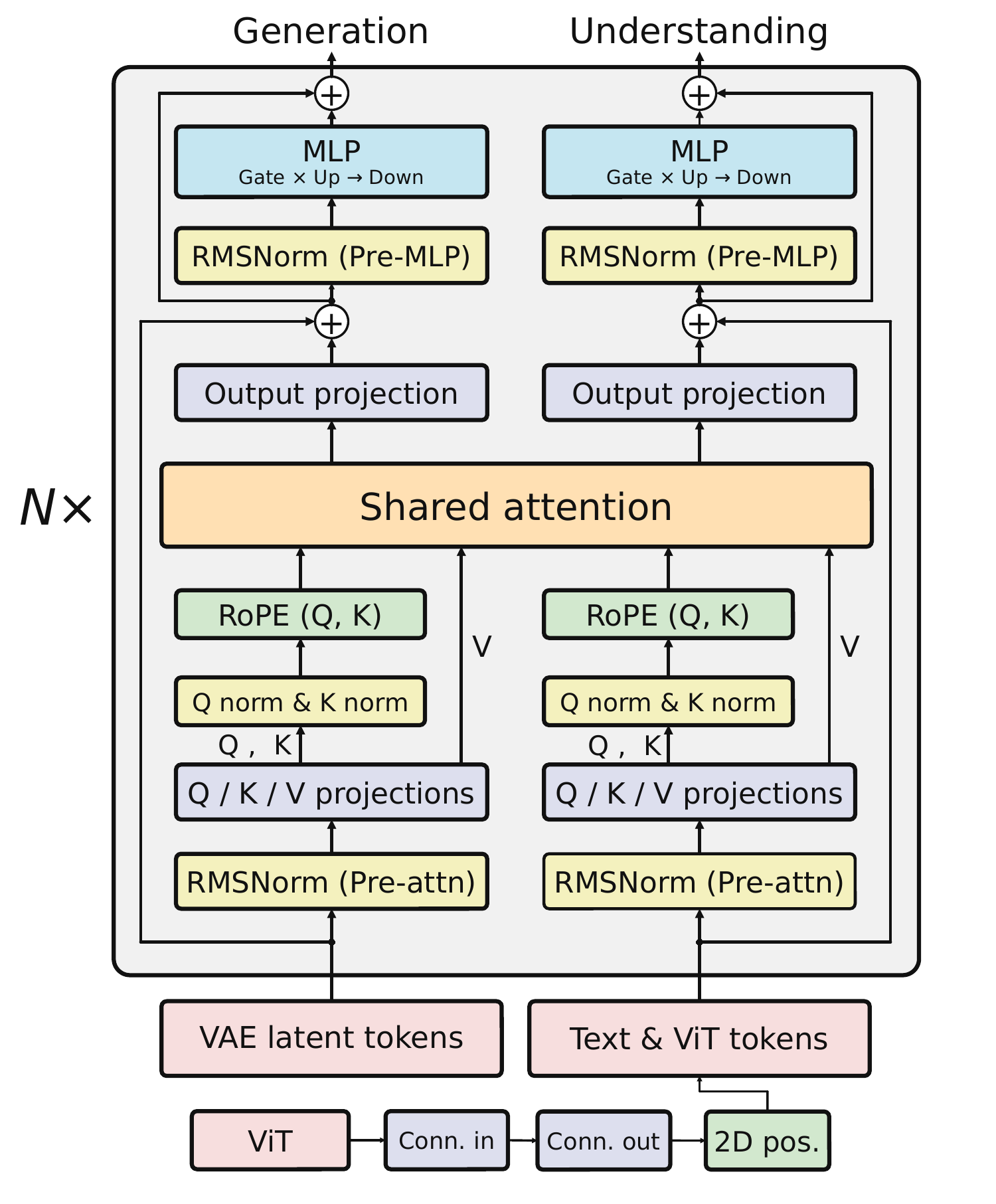}
    \caption{\textbf{BAGEL architecture.}}
    \label{fig:bagel-architecture}
\end{wrapfigure}
The base model we use, BAGEL, is a Mixture-of-Transformers model (Figure~\ref{fig:bagel-architecture}). Its generation and understanding branches have separate projections, normalization layers, and MLPs, and interact through shared attention.

\paragraph{Visual representations.}
BAGEL encodes images with both a SigLIP ViT and a VAE. The ViT uses $14\times14$-pixel patches and produces 1,152-dimensional features, which a two-layer MLP connector maps to the 3,584-dimensional transformer space. The VAE downsamples each spatial dimension by a factor of 8 and produces 16-channel latent maps. We group $2\times2$ latent cells into each token, which gives a 64-dimensional vector and an effective image-space stride of 16 pixels. Linear projections map these latent tokens into and out of the transformer, and timestep embeddings condition the latent inputs. Both visual token types receive fixed two-dimensional sinusoidal position embeddings. Images are resized bicubically to a maximum edge of 518 pixels for the ViT and 512 pixels for the VAE, with dimensions rounded to multiples of 14 and 16, respectively.

\paragraph{MoT routing and gradient flow.}
Text and ViT tokens are processed by the understanding expert, and VAE tokens by the generation expert. The two experts interact through joint attention within each example. I2I examples condition on both the ViT features and the clean VAE latents of the source image. Training perturbs the target-image latent as $z_t=(1-t)z+t\epsilon$, where $z$ is the encoded target and $\epsilon$ is Gaussian noise, and minimizes the mean squared error between the predicted flow and $\epsilon-z$. This loss updates the generation pathway and can also backpropagate through attention into the understanding expert, the ViT connector, and the ViT. I2T examples condition on the source-image ViT features and the instruction text, and are trained with autoregressive cross-entropy on the target permutation. Their gradients update the text output head, the understanding expert, the connector, and the ViT. Mixed training combines the text and latent losses in a single objective. The understanding pathway therefore receives training signal from both output modalities.

\paragraph{Trainable and frozen modules.}
Standard recipes finetune the ViT, its connector, and the MoT parameters used by the active objectives. When generation supervision is present, the latent input and output projections and the timestep network are also trained. The VAE encoder and decoder stay frozen throughout: images are encoded without gradient tracking, and I2T-only stages do not use the VAE. In the first stage of Frozen I2I $\rightarrow$ Mixed, the ViT and shared parameters are frozen, and only generation-specific parameters are
updated. The subsequent Mixed stage unfreezes these parameters in the language model, with the VAE still frozen.

\subsection{Instance-level agreement}
\label{app:controlled-eval}
\label{app:controlled-instance}
Table~\ref{tab:controlled_instance_alignment} compares paired I2I and I2T predictions on the full evaluation set of each task, which contains 1k I2I examples and their 1k corresponding I2T examples. For each task and training seed, both predictions come from the same checkpoint and the same input image and question. An I2T prediction is correct if its parsed permutation exactly matches the reference. We report the accuracy of each output modality and the I2T accuracy conditioned on whether the corresponding I2I prediction is correct. All values are averaged over the checkpoints trained on 30k I2I samples with one epoch and 1k I2T samples with 15 epochs on three different random seeds (Appendix~\ref{app:controlled-model}).

Because I2I predictions are images, we score them with a VLM judge, \texttt{gemini-3.7-flash}~\citep{gemini37flash}, for both tasks. Each request provides the task input, the reference target, and the generated candidate, in that order, and the judge returns one structured judgment. For Jigsaw, all four pieces must occupy their reference cells. For Zoom-In, all four views must follow the reference ordering from least to most zoomed, and the leftmost original-image anchor must be preserved. A deterministic check then marks a generation as correct only if all four positions match, every source piece or view appears exactly once, the task-fulfillment and content-fidelity scores are both at least 3 on a 0--4 scale, no decisive failure is reported, and the overall verdict of the judge is correct. Minor seams, blur, color shifts, and label-rendering differences are tolerated as long as content and ordering remain clear. Figures~\ref{fig:jigsaw-generation-prompt} and~\ref{fig:zoomin-generation-prompt} show the evaluation prompts.

%% file: iclr2026/sections/appendix/4_taskonomy_construction.tex
\section{Construction of \methodname}
\label{app:taxonomy-construction}
\subsection{Overview of the construction pipeline}
\label{app:construction-overview}

\begin{figure}[!htbp]
\centering
\includegraphics[width=\linewidth]{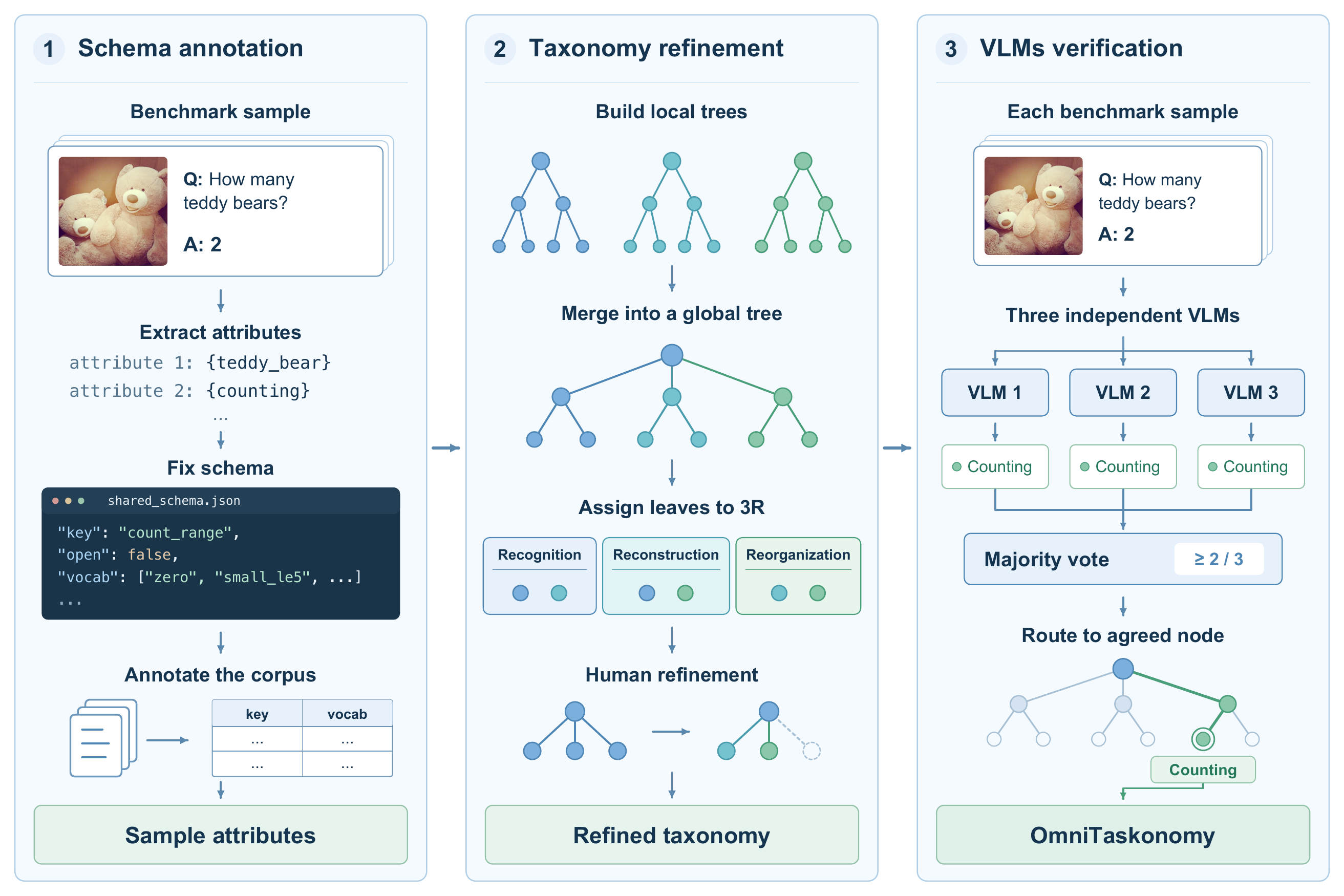}
\caption{\textbf{\methodname construction pipeline.} The three stages are schema annotation, taxonomy refinement, and VLM verification.}
\label{fig:construction-pipeline}
\end{figure}

Figure~\ref{fig:construction-pipeline} summarizes the three stages used to build \methodname. \emph{Schema annotation} extracts free-form attributes from a sample of benchmark questions, consolidates them into a shared schema, and uses this schema to annotate the full corpus. \emph{Taxonomy refinement} uses a VLM to group annotated samples into local trees and merge them into a global tree. We then review the leaf definitions, organize the leaves under the 3Rs with a VLM, and refine their boundaries by splitting, merging, and human review. \emph{VLM verification} has three VLMs assign a primary capability to each eligible example and retains examples with at least two matching votes. Appendices~\ref{app:construction-schema}, \ref{app:construction-trees}, and~\ref{app:unitaskonomy-annotation} describe the three stages in turn.

\clearpage
\subsection{From sample attributes to a fixed schema}
\label{app:construction-schema}

\paragraph{Extracting descriptive attributes.}
The first stage in Figure~\ref{fig:construction-pipeline} starts from 1,000 examples sampled from the seven benchmark collections. For each example, \texttt{gemini-3-flash-preview} receives the image(s), question, answer choices, and reference answer, and produces 50 to 80 free-form attributes. These attributes describe the image source and content, the required perceptual operations, task-specific difficulty, question--answer properties, and data quality. We include the reference answer because it helps identify which visual operation the question tests.

\paragraph{Fixing the schema and annotating samples.}
We pool the attribute names and their observed value frequencies, and ask the same model to merge synonymous keys into a shared schema. The resulting schema has 57 fields. Each field specifies a canonical key, a description, a vocabulary, whether values outside the vocabulary are allowed, and the variants merged into it. We then apply this fixed schema to all 10,422 samples in the benchmark collection. Each call receives one sample and the schema, and returns a flat attribute record with \texttt{n/a} for fields that do not apply.

Figure~\ref{fig:prompt-worked-example} walks through these steps for a CV-Bench counting example. The question ``How many walls are in the image?'' yields free-form attributes such as \texttt{main\_task: counting} and \texttt{count\_range: small\_le5}. The pooled schema defines \texttt{count\_range} with a closed vocabulary, and a separate pass annotates the example under this shared definition. As a result, fields are directly comparable across samples.

\input{iclr2026/sections/appendix/7_worked_example}

\clearpage
\subsection{From candidate trees to the refined 3R taxonomy}
\label{app:construction-trees}

\paragraph{Local grouping and global merging.}
We partition the attribute-annotated corpus from Appendix~\ref{app:construction-schema} into ten uniform batches. For each batch, \texttt{gemini-3.1-pro-preview}~\citep{gemini31pro} uses the schema fields to propose a hierarchy of image domains, broad abilities, finer skills, and task subtypes, and the prompt requires sibling nodes to partition their parent. The model then assigns each sample to one local leaf based on its images, question, reference answer, and attributes; when several visual operations are involved, it selects the primary one. The ten local trees contain 474 leaves in total. A merge step consolidates synonymous nodes into a global candidate tree with 42 leaves and records a mapping from every local leaf to a global leaf. Composing the two mappings gives each sample one provisional global leaf.

\paragraph{Organizing and refining the 3R leaves.}
We review the candidate vocabulary and write capability definitions, which yields a human-authored catalogue of 35 leaves. We then organize these leaves under Recognition, Reconstruction, and Reorganization~\citep{malik2016three}. For this first 3R assignment, \texttt{gpt-5.6-sol}~\citep{openai2026gpt56} receives the leaf definitions and the three family descriptions, and assigns each leaf to one family according to the visual information it requires: semantic content, geometric and photometric properties, or the grouping and organization of visual elements.

Human review then refines this hierarchy. We split leaves that mix distinct capabilities, merge redundant leaves, and sharpen inclusion and exclusion rules so that every sample has an unambiguous primary label. For example, camera-relative depth is separated from metric inter-object distance and physical size, while object, text, and interface localization are merged into referential localization. When a leaf is retired or narrowed, \texttt{gemini-3.7-flash}~\citep{gemini37flash} or a human annotator re-reviews its samples and either reroutes them to another leaf or excludes them; Appendix~\ref{app:unitaskonomy-annotation} reports these exclusions. Each retained sample is routed by its primary capability.

Table~\ref{tab:construction-counting-example} traces the counting example through the recorded local and global trees to its final 3R label. The final leaf keeps the counting operation, whereas image domain and small-count difficulty no longer define separate capability leaves.

\begin{table}[!htbp]
\centering\small
\caption{\textbf{A counting example through taxonomy construction.}}
\label{tab:construction-counting-example}
\begin{tabularx}{\linewidth}{@{}l X@{}}
\toprule
Stage & Recorded assignment \\
\midrule
Local tree & Natural and Surveillance Photos $\rightarrow$ Counting and Enumeration $\rightarrow$ Object Counting $\rightarrow$ Small Count Enumeration \\
Global candidate tree & Natural Images $\rightarrow$ Quantitative and Logical Reasoning $\rightarrow$ Object Counting $\rightarrow$ Small Count Enumeration \\
Final 3R label & Reorganization $\rightarrow$ Counting \\
\bottomrule
\end{tabularx}
\end{table}

\subsection{VLM annotation and exclusions}
\label{app:unitaskonomy-annotation}

After taxonomy refinement (Appendix~\ref{app:construction-trees}), three VLM judges annotate the final samples: \texttt{gemini-3.7-flash}, \texttt{gpt-5.6-sol}, and \texttt{claude-opus-4-8}~\citep{anthropic2026claudeopus48}. Each judge assigns a primary capability label, and we retain an example when at least two of the three labels agree. Of the 9,444 retained examples, 8,811 (93.30\%) receive unanimous labels and 633 (6.70\%) receive two-of-three agreement.
The remaining 978 examples are excluded: 949 during taxonomy refinement (Appendix~\ref{app:construction-trees}) and 29 for lacking a two-of-three majority.

Table~\ref{tab:app-annotation-examples} shows representative retained annotations and their routing rationales.

\begin{table}[!htbp]
\centering\small
\caption{\textbf{Representative retained annotations.} Questions are shortened for space; rationales summarize the recorded assignments.}
\label{tab:app-annotation-examples}
\begin{tabularx}{\linewidth}{@{}p{.19\linewidth} X X@{}}
\toprule
Capability & Benchmark question & Routing rationale \\
\midrule
Appearance understanding & MMStar: ``What is the dominant color in the image?'' & Queries a named appearance property. \\
Depth understanding & BLINK: ``Which point is closer to the camera?'' & Compares camera-relative depth. \\
Metric 3D relation & CV-Bench-3D: ``Which highlighted object is closer to the car in real-world distance?'' & Requires metric inter-object proximity. \\
2D spatial relation & CV-Bench-2D: ``Where is the mouse relative to the potted plant?'' & Queries their visible left/right arrangement. \\
Connectivity & BLINK: ``Is the bed touching the book?'' & Queries contact rather than metric distance. \\
Counting & CV-Bench-2D: ``How many trees are in the image?'' & Requires cardinality of a visually selected set. \\
\bottomrule
\end{tabularx}
\end{table}
\subsection{Human validation}
\label{app:human-validation}

\paragraph{Review setting.}
Four human reviewers audit 250 of the 9,444 retained I2T examples. To cover every capability, we randomly select one example from each of the 25 leaves, sample another 225 uniformly without replacement from the remaining pool, and shuffle the result with seed 20260919. Each reviewer assesses the same 50 shared examples and 50 distinct examples, which gives 100 judgments per reviewer and 400 in total. Reviewers see the image(s), question, answer choices, reference answer, assigned leaf, and the catalogue of leaf definitions and boundary rules. For each example, a reviewer can accept the assigned leaf, reject it and select a replacement, or reject it without a confident replacement (\texttt{unsure}). We analyze the final saved judgments, including revisions.

\paragraph{Acceptance of VLM labels.}
Reviewers accept the VLM label in 378 of the 400 judgments (94.50\%), select a replacement in 10, and mark 12 as \texttt{unsure}. Among the 388 definite judgments, 97.4\% accept the VLM label. Weighting examples equally instead of judgments gives the same 94.50\% acceptance rate. Of the 50 shared examples, 47 (94.00\%) are accepted by a strict majority of at least three reviewers, including 43 (86.00\%) accepted unanimously. Of the remaining three, two are split two--two and one is rejected by the majority.

\paragraph{Inter-reviewer agreement.}
We next compare the reviewers' final labels on the 50 shared examples
(Table~\ref{tab:human-reviewer-agreement}). A reviewer's final label is
the VLM-assigned leaf if accepted and the selected replacement leaf
otherwise. Because leaves are unordered categories, we report exact
agreement and Cohen's $\kappa$ for each reviewer pair, excluding examples
that either reviewer marked \texttt{unsure}.

\begin{table}[!htbp]
\centering\small
\setlength{\tabcolsep}{6pt}
\caption{\textbf{Pairwise agreement between human reviewers.}
Each comparison excludes examples that either reviewer marked
\texttt{unsure}; $n$ is the number of remaining shared examples.}
\label{tab:human-reviewer-agreement}
\begin{tabular}{@{}c r r r@{}}
\toprule
Reviewers & $n$ & Agree (\%) & $\kappa$ \\
\midrule
R1--R2 & 48 & 97.92 & 0.98 \\
R1--R3 & 47 & 95.74 & 0.95 \\
R1--R4 & 47 & 97.87 & 0.98 \\
R2--R3 & 48 & 91.67 & 0.91 \\
R2--R4 & 48 & 93.75 & 0.93 \\
R3--R4 & 47 & 95.74 & 0.95 \\
\bottomrule
\end{tabular}
\end{table}

Pairwise exact agreement ranges from 91.67\% to 97.92\%, and Cohen's
$\kappa$ from 0.91 to 0.98. Across all four reviewers, Fleiss'
$\kappa$ is 0.96 on the 46 shared examples without any \texttt{unsure} response.
The reviewers therefore agree closely on the final labels of the reviewed examples.

\subsection{Prompts for construction and annotation}
\label{app:construction-prompt-guide}

Appendix~\ref{app:prompts} collects the full prompt templates. Figures~\ref{fig:taxonomy-freeform}--\ref{fig:taxonomy-fill} cover attribute extraction, schema induction, and fixed-schema annotation (Appendix~\ref{app:construction-schema}). Figures~\ref{fig:taxonomy-local}--\ref{fig:taxonomy-3r} cover local tree induction, sample-to-leaf assignment, global merging, and 3R assignment (Appendix~\ref{app:construction-trees}). The human splitting, merging, and refinement steps do not use a model prompt.

Figure~\ref{fig:taxonomy-annotation} gives the shared instructions for the three VLM annotators, and Figure~\ref{fig:taxonomy-annotation-item} gives the per-example input and batch footer. Appendix~\ref{app:prompt-annotation} describes their inputs, model settings, and voting procedure.

%% file: iclr2026/sections/appendix/7_worked_example.tex
\begin{figure}[!htbp]
\centering
\begingroup
\lstset{style=paperprompt,basicstyle=\fontsize{7.8}{9.2}\selectfont\ttfamily}
\begin{minipage}[t]{0.49\linewidth}\vspace{0pt}
\begin{examplebox}[equal height group=prompt-example-top]{(a) A CV-Bench counting example}
\centering
\includegraphics[width=\linewidth,height=80pt,keepaspectratio]{}
\par\vspace{5pt}\raggedright
\textbf{Question.} How many walls are in the image?\\[3pt]
\textbf{Options.} A.~1 \quad B.~3 \quad C.~0.0 \quad D.~2.0\\[3pt]
\textbf{Reference answer.} A (one wall).
\end{examplebox}
\end{minipage}\hfill
\begin{minipage}[t]{0.49\linewidth}\vspace{0pt}
\begin{examplebox}[equal height group=prompt-example-top]{(b) Free-form attributes}
\textcolor{promptink}{\textbf{Image + question + answer $\rightarrow$ attributes}}
\par\vspace{4pt}
\begin{lstlisting}
{
  "image_source": "natural_photo",
  "main_task": "counting",
  "perceptual_sub_mechanism": "semantic_segmentation",
  "count_range": "small_le5",
  "occlusion_level": "partial"
}
\end{lstlisting}
\vspace{4pt}
Five of the 76 recorded attributes for this example.
\end{examplebox}
\end{minipage}

\vspace{8pt}
\begin{minipage}[t]{0.49\linewidth}\vspace{0pt}
\begin{examplebox}[equal height group=prompt-example-bottom]{(c) One canonical schema field}
\begin{lstlisting}
{
  "key": "count_range",
  "description": "The approximate number of items to be counted.",
  "open": false,
  "vocab": [
    "zero",
    "small_le5",
    "medium_6_20",
    "large_21_50",
    "very_large_50plus"
  ],
  "merged_variants": ["count_range"]
}
\end{lstlisting}
\end{examplebox}
\end{minipage}\hfill
\begin{minipage}[t]{0.49\linewidth}\vspace{0pt}
\begin{examplebox}[equal height group=prompt-example-bottom]{(d) Fixed-schema attributes}
\textcolor{promptink}{\textbf{Same example + shared schema}}
\par\vspace{4pt}
\begin{lstlisting}
{
  "image_source": "natural_photo",
  "main_task": "counting",
  "perceptual_sub_mechanism": "object_enumeration",
  "count_range": "small_le5",
  "occlusion_level": "partial",
  "estimated_difficulty": "easy"
}
\end{lstlisting}
\vspace{4pt}
Six of the 57 recorded fields. The value \texttt{small\_le5} is selected from the closed vocabulary for \texttt{count\_range}.
\end{examplebox}
\end{minipage}
\endgroup
\caption{\textbf{From an example to a shared schema.} Free-form and fixed-schema attributes come from separate calls on the same example.}
\label{fig:prompt-worked-example}
\end{figure}

%% file: iclr2026/sections/appendix/4_taskonomy.tex
\section{\methodname: Capabilities, data sources, and examples}
\label{app:unitaskonomy}

\subsection{Understanding capabilities}
\label{app:unitaskonomy-definitions}
Table~\ref{tab:app-capability-defs} describes the visual information required by each understanding capability and gives its number of retained evaluation examples ($N$).
\input{iclr2026/sections/appendix/generated/capability_definitions}
\subsection{I2I task inventory}
\label{app:unitaskonomy-i2i}
Table~\ref{tab:app-i2i-inventory} lists the 19 I2I tasks and the target image that each task produces: two in Recognition, nine in Reconstruction, and eight in Reorganization.

\begin{table}[!htbp]
\centering\small
\caption{\textbf{I2I tasks and their supervised targets.}}
\label{tab:app-i2i-inventory}
\begin{tabularx}{\linewidth}{@{}l X@{}}
\toprule
I2I task & Target image \\
\midrule
\multicolumn{2}{@{}l}{\textbf{Recognition}}\\
Object editing & An image with objects of a requested category added or replaced. \\
Attribute editing & An image with a specified color, material, or other appearance attribute changed. \\
\midrule
\multicolumn{2}{@{}l}{\textbf{Reconstruction}}\\
Colorization & The color image corresponding to a grayscale input. \\
Z-depth & Depth along the camera's viewing axis. \\
Euclidean depth & Radial distance from the camera. \\
Surface normals & Local 3D surface orientation. \\
Principal curvature & Local surface bending. \\
Occlusion edges & Boundaries induced by occlusion or depth discontinuities. \\
3D keypoints & Geometrically salient scene points. \\
Reshading & Image appearance under a shading target. \\
Inpainting & Completed rectangular regions of a masked image. \\
\midrule
\multicolumn{2}{@{}l}{\textbf{Reorganization}}\\
Semantic segmentation & A semantic class assignment to pixels or regions. \\
2D edges & Image-plane appearance boundaries. \\
2D keypoints & Salient image-plane locations. \\
2D segmentation & Grouping into image regions. \\
2.5D segmentation & Grouping into regions informed by surface geometry. \\
Object pointing & Marks on instances matching a category query. \\
Jigsaw & A correctly arranged image from shuffled patches. \\
Localization & A marked region matching a referring expression. \\
\bottomrule
\end{tabularx}
\end{table}

The inventory includes the dense prediction tasks formalized in Taskonomy~\citep{zamir2018taskonomy}, together with editing and puzzle tasks such as object editing, inpainting, and Jigsaw. We assign each I2I task to a family by the output it supervises, not by its dataset or by whether it uses semantic labels. I2I tasks and understanding capabilities remain separate leaves: an I2I task and an understanding capability can share a family without being the same task.

\subsection{Evaluation sources and coverage}
\label{app:unitaskonomy-induction}
\label{app:unitaskonomy-consolidation}

\begin{figure}[tb]
\centering
\includegraphics[width=\linewidth]{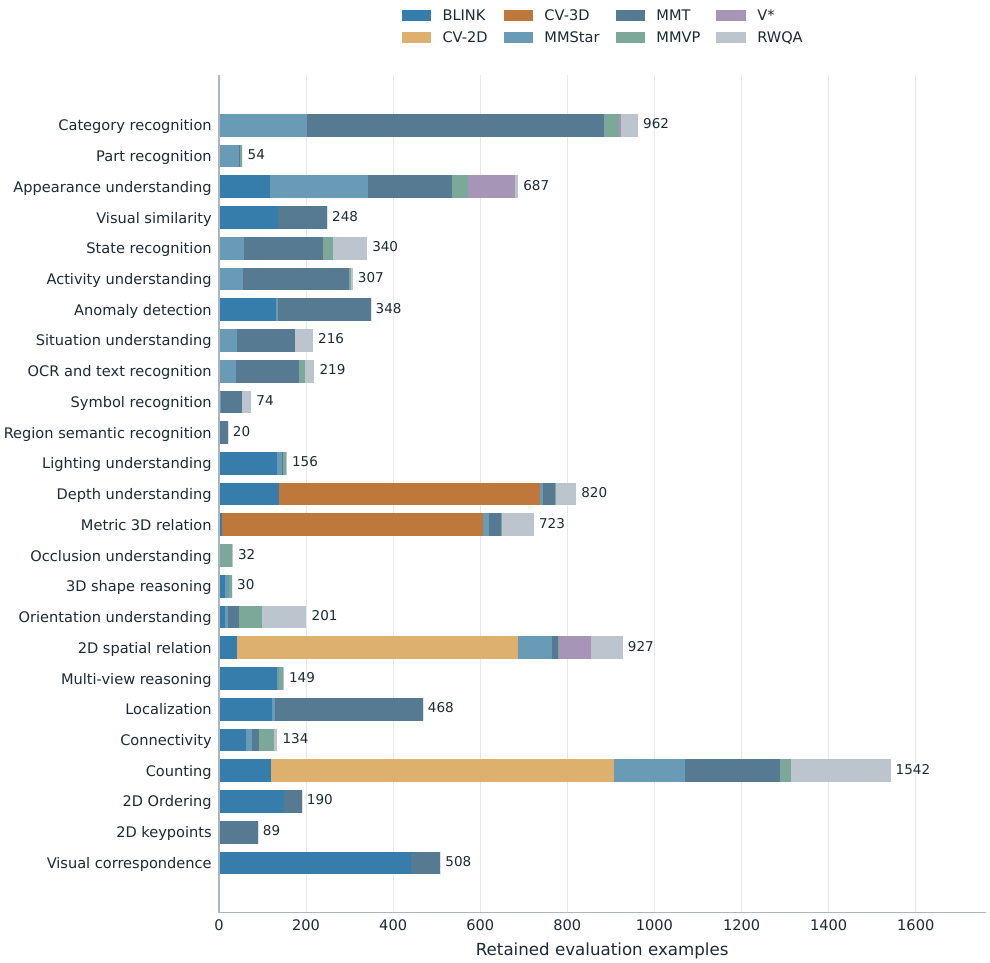}
\caption{\textbf{Capability coverage and benchmark composition.} Bar lengths give the number of retained evaluation examples per capability, with totals labeled, and colors indicate the source benchmark. The 19 capabilities with more than 100 examples appear in the main transfer map (Figure~\ref{fig:transfer}); Figure~\ref{fig:app-transfer-full} includes all 25.}
\label{fig:app-taxonomy-coverage}
\end{figure}

The evaluation corpus contains 10,422 examples from seven benchmark collections: BLINK, CV-Bench, MMStar, MMT-Bench, MMVP, VStarBench, and RealWorldQA. We count CV-Bench-2D and CV-Bench-3D as separate evaluation sets throughout, which gives eight sets in total. Of the 10,422 examples, 9,444 receive a retained capability label and 978 are excluded (Table~\ref{tab:app-benchmark-stats}). The retained examples span Recognition (3,475), Reconstruction (3,038), and Reorganization (2,931). Figure~\ref{fig:app-taxonomy-coverage} shows how each capability draws on the source benchmarks.

\input{iclr2026/sections/appendix/generated/benchmark_coverage}

\clearpage
\subsection{I2I training data}
\label{app:i2i-training-data}

Table~\ref{tab:i2i-training-sources} lists the training source for each of the 19 I2I tasks. Each task is trained in a separate run, and tasks that share source images use different targets.

\begin{table}[!htbp]
\centering\small
\setlength{\tabcolsep}{5pt}
\caption{\textbf{I2I training sources.} Each task uses 50,000 I2I training examples.}
\label{tab:i2i-training-sources}
\begin{tabular}{@{}ll@{}}
\toprule
Task & Source \\
\midrule
\multicolumn{2}{@{}l}{\textbf{Recognition}}\\
Object editing & ConceptEdit-12M, category-routed pairs \\
Attribute editing & ConceptEdit-12M, appearance-routed pairs \\
\midrule
\multicolumn{2}{@{}l}{\textbf{Reconstruction}}\\
Colorization & Taskonomy tiny \\
Z-depth & Taskonomy tiny \\
Euclidean depth & Taskonomy tiny \\
Surface normals & Taskonomy tiny \\
Principal curvature & Taskonomy tiny \\
Occlusion edges & Taskonomy tiny \\
3D keypoints & Taskonomy tiny \\
Reshading & Taskonomy tiny \\
Inpainting & COCO 2017 \citep{lin2014coco} + ConceptEdit originals \\
\midrule
\multicolumn{2}{@{}l}{\textbf{Reorganization}}\\
Semantic segmentation & COCO 2017 panoptic annotations \\
2D edges & Taskonomy tiny \\
2D keypoints & Taskonomy tiny \\
2D segmentation & Taskonomy tiny \\
2.5D segmentation & Taskonomy tiny \\
Object pointing & VisGym \\
Jigsaw & VisGym \\
Localization & RefCOCOg \citep{mao2016generation} \\
\bottomrule
\end{tabular}
\end{table}

\paragraph{Data from Taskonomy and VisGym.}
Twelve tasks use paired images and targets from the tiny split of the Taskonomy dataset~\citep{zamir2018taskonomy}. Jigsaw and object pointing use data from VisGym~\citep{wang2026visgym}.

\paragraph{Recognition editing pairs.}
Object editing and attribute editing use existing pairs of source and edited images from ConceptEdit-12M~\citep{cui2026conceptedit}. To select data pairs whose edits supervise a Recognition capability, we use \texttt{gpt-5.6-luna} to route which capability one given pair belongs to. For each pair, \texttt{gpt-5.6-luna} receives the edit-concept labels of the dataset, the short and detailed English instructions, and the source-image caption. It assigns the primary supervision of the edit to a Recognition capability, or rejects the pair as belonging to another family. 
We exclude pure object removal from Recognition because its target reconstructs the hidden background.

We keep pairs routed to category recognition for object editing and to appearance understanding for attribute editing, which gives eligible pools of 61,694 and 53,652 pairs, respectively. For each task, we sort the pool by sample identifier, shuffle it with seed 42, and take the first 50,000 pairs. In terms of the task groups defined by ConceptEdit, 69.31\% of the object-editing pairs come from the add/remove group and 30.23\% from replacement. For attribute editing, 59.02\% of the pairs come from color/material changes and 35.51\% from replacement. These groups only describe the composition of the selected data, which is determined by the capability-based routing. Each training example keeps the original short English edit instruction of its pair. Figures~\ref{fig:recognition-routing-system}--\ref{fig:recognition-training-instructions} give the routing templates and example training instructions.

\paragraph{Segmentation, inpainting, and localization.}
Semantic segmentation uses 50,000 distinct COCO train2017 images~\citep{lin2014coco}. We convert their panoptic annotations into a fixed 133-category color target, in which instances of the same category share a color and unlabeled pixels are black. Inpainting combines 38,400 COCO train2017 photographs (76.80\%) with 11,600 synthetic ConceptEdit source images (23.20\%). Each input contains one black rectangle that covers 15--35\% of the image area, with a width-to-height ratio between 0.5 and 2.0, and the target is the unchanged complete image.

Localization uses the RefCOCOg UMD training split~\citep{nagaraja2016refexp}, whose 42,226 distinct referring-object records are reused to reach 50,000 I2I training examples. The input is the source image with a referring expression, and the target image marks the bounding box of the referred object. Box coordinates are normalized to $[0,1]$, and each box is drawn in a random color.

\input{iclr2026/sections/appendix/generated/unitaskonomy_examples}

%% file: iclr2026/sections/appendix/generated/capability_definitions.tex
\begingroup
\footnotesize
\setlength{\LTcapwidth}{\linewidth}
\begin{longtable}{@{}>{\raggedright\arraybackslash}p{\dimexpr.29\linewidth-\tabcolsep\relax}>{\raggedleft\arraybackslash}p{\dimexpr.07\linewidth-2\tabcolsep\relax}>{\raggedright\arraybackslash}p{\dimexpr.64\linewidth-\tabcolsep\relax}@{}}
\caption{\textbf{Understanding capabilities in \methodname.} $N$ is the number of retained evaluation examples for each capability.}\label{tab:app-capability-defs}\\
\toprule Capability & $N$ & Description \\ \midrule \endfirsthead
\toprule Capability & $N$ & Description \\ \midrule \endhead
\bottomrule\endfoot
\bottomrule\endlastfoot
\multicolumn{3}{@{}l}{\textbf{Recognition}}\\*[2pt]
Category recognition & 962 & Recognize the semantic category, name, stable identity attribute, or known identity of an object, person, place, or scene. \\[2pt]
Part recognition & 54 & Identify or name a sub-part / component of an object (a handle, a wing, an engine bay). \\[2pt]
Appearance understanding & 687 & Understand visible appearance in the image plane, including named non-geometric properties and projected 2D shape, silhouette, outline, contour morphology, or edge structure. \\[2pt]
Visual similarity & 248 & Judge holistic visual resemblance, difference, or match among complete images, regions, or objects when no localized indexed correspondence and no specific appearance property is requested. \\[2pt]
State recognition & 340 & Recognize an ordinary temporary, operational, environmental, or emotional state without judging it against a norm. \\[2pt]
Activity understanding & 307 & Understand what activity or event is happening, how it develops, or which semantic stage of an activity or process is shown. \\[2pt]
Anomaly detection & 348 & Detect whether an observation departs from what is normal, correct, authentic, healthy, intact, or properly performed. \\[2pt]
Situation understanding & 216 & Understand what a whole situation means by combining multiple entities and contextual cues, including relationships, purpose, intent, affordance, safety, and narrative meaning. \\[2pt]
OCR and text recognition & 219 & Read written text, numbers, handwriting, or mathematical expressions rendered in the image, and verify that a given string is present. \\[2pt]
Symbol recognition & 74 & Recognize the conventional meaning of a non-text symbol, icon, pictogram, traffic sign, flag, map key, or legend item. \\[2pt]
Region semantic recognition & 20 & Connect semantic concepts to image regions by assigning class labels to pixels or regions across the image. \\[2pt]
\midrule
\multicolumn{3}{@{}l}{\textbf{Reconstruction}}\\*[2pt]
Lighting understanding & 156 & Illumination, shading, shadows, highlights, reflectance vs. illumination (intrinsic-image) reasoning, and light-source direction. \\[2pt]
Depth understanding & 820 & Recover camera-relative depth or depth order: z/radial distance from the observer, including pairwise front/behind and nearer/farther order along the viewing direction when no physical metric is requested. \\[2pt]
Metric 3D relation & 723 & Recover a real-world metric or scale-bearing relation among scene entities: inter-object distance, physical proximity, size, height, length, volume, or quantitative 3D layout. \\[2pt]
Occlusion understanding & 32 & Determine whether content is visible, occluded, cropped, or out of frame; recover the occluder/occluded order and causal boundary where one surface blocks another. \\[2pt]
3D shape reasoning & 30 & Reason about intrinsic 3D form that is independent of the observer, including solid identity, congruence, cross-section, folding and unfolding, or local surface bending such as curvature, ridges, and valleys. \\[2pt]
Orientation understanding & 201 & Understand the orientation or geometric configuration of a focal entity or visible surface, including heading, facing, gaze, articulated pose, door opening or swing direction, object rotation, relative axis alignment, surface normal, slant, tilt, or slope. \\[2pt]
2D spatial relation & 927 & Understand a directly visible, qualitative spatial relationship in the projected image plane, including position relative to the frame or to another visible entity, without inferring physical depth, metric geometry, or 3D orientation. \\[2pt]
Multi-view reasoning & 149 & Transform or integrate camera/observer coordinate frames: camera motion/pose, requested projection, novel view, multi-view layout, or egocentric navigation. \\[2pt]
\multicolumn{3}{@{}l}{\textbf{Reorganization}}\\*[2pt]
Localization & 468 & Return, select, or evaluate the image support of a target; point, box, crop, mask, polygon, foreground region, or 2D segmentation; regardless of whether the target is an object, part, text, symbol, or interface element. \\[2pt]
Connectivity & 134 & Understand whether entities, regions, or graph elements are connected through contact, attachment, containment, support, overlap, adjacency, intersection, a shared carrier, or a traversable path. \\[2pt]
Counting & 1542 & Count a visually selected set, compare set cardinalities, or derive a fraction, proportion, or percentage from visibly identifiable instances, parts, regions, marks, or text elements. \\[2pt]
2D ordering & 190 & Recover the original permutation of externally shuffled visual units using spatial or visual continuity. \\[2pt]
2D keypoints & 89 & Recover a predefined image-plane keypoint representation: corners, junctions, salient points, or 2D landmarks. \\[2pt]
Visual correspondence & 508 & Establish an explicit localized identity- or function-preserving mapping of a point, patch, part, or entity across indexed images, views, or frames, including tracking and identity-preserving trajectory continuation. \\[2pt]
\end{longtable}
\endgroup

%% file: iclr2026/sections/appendix/generated/benchmark_coverage.tex
\begin{table}[tb]
\centering
\small
\setlength{\tabcolsep}{4pt}
\caption{\textbf{Benchmark coverage.} CV-Bench is reported as separate 2D and 3D evaluation sets.}
\label{tab:app-benchmark-stats}
\begin{tabular}{@{}lrrrr@{}}
\toprule
Benchmark & Collected & Retained & Excluded & Retained (\%) \\
\midrule
BLINK & 1901 & 1758 & 143 & 92.5 \\
CV-Bench-2D & 1438 & 1434 & 4 & 99.7 \\
CV-Bench-3D & 1200 & 1200 & 0 & 100.0 \\
MMStar & 1500 & 986 & 514 & 65.7 \\
MMT-Bench & 3127 & 2840 & 287 & 90.8 \\
MMVP & 300 & 290 & 10 & 96.7 \\
VStarBench & 191 & 191 & 0 & 100.0 \\
RealWorldQA & 765 & 745 & 20 & 97.4 \\
Total & 10,422 & 9,444 & 978 & 90.6 \\
\bottomrule
\end{tabular}
\end{table}

%% file: iclr2026/sections/appendix/generated/unitaskonomy_examples.tex
\clearpage
\subsection{Representative examples across \methodname}
\label{app:unitaskonomy-examples}
{\small We illustrate all 19 I2I objectives and 25 understanding capabilities in \methodname,
grouped by the three families. I2I examples show input $\rightarrow$ reference target, and
I2T examples show an image, question, and reference answer.\par}
\vspace{6pt}

\begingroup
\definecolor{utrec}{HTML}{B77939}
\definecolor{utrcn}{HTML}{527EAF}
\definecolor{utrorg}{HTML}{3D8B71}
\definecolor{utink}{HTML}{243746}
\definecolor{utmuted}{HTML}{60717D}
\definecolor{utanswer}{HTML}{285E4C}
\definecolor{utbg}{HTML}{F3F5F7}
\newcommand{\utimg}[2]{{\setlength{\fboxsep}{0pt}\colorbox{utbg}{\begin{minipage}[c][#1][c]{\linewidth}\centering\includegraphics[width=\linewidth,height=#1,keepaspectratio]{#2}\end{minipage}}}}
\newcommand{\utimgplain}[2]{\begin{minipage}[c][#1][c]{\linewidth}\centering\includegraphics[width=\linewidth,height=#1,keepaspectratio]{#2}\end{minipage}}
\newcommand{\uttag}[1]{\hspace{2pt}\tcbox[on line,boxsep=0pt,left=2.2pt,right=2.2pt,top=1pt,bottom=.6pt,arc=1.6pt,boxrule=0pt,colback=utaccent!14!white,coltext=utaccent!85!black,fontupper=\sffamily\bfseries\fontsize{5.8}{6}\selectfont]{#1}}

\tcbset{utfamily/.style={enhanced,boxrule=.55pt,arc=2pt,outer arc=2pt,
  colback=white,coltitle=utink,coltext=utink,
  fonttitle=\sffamily\bfseries\fontsize{11}{13}\selectfont,
  fontupper=\fontsize{7.6}{8.8}\selectfont,
  left=9pt,right=9pt,top=5pt,bottom=6pt,toptitle=5pt,bottomtitle=5pt,
  before skip=0pt,after skip=0pt,valign=top,
  before upper={\raggedright\setlength{\parskip}{0pt}\setlength{\parindent}{0pt}}}}
\setlength{\parindent}{0pt}
\setlength{\parskip}{0pt}
\newsavebox{\utleft}
\newsavebox{\utright}
\newlength{\utrowheight}
\newlength{\utrowdepth}
\newlength{\utrulepad}
\newcommand{\utpair}[3][1]{%
  \sbox{\utleft}{#2}%
  \sbox{\utright}{#3}%
  \setlength{\utrowheight}{\ht\utleft}%
  \ifdim\ht\utright>\utrowheight\setlength{\utrowheight}{\ht\utright}\fi
  \setlength{\utrowdepth}{\dp\utleft}%
  \ifdim\dp\utright>\utrowdepth\setlength{\utrowdepth}{\dp\utright}\fi
  \addtolength{\utrowheight}{\utrulepad}%
  \addtolength{\utrowdepth}{\utrulepad}%
  \noindent\usebox{\utleft}\hfill%
  \ifnum#1=1 %
    {\color{utrule}\smash{\vrule width .3pt height\utrowheight depth\utrowdepth}}%
  \else\hspace{.3pt}\fi\hfill%
  \usebox{\utright}\par}
\newlength{\utcellw}
\newlength{\utW}\setlength{\utW}{50pt}
\newcommand{\utimgw}[1]{\includegraphics[width=\utW]{#1}}
\newcommand{\utarrow}[1]{\raisebox{.5\utWi}{\hspace{3pt}\begin{tabular}{@{}c@{}}{\fontsize{6}{6.5}\selectfont\color{utmuted}#1}\\[-1pt]\tikz\draw[->,>=latex,line width=.8pt,utaccent](0,0)--(15pt,0);\end{tabular}\hspace{3pt}}}
\newcommand{\utcell}[5]{\begin{minipage}[t]{.315\linewidth}\vspace{0pt}\raggedright
{\sffamily\bfseries\fontsize{7.4}{9}\selectfont\mbox{#2\hspace{1pt}\uttag{I2T}}\par}\vspace{4pt}\nointerlineskip%
\begin{minipage}[t][#1][t]{\linewidth}\vspace{0pt}\utimgw{#3}\end{minipage}\par\vspace{3pt}%
\textbf{Q:} #4\par\vspace{1pt}{\color{utanswer}\textbf{Ref:} #5\par}\end{minipage}}
\newcommand{\utmultiimg}[2]{\begin{minipage}[t]{\utW}\vspace{0pt}\centering\utimgw{#1}\par\vspace{1pt}{\fontsize{6.5}{7.5}\selectfont\color{utmuted}#2\par}\end{minipage}}
\newlength{\utWi}\setlength{\utWi}{46pt}
\newcommand{\utimgwi}[1]{\includegraphics[width=\utWi]{#1}}
\newcommand{\uticell}[6]{\begin{minipage}[t]{.315\linewidth}\vspace{0pt}\raggedright
{\sffamily\bfseries\fontsize{7.4}{9}\selectfont\mbox{#2\hspace{1pt}\uttag{I2I}}\par}\vspace{4pt}\nointerlineskip%
\begin{minipage}[t][#1][t]{\linewidth}\vspace{0pt}\utimgwi{#3}\utarrow{#5}\utimgwi{#4}\end{minipage}%
\if\relax\detokenize{#6}\relax\else\par\vspace{3pt}{\fontsize{7}{8.2}\selectfont #6\par}\fi\end{minipage}}
\newcommand{\utpairwide}[2]{%
  \sbox{\utleft}{#1}\sbox{\utright}{#2}%
  \setlength{\utrowheight}{\ht\utleft}\ifdim\ht\utright>\utrowheight\setlength{\utrowheight}{\ht\utright}\fi
  \setlength{\utrowdepth}{\dp\utleft}\ifdim\dp\utright>\utrowdepth\setlength{\utrowdepth}{\dp\utright}\fi
  \addtolength{\utrowheight}{\utrulepad}\addtolength{\utrowdepth}{\utrulepad}%
  \noindent\usebox{\utleft}\hfill{\color{utrule}\smash{\vrule width .3pt height\utrowheight depth\utrowdepth}}\hfill\usebox{\utright}\par}
\newsavebox{\utmid}
\newcommand{\uttriple}[3]{%
  \sbox{\utleft}{#1}\sbox{\utmid}{#2}\sbox{\utright}{#3}%
  \setlength{\utrowheight}{\ht\utleft}%
  \ifdim\ht\utmid>\utrowheight\setlength{\utrowheight}{\ht\utmid}\fi
  \ifdim\ht\utright>\utrowheight\setlength{\utrowheight}{\ht\utright}\fi
  \setlength{\utrowdepth}{\dp\utleft}%
  \ifdim\dp\utmid>\utrowdepth\setlength{\utrowdepth}{\dp\utmid}\fi
  \ifdim\dp\utright>\utrowdepth\setlength{\utrowdepth}{\dp\utright}\fi
  \addtolength{\utrowheight}{\utrulepad}\addtolength{\utrowdepth}{\utrulepad}%
  \noindent\usebox{\utleft}\hfill{\color{utrule}\smash{\vrule width .3pt height\utrowheight depth\utrowdepth}}\hfill
  \usebox{\utmid}\hfill\ifdim\ht\utright>0pt{\color{utrule}\smash{\vrule width .3pt height\utrowheight depth\utrowdepth}}\else\hspace{.3pt}\fi\hfill
  \usebox{\utright}\par}
\newcommand{\utrowseparator}[1]{%
  \par\vspace{\utrulepad}%
  {\color{utrule}\hrule height #1}\vspace{\utrulepad}}

\colorlet{utrule}{utrec!25!white}\colorlet{utaccent}{utrec}
\setlength{\utrulepad}{3pt}
\begin{tcolorbox}[utfamily,colframe=utrec!45!white,colbacktitle=utrec!8!white,title={Recognition}]
\utpair{%
\begin{minipage}[t]{.484\linewidth}\vspace{0pt}\raggedright
{\sffamily\bfseries\fontsize{8.1}{9.5}\selectfont Object editing \uttag{I2I}\par}\vspace{4pt}
\utimgw{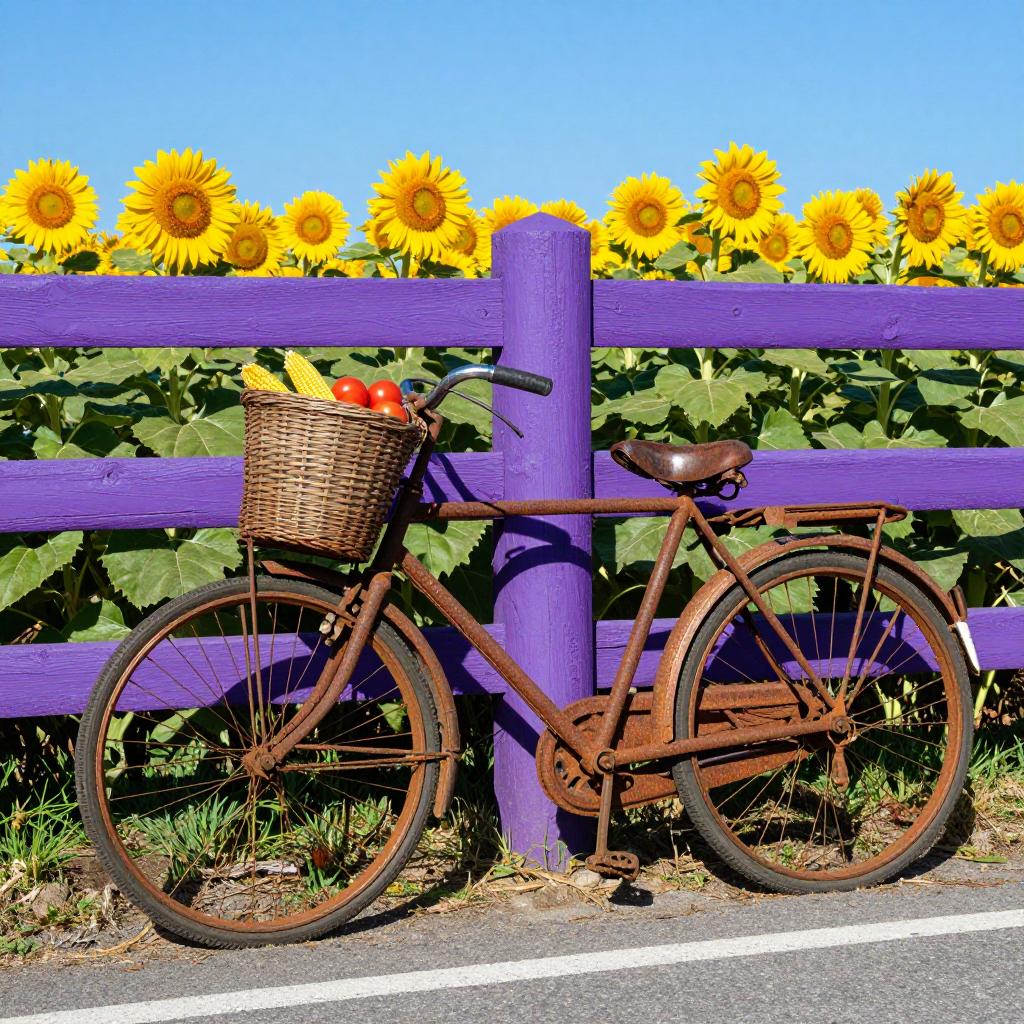}\utarrow{}\utimgw{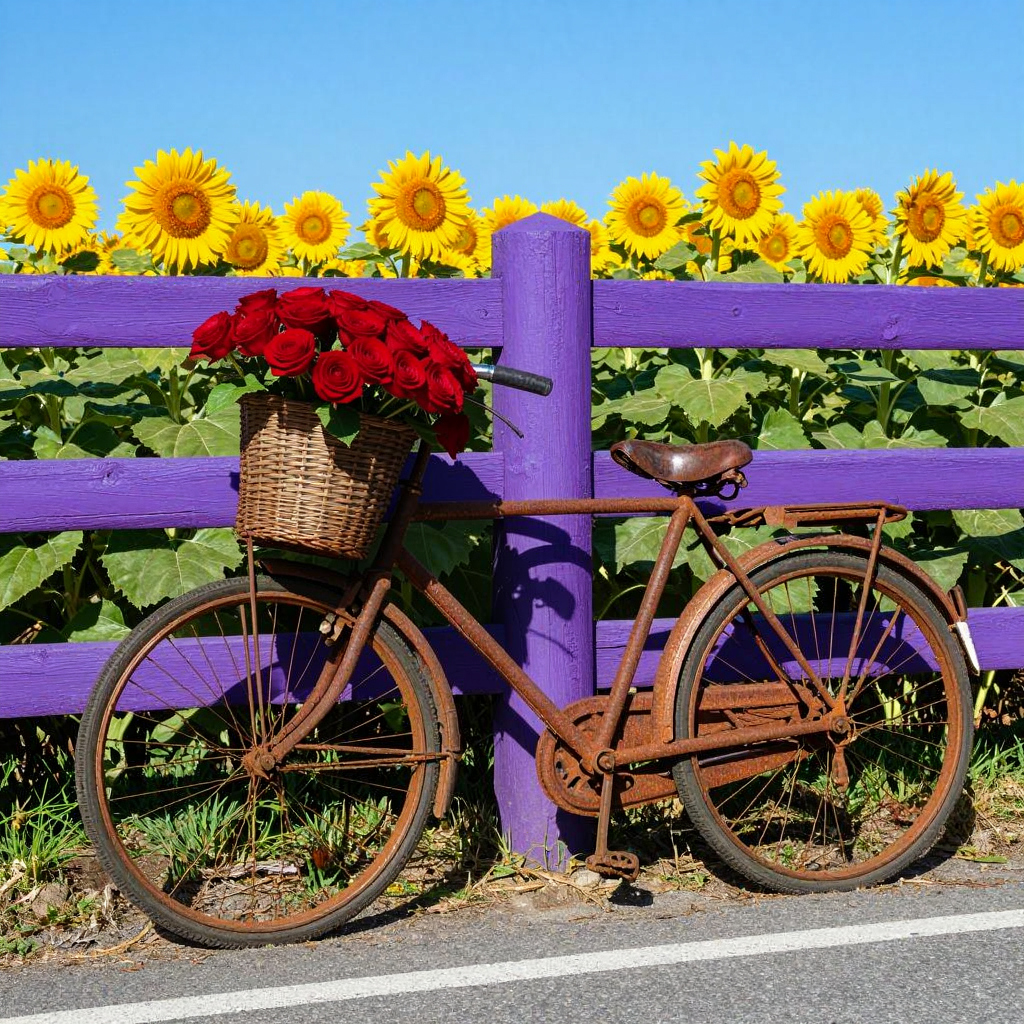}\par\vspace{3pt}
\textbf{Instruction:} Swap the corn and tomatoes in the bicycle basket with a bouquet of red roses.\par
\end{minipage}
}{%
\begin{minipage}[t]{.484\linewidth}\vspace{0pt}\raggedright
{\sffamily\bfseries\fontsize{8.1}{9.5}\selectfont Attribute editing \uttag{I2I}\par}\vspace{4pt}
\utimgw{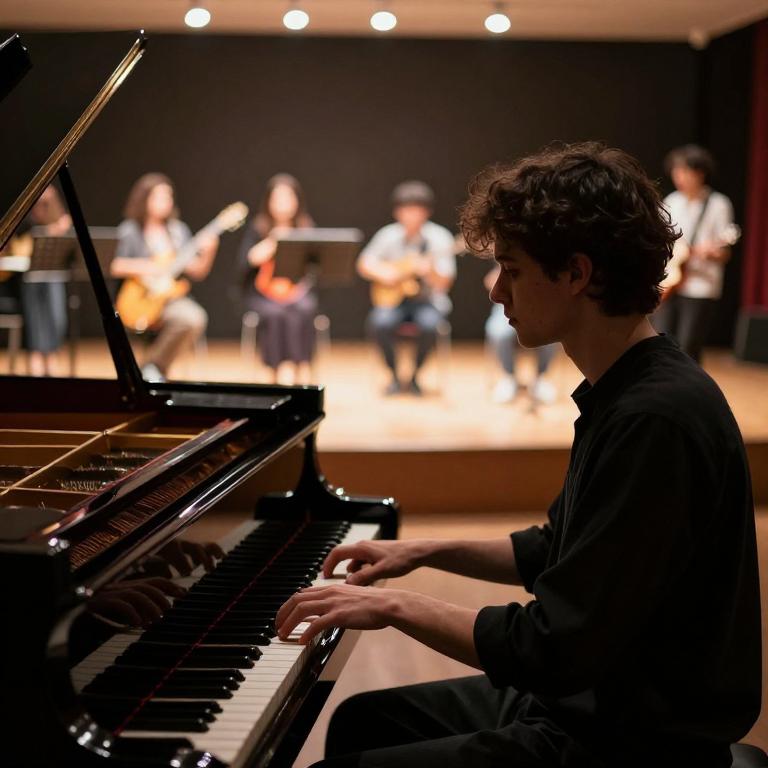}\utarrow{}\utimgw{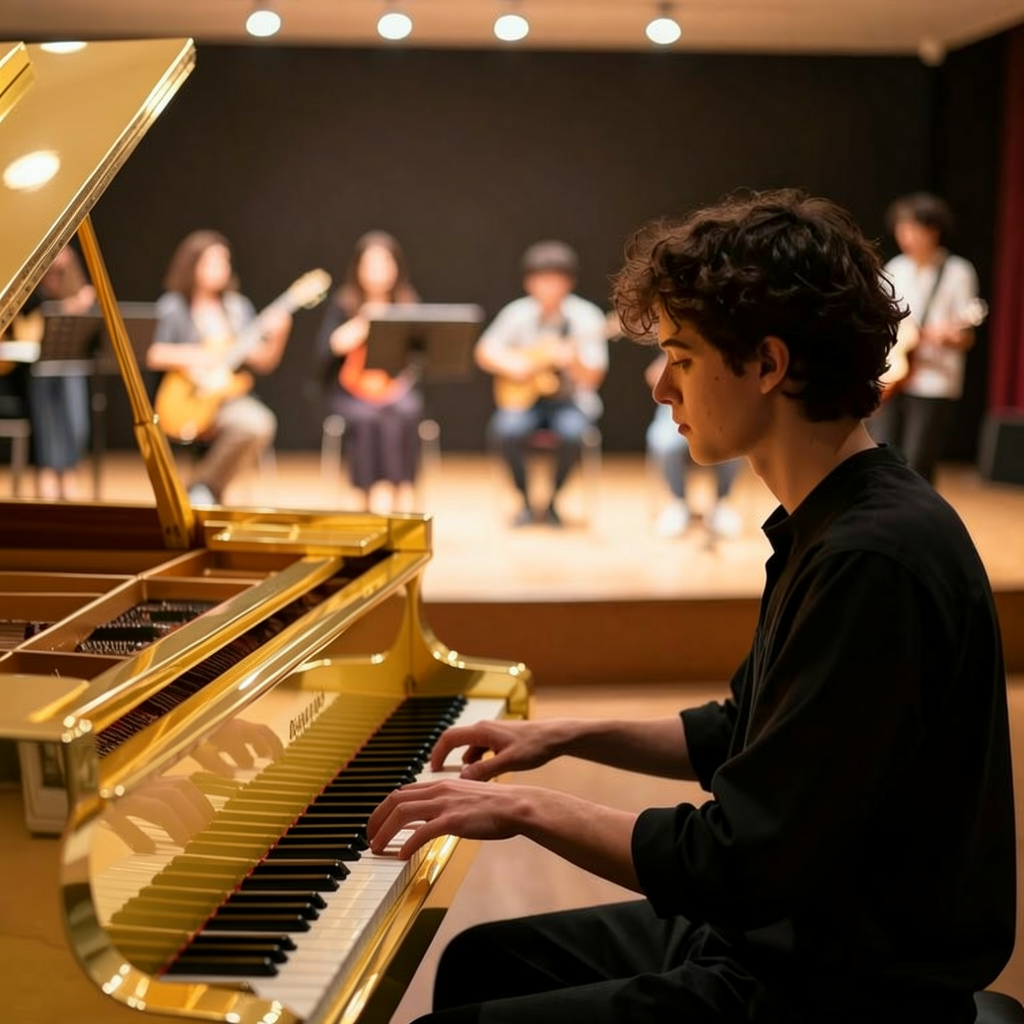}\par\vspace{3pt}
\textbf{Instruction:} Change the black grand piano to a gold finish.\par
\end{minipage}
}
\utrowseparator{.55pt}
\uttriple{%
\utcell{39.0pt}{Category recognition}{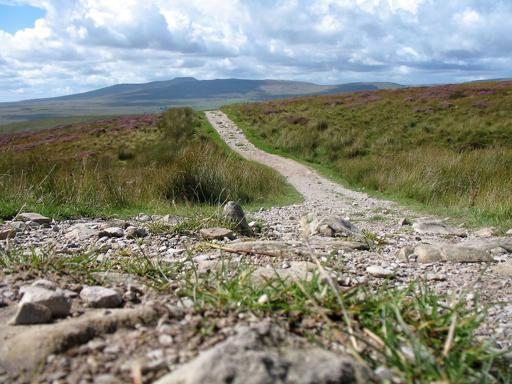}{What dominates the foreground?}{C\,---\, A dirt road leading up a grassy hill to a rocky path}%
}{%
\utcell{39.0pt}{Appearance understanding}{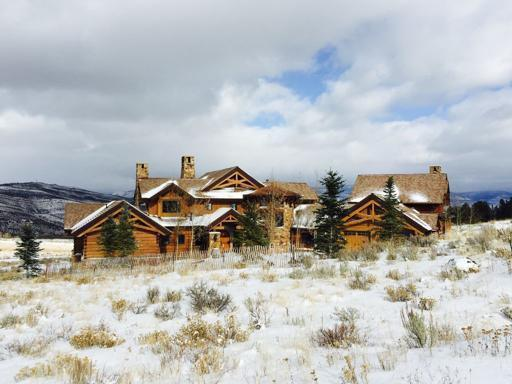}{What is the predominant color?}{B\,---\, White}%
}{%
\utcell{39.0pt}{Part recognition}{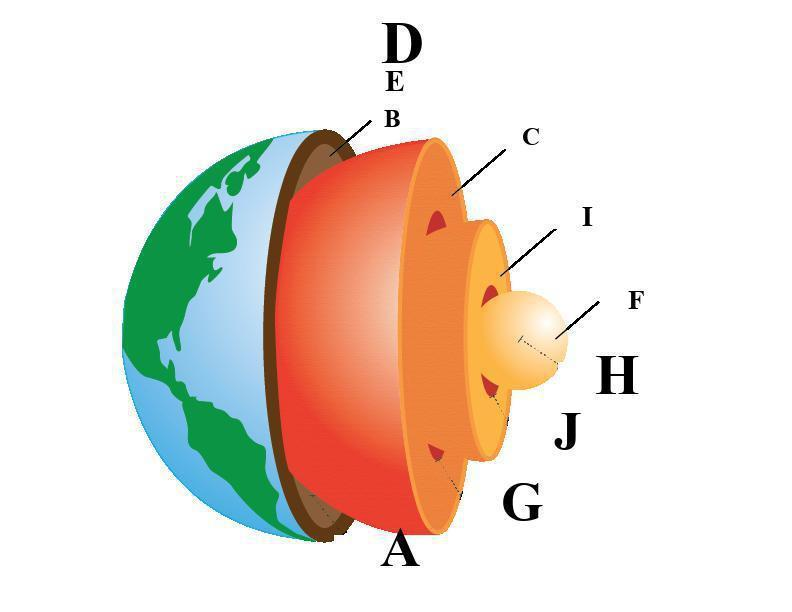}{Which labeled region is the crust?}{B\,---\, b}%
}
\utrowseparator{.3pt}
\uttriple{%
\utcell{50.5pt}{OCR and text recognition}{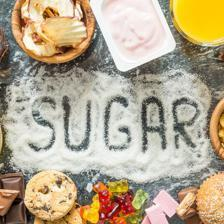}{Which letters are visible?}{A\,---\, SUGAR}%
}{%
\utcell{50.5pt}{Activity understanding}{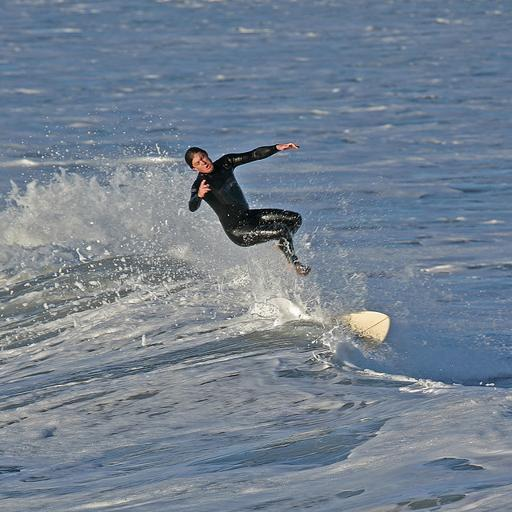}{What is the main activity?}{C\,---\, A man surfing}%
}{%
\utcell{50.5pt}{Symbol recognition}{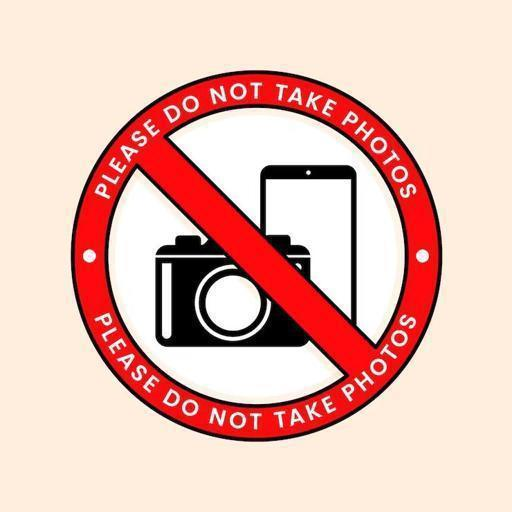}{What does this sign mean?}{C\,---\, No photography allowed}%
}
\utrowseparator{.3pt}
\uttriple{%
\utcell{75.5pt}{Situation understanding}{}{What is the likely purpose of this setting?}{A\,---\, A photoshoot for greeting cards}%
}{%
\utcell{75.5pt}{Region semantic recognition}{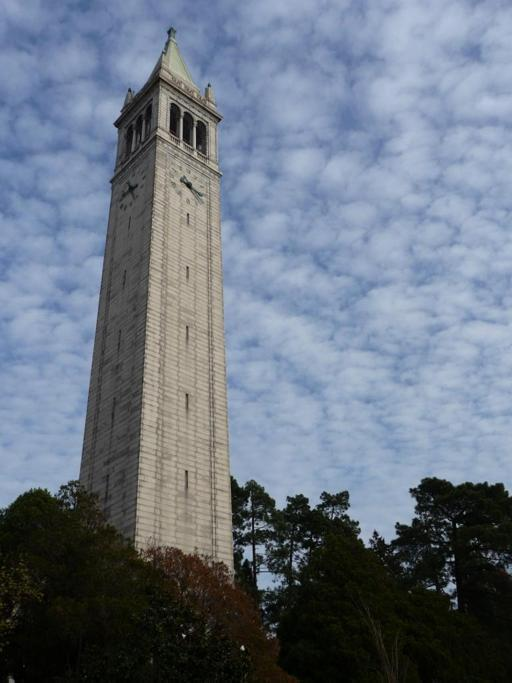}{Which category is at pixel (1.184, 0.051) in the 512-by-683 image?}{C\,---\, tree}%
}{%
\utcell{75.5pt}{State recognition}{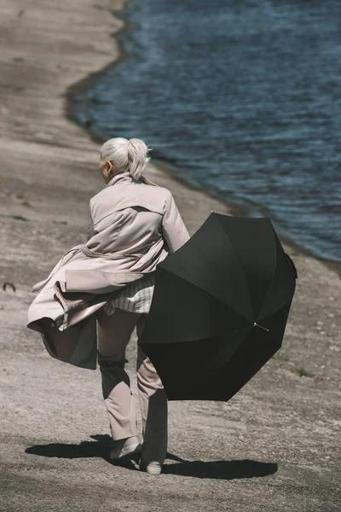}{What weather is depicted?}{C\,---\, windy}%
}
\utrowseparator{.3pt}
\begin{minipage}[t]{\linewidth}\vspace{0pt}\raggedright
{\sffamily\bfseries\fontsize{8.1}{9.5}\selectfont Visual similarity \uttag{I2T}\par}\vspace{4pt}\nointerlineskip
\noindent\utmultiimg{}{Reference}\hspace{6pt}%
\utmultiimg{}{A (image 2)}\hspace{6pt}%
\utmultiimg{}{B (image 3)}\hspace{14pt}%
\begin{minipage}[t]{\dimexpr\linewidth-3\utW-28pt\relax}\vspace{0pt}\raggedright\textbf{Q:} Which candidate best matches the reference?\par\vspace{2pt}{\color{utanswer}\textbf{Ref:} A\,---\, the second image\par}\end{minipage}
\end{minipage}\par
\utrowseparator{.3pt}
\begin{minipage}[t]{\linewidth}\vspace{0pt}\raggedright
{\sffamily\bfseries\fontsize{8.1}{9.5}\selectfont Anomaly detection \uttag{I2T}\par}\vspace{4pt}\nointerlineskip
\noindent\utmultiimg{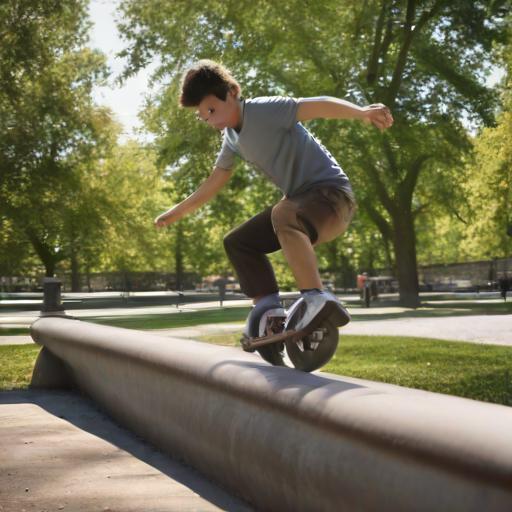}{A}\hspace{6pt}%
\utmultiimg{}{B}\hspace{6pt}%
\utmultiimg{}{C}\hspace{6pt}%
\utmultiimg{}{D}\hspace{14pt}%
\begin{minipage}[t]{\dimexpr\linewidth-4\utW-34pt\relax}\vspace{0pt}\raggedright\textbf{Q:} Which image is most likely to be a real photograph?\par\vspace{2pt}{\color{utanswer}\textbf{Ref:} C\,---\, the third image\par}\end{minipage}
\end{minipage}\par
\end{tcolorbox}
\clearpage
\colorlet{utrule}{utrcn!25!white}\colorlet{utaccent}{utrcn}
\setlength{\utrulepad}{3pt}
\begin{tcolorbox}[utfamily,colframe=utrcn!45!white,colbacktitle=utrcn!8!white,title={Reconstruction}]
\uttriple{%
\uticell{46.5pt}{Colorization}{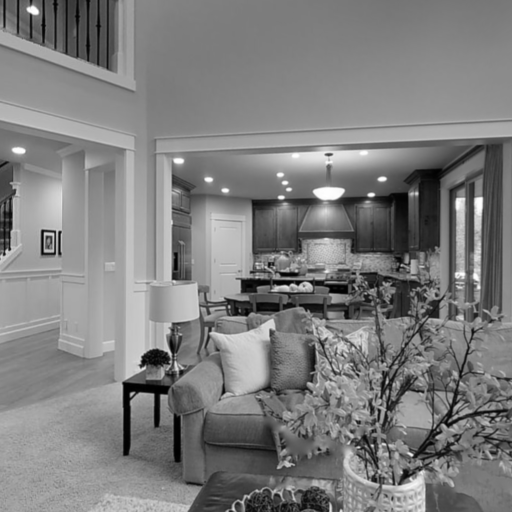}{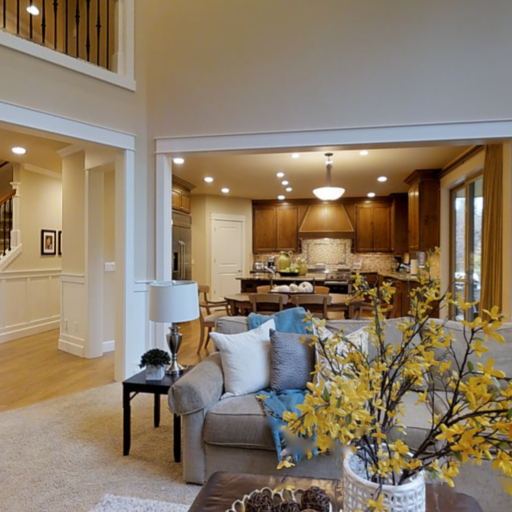}{}{}%
}{%
\uticell{46.5pt}{Z-depth}{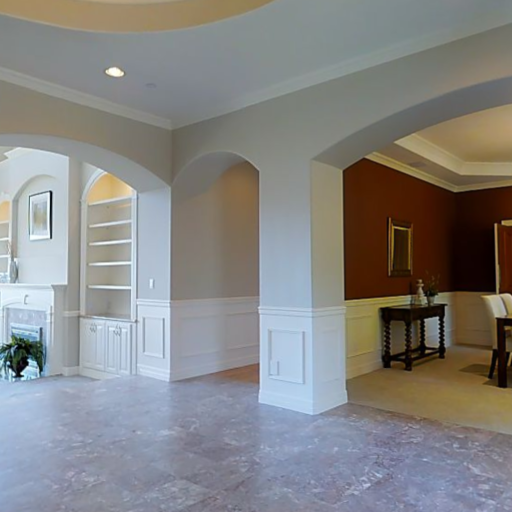}{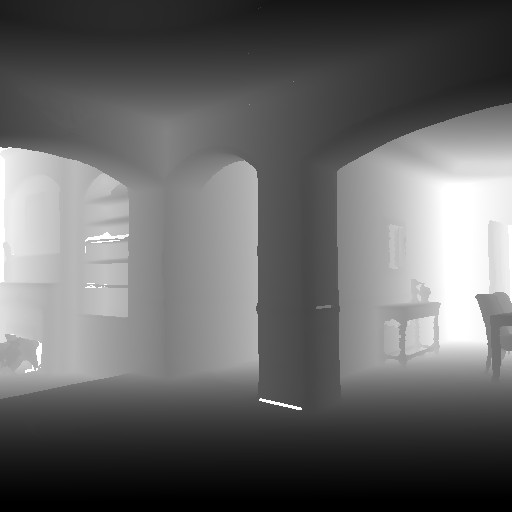}{}{}%
}{%
\uticell{46.5pt}{Euclidean depth}{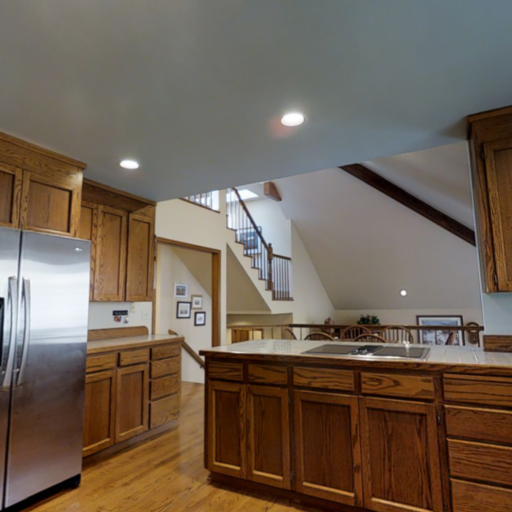}{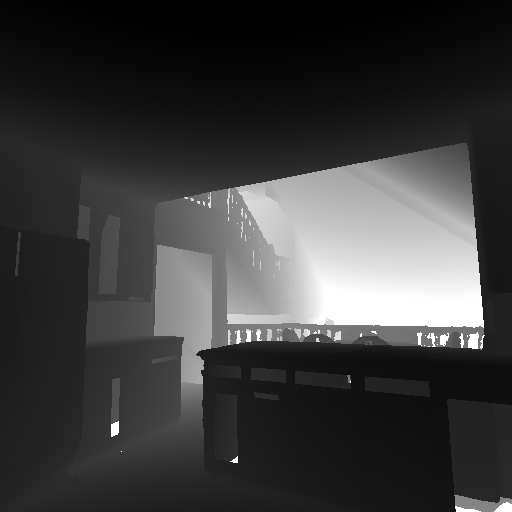}{}{}%
}
\utrowseparator{.3pt}
\uttriple{%
\uticell{46.5pt}{Surface normals}{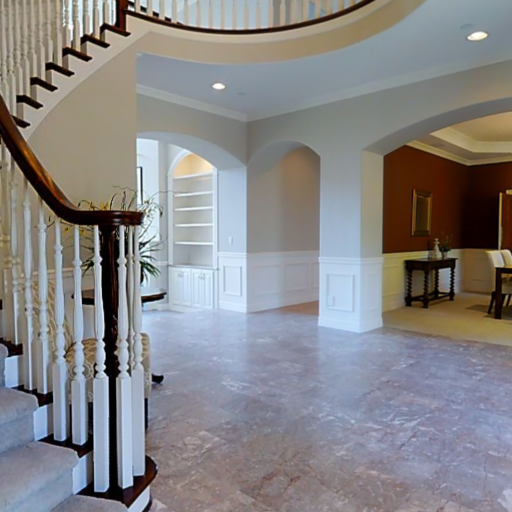}{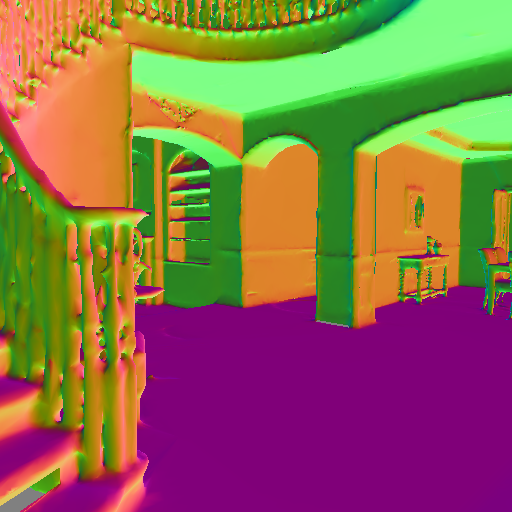}{}{}%
}{%
\uticell{46.5pt}{Principal curvature}{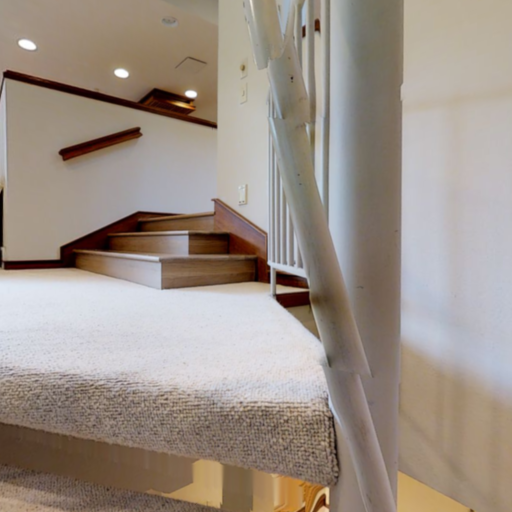}{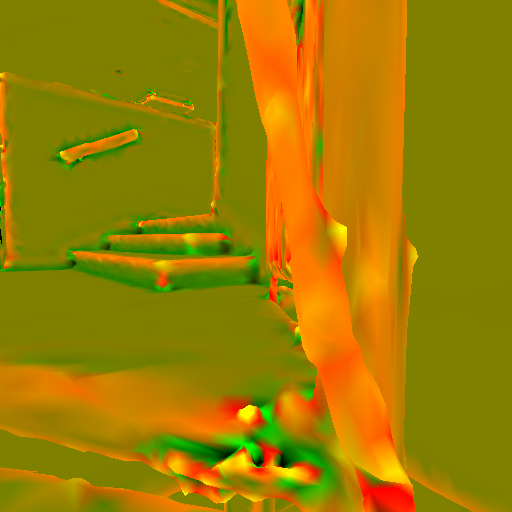}{}{}%
}{%
\uticell{46.5pt}{Occlusion edges}{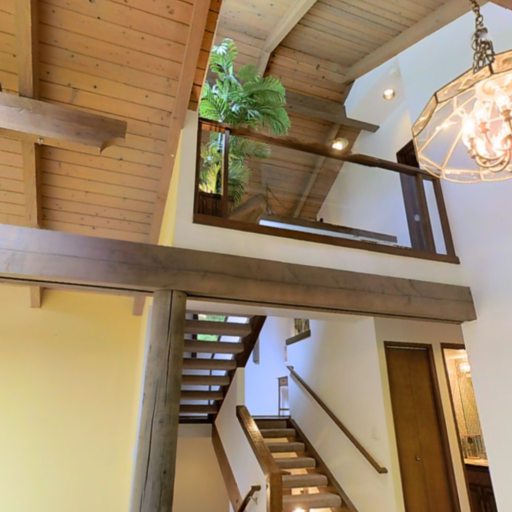}{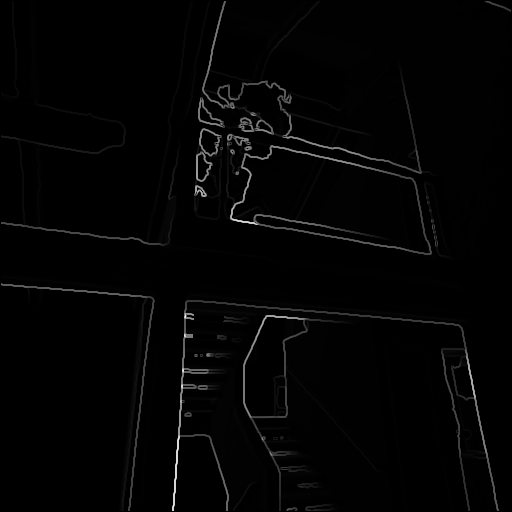}{}{}%
}
\utrowseparator{.3pt}
\uttriple{%
\uticell{46.5pt}{3D keypoints}{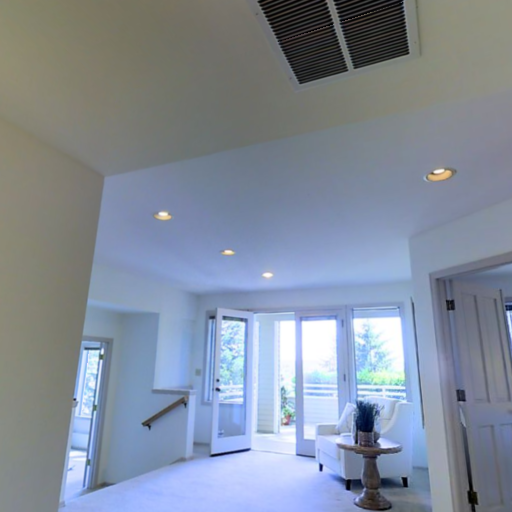}{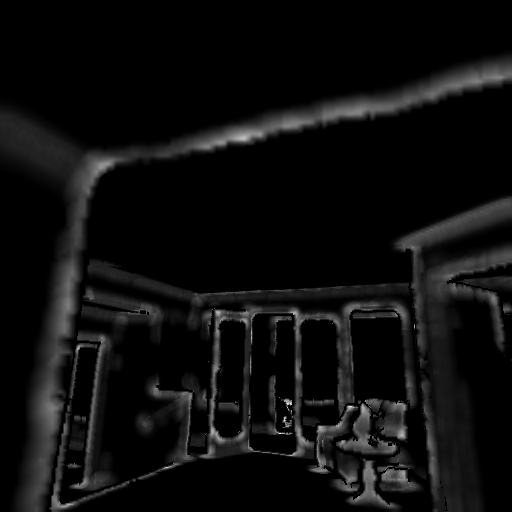}{}{}%
}{%
\uticell{46.5pt}{Reshading}{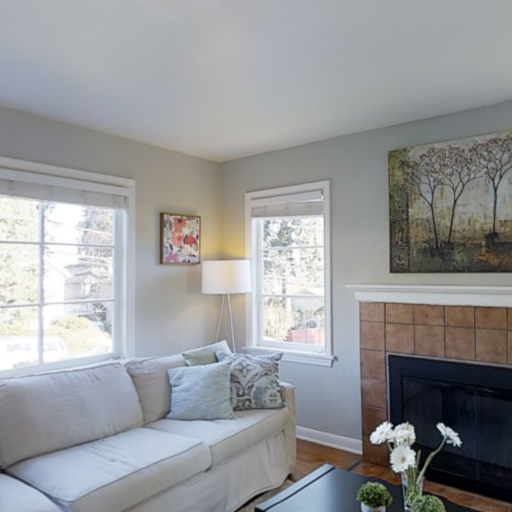}{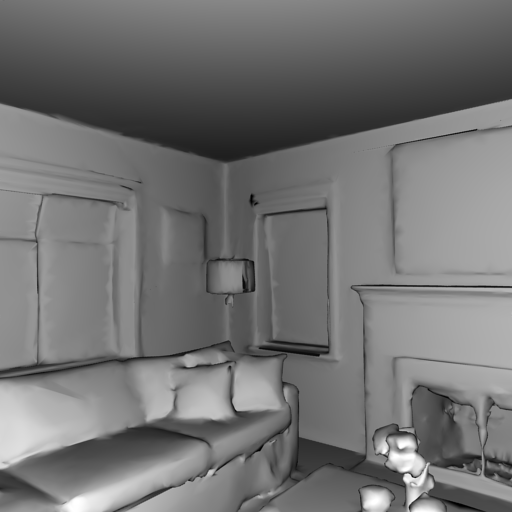}{}{}%
}{%
\uticell{35.0pt}{Inpainting}{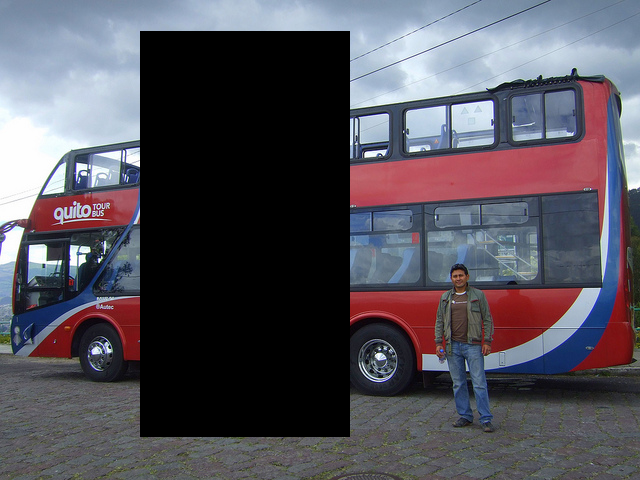}{}{}{\textbf{Instruction:} Inpaint the black rectangular region to reconstruct the complete original image. Keep all visible areas unchanged. Output only the completed image.}%
}
\utrowseparator{.55pt}
\uttriple{%
\utcell{38.0pt}{Lighting understanding}{}{Ignoring shading, which point has darker surface color: A, B, or the same?}{A\,---\, A is darker}%
}{%
\utcell{38.0pt}{Depth understanding}{}{Where is the sheep?}{B\,---\, The sheep is in the front of the car}%
}{%
\utcell{38.0pt}{Metric 3D relation}{}{Is the airplane far from the bicycle?}{A\,---\, yes}%
}
\utrowseparator{.3pt}
\uttriple{%
\utcell{65.5pt}{Occlusion understanding}{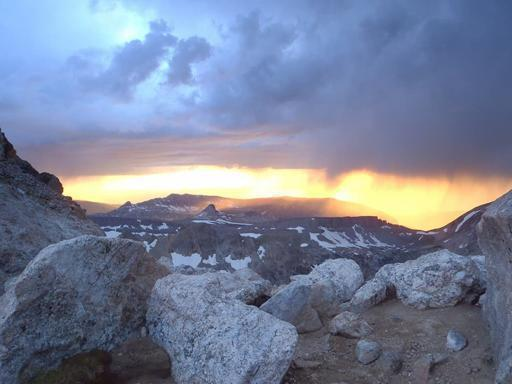}{What is the position of the sun?}{A\,---\, Hidden by storm clouds}%
}{%
\utcell{65.5pt}{Orientation understanding}{}{Which direction is the dog facing?}{A\,---\, The dog is facing forward.}%
}{%
\utcell{65.5pt}{2D spatial relation}{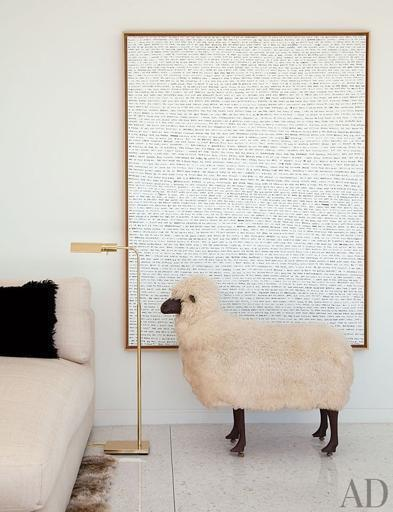}{Where are the sheep's legs in the image?}{B\,---\, On the right side}%
}
\utrowseparator{.3pt}
\uttriple{%
\utcell{75.5pt}{3D shape reasoning}{}{Which cube matches the unfolded net?}{D}%
}{%
\utcell{75.5pt}{Multi-view reasoning}{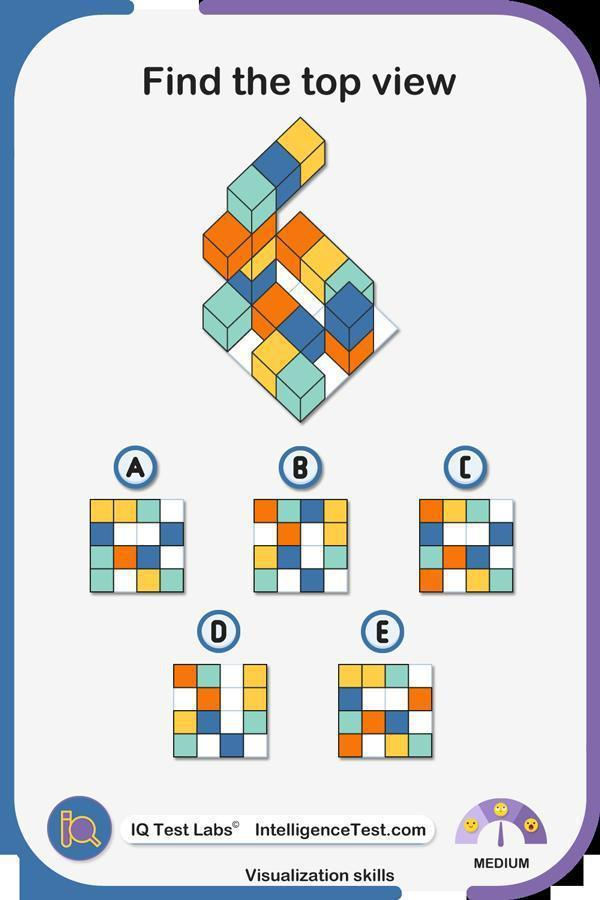}{Which option is the top view?}{B}%
}{%
\rule{.315\linewidth}{0pt}%
}
\end{tcolorbox}
\clearpage
\colorlet{utrule}{utrorg!25!white}\colorlet{utaccent}{utrorg}
\setlength{\utrulepad}{3pt}
\begin{tcolorbox}[utfamily,colframe=utrorg!45!white,colbacktitle=utrorg!8!white,title={Reorganization}]
\uttriple{%
\uticell{46.5pt}{2D segmentation}{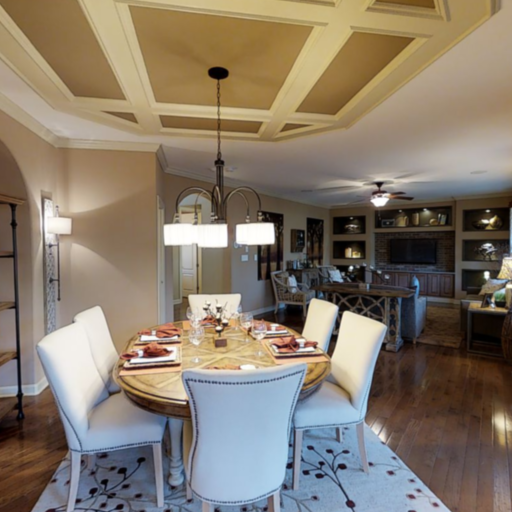}{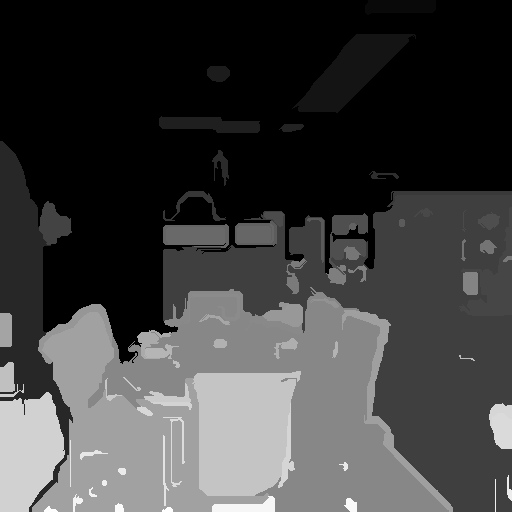}{}{}%
}{%
\uticell{46.5pt}{2D edges}{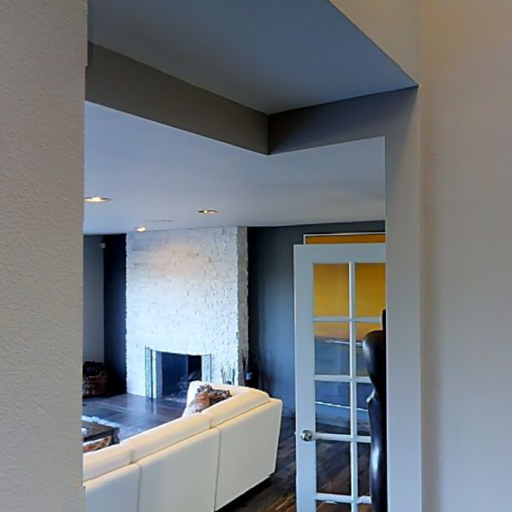}{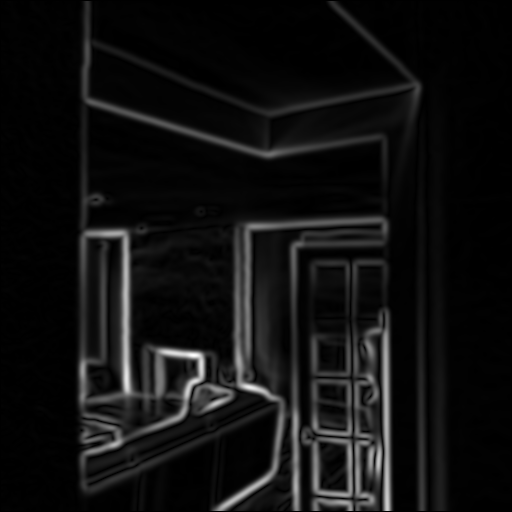}{}{}%
}{%
\uticell{46.5pt}{2D keypoints}{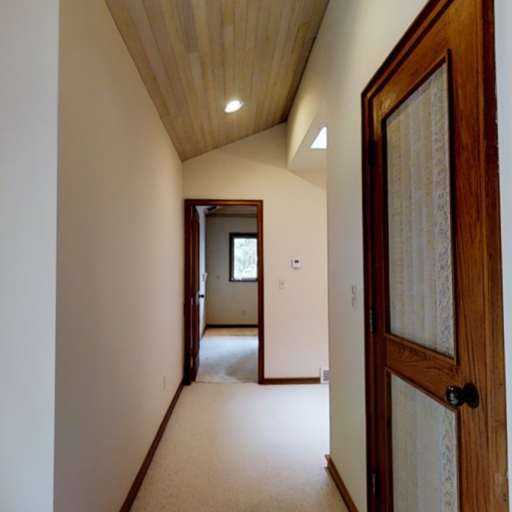}{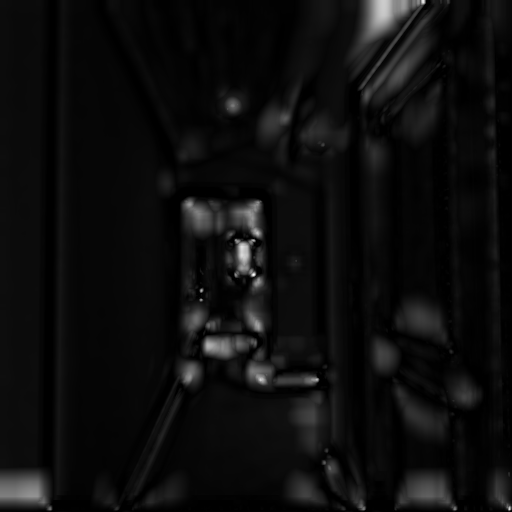}{}{}%
}
\utrowseparator{.3pt}
\uttriple{%
\uticell{70.0pt}{2.5D segmentation}{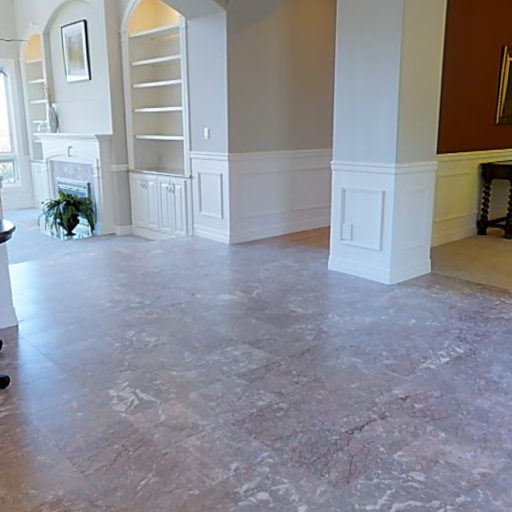}{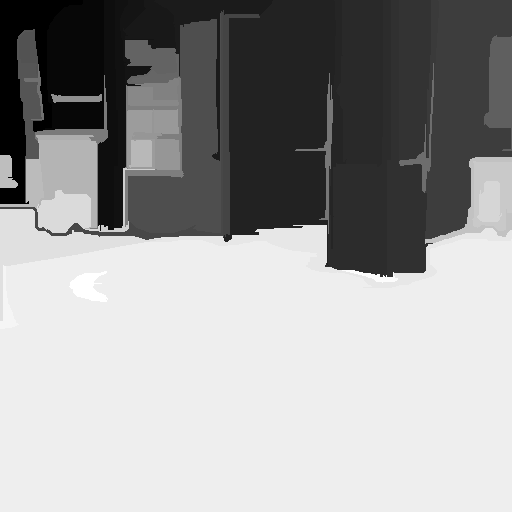}{}{}%
}{%
\uticell{70.0pt}{Semantic segmentation}{}{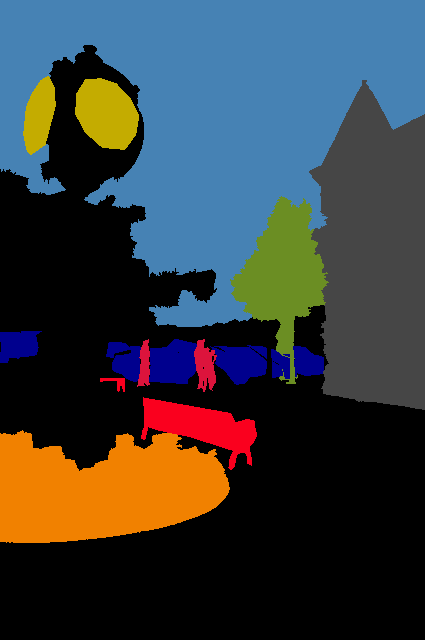}{}{}%
}{%
\uticell{70.0pt}{Object pointing}{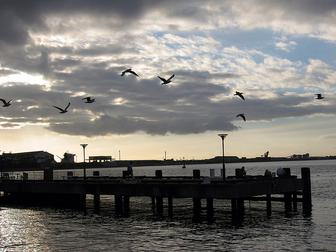}{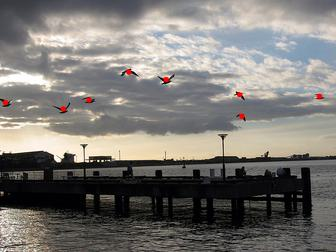}{gull}{}%
}
\utrowseparator{.3pt}
\utpairwide{%
\begin{minipage}[t]{.645\linewidth}\vspace{0pt}\raggedright
{\sffamily\bfseries\fontsize{7.4}{9}\selectfont Jigsaw\hspace{1pt}\uttag{I2I}\par}\vspace{4pt}\nointerlineskip
\utimgwi{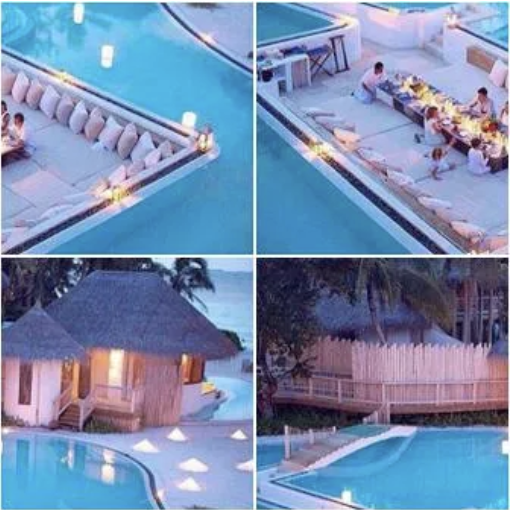}\utarrow{}\utimgwi{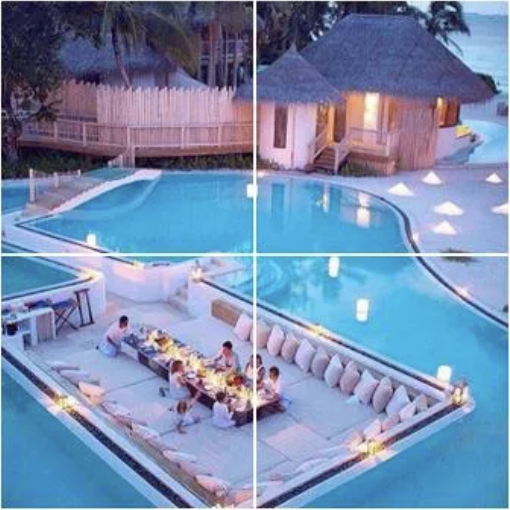}\hspace{6pt}%
\utimgwi{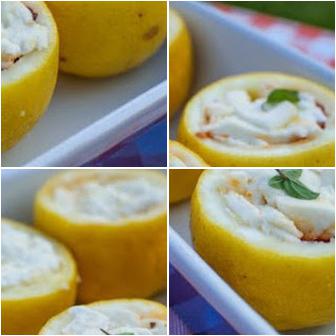}\utarrow{}\utimgwi{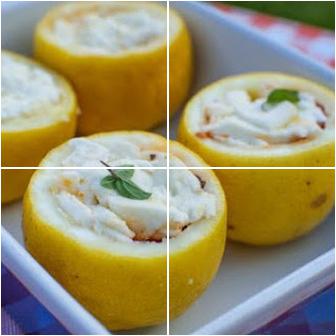}\par
\end{minipage}%
}{%
\uticell{29.9pt}{Localization}{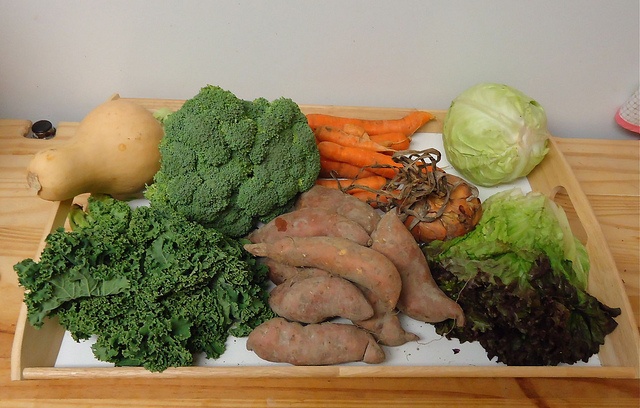}{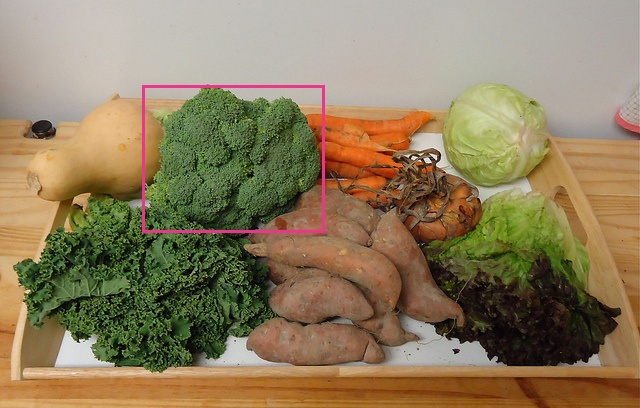}{}{\textbf{Instruction:} Locate the object referred to by the text description and mark it with a single clean bounding box. Referring expression: green color vegetable in between potato and carrot}%
}
\utrowseparator{.55pt}
\uttriple{%
\utcell{67.0pt}{Localization}{}{Which box best encloses the flower arrangement?}{B\,---\, Box B}%
}{%
\utcell{67.0pt}{Counting}{}{How many slices of bread are shown?}{B\,---\, There are 14 slices of bread in this image.}%
}{%
\utcell{67.0pt}{Connectivity}{}{What is the relation between the cat and the vase?}{B\,---\, The cat is inside the vase.}%
}
\utrowseparator{.3pt}
\utpairwide{%
\begin{minipage}[t]{.645\linewidth}\vspace{0pt}\raggedright
{\sffamily\bfseries\fontsize{7.4}{9}\selectfont 2D keypoints\hspace{1pt}\uttag{I2T}\par}\vspace{4pt}\nointerlineskip
\begin{minipage}[t]{\utW}\vspace{0pt}\utimgw{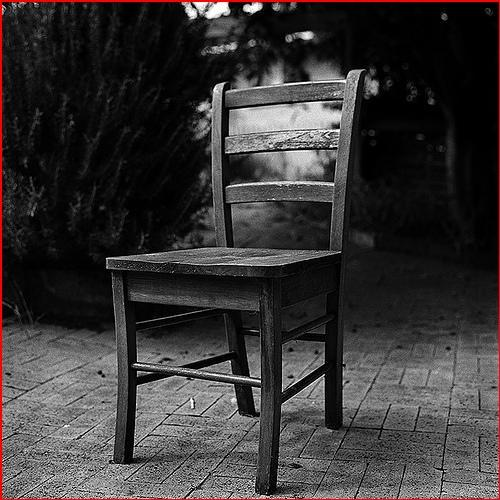}\end{minipage}\hspace{8pt}%
\begin{minipage}[t]{\dimexpr\linewidth-\utW-8pt\relax}\vspace{0pt}\raggedright
\textbf{Q:} Detect the visible furniture keypoints.\par\vspace{1pt}
{\color{utanswer}\textbf{Ref:} C; normalized $(x,y)$:\par}\vspace{2pt}
{\fontsize{6.6}{7.4}\selectfont\setlength{\tabcolsep}{2.5pt}\renewcommand{\arraystretch}{.95}%
\begin{tabular}{@{}lr@{\hspace{8pt}}lr@{}}
Legs RT & [0.244, 0.924] & Base LB & [0.674, 0.498]\\
Legs LT & [0.532, 0.978] & Base RB & [0.438, 0.484]\\
Legs LB & [0.668, 0.840] & Head RT & [0.436, 0.168]\\
Legs RB & [0.490, 0.816] & Head LT & [0.710, 0.142]\\
Base RT & [0.212, 0.518] & Base LT & [0.540, 0.540]\\
\end{tabular}\par}\vspace{2pt}
{\fontsize{6.3}{7.2}\selectfont\color{utmuted}R/L: right/left; T/B: top/bottom; Head: headboard.\par}
\end{minipage}
\end{minipage}%
}{%
\utcell{37.0pt}{2D ordering}{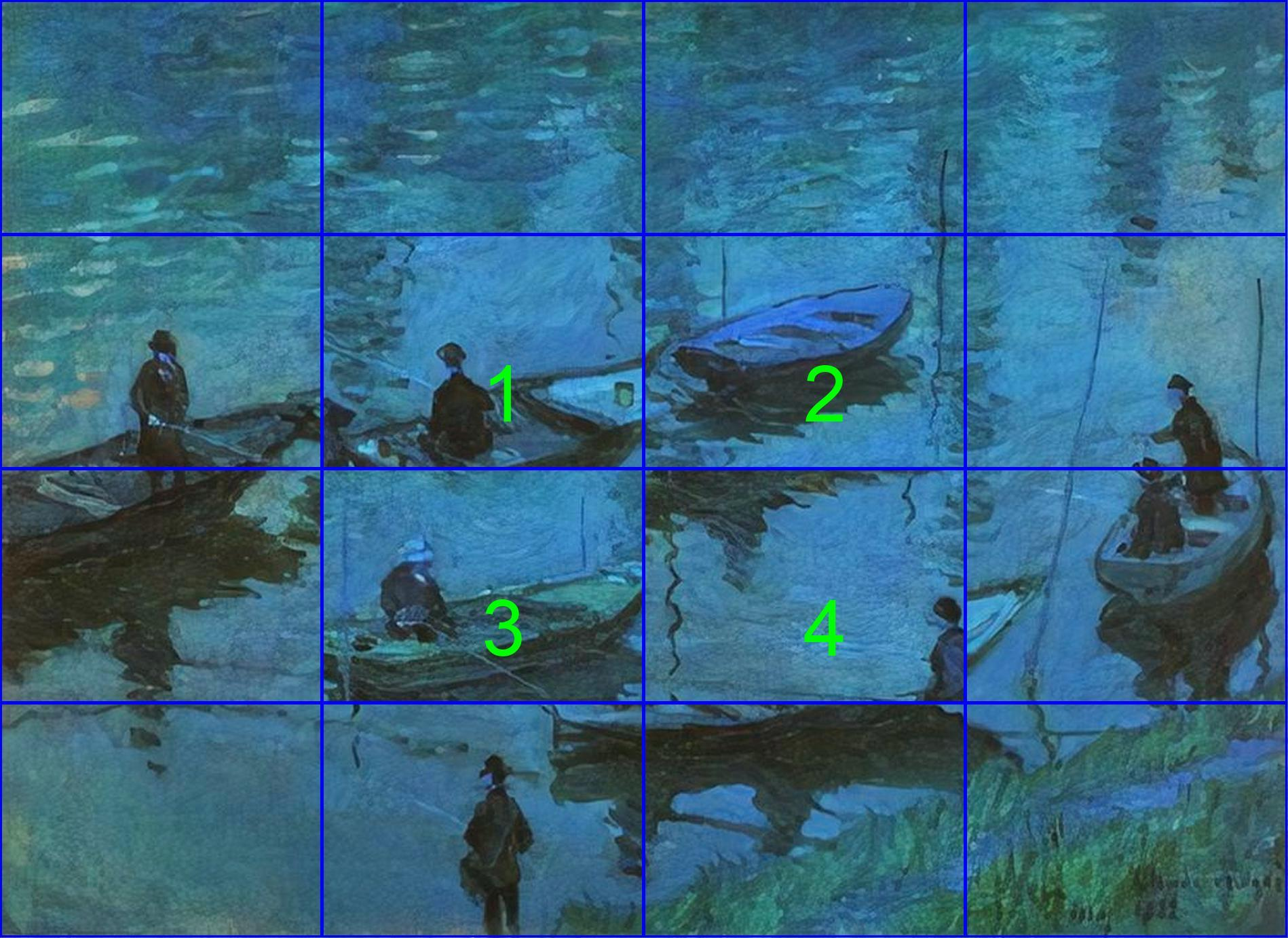}{Recover the patch order: top left, top right, bottom left, bottom right.}{C\,---\, [3, 2, 4, 1]}%
}
\utrowseparator{.3pt}
\begin{minipage}[t]{\linewidth}\vspace{0pt}\raggedright
{\sffamily\bfseries\fontsize{8.1}{9.5}\selectfont Visual correspondence \uttag{I2T}\par}\vspace{4pt}\nointerlineskip
\noindent\utmultiimg{}{REF}\hspace{6pt}%
\utmultiimg{}{Candidate points}\hspace{14pt}%
\begin{minipage}[t]{\dimexpr\linewidth-2\utW-20pt\relax}\vspace{0pt}\raggedright\textbf{Q:} Which point in the second image corresponds to REF?\par\vspace{2pt}{\color{utanswer}\textbf{Ref:} A\,---\, Point A\par}\end{minipage}
\end{minipage}\par
\end{tcolorbox}
\clearpage
\endgroup

%% file: iclr2026/sections/appendix/5_transfer.tex
\section{Complete transfer results}
\label{app:transfer}
\subsection{Training setup}
\label{app:transfer-setup}
Each source checkpoint follows the I2I $\rightarrow$ I2T recipe. We first finetune pretrained BAGEL on one I2I task, then initialize the I2T stage from the resulting weights and train on 50,000 LLaVA-Instruct examples. The optimizer is reinitialized between the two stages. The I2T-only baseline starts from the same pretrained BAGEL checkpoint and uses identical I2T data and optimization settings (Table~\ref{tab:transfer-training}). The language model, understanding components, and ViT are trainable, while the VAE is frozen during I2I training and unused in the I2T stage. For each training, we use three different random seeds and average the evaluation results.

\begin{table}[!htbp]
\centering\small
\setlength{\tabcolsep}{8pt}
\caption{\textbf{Transfer training settings.}}
\label{tab:transfer-training}
\begin{tabular}{@{}lcc@{}}
\toprule
Setting & Stage 1 (I2I) & Stage 2 (I2T) \\
\midrule
Training-example budget & 50,000 & 50,000 \\
Batch composition & I2I only & I2T only \\
GPUs / effective batch size & 4 / 64 & 4 / 64 \\
Optimizer updates & 781--782 & 781 \\
Optimizer & \multicolumn{2}{c}{AdamW, $\beta=(0.9,0.95)$} \\
Learning rate & \multicolumn{2}{c}{$2\times10^{-5}$} \\
Warmup & 50 updates & 50 updates \\
Schedule after warmup & Constant & Constant \\
Conditioning dropout (text, VAE, ViT) & $(0,0,0)$ & $(0,0,0)$ \\
Weight decay / gradient-norm limit & 0 / 1.0 & 0 / 1.0 \\
\bottomrule
\end{tabular}
\end{table}

Both stages use an effective batch of 64 examples on four GPUs. Each I2I stage uses either 16 examples per GPU without gradient accumulation or four examples per GPU with four accumulation steps, and all I2T stages use the latter configuration. With the former, the 50k budget ends at 50,048 examples after 782 updates; with the latter, it ends after 781 updates.

\input{iclr2026/sections/appendix/generated/capability_baselines}

\subsection{Full task-by-capability matrix}
\label{app:transfer-full}
\begin{figure}[p]
\centering
\includegraphics[width=\linewidth,height=.85\textheight,keepaspectratio]{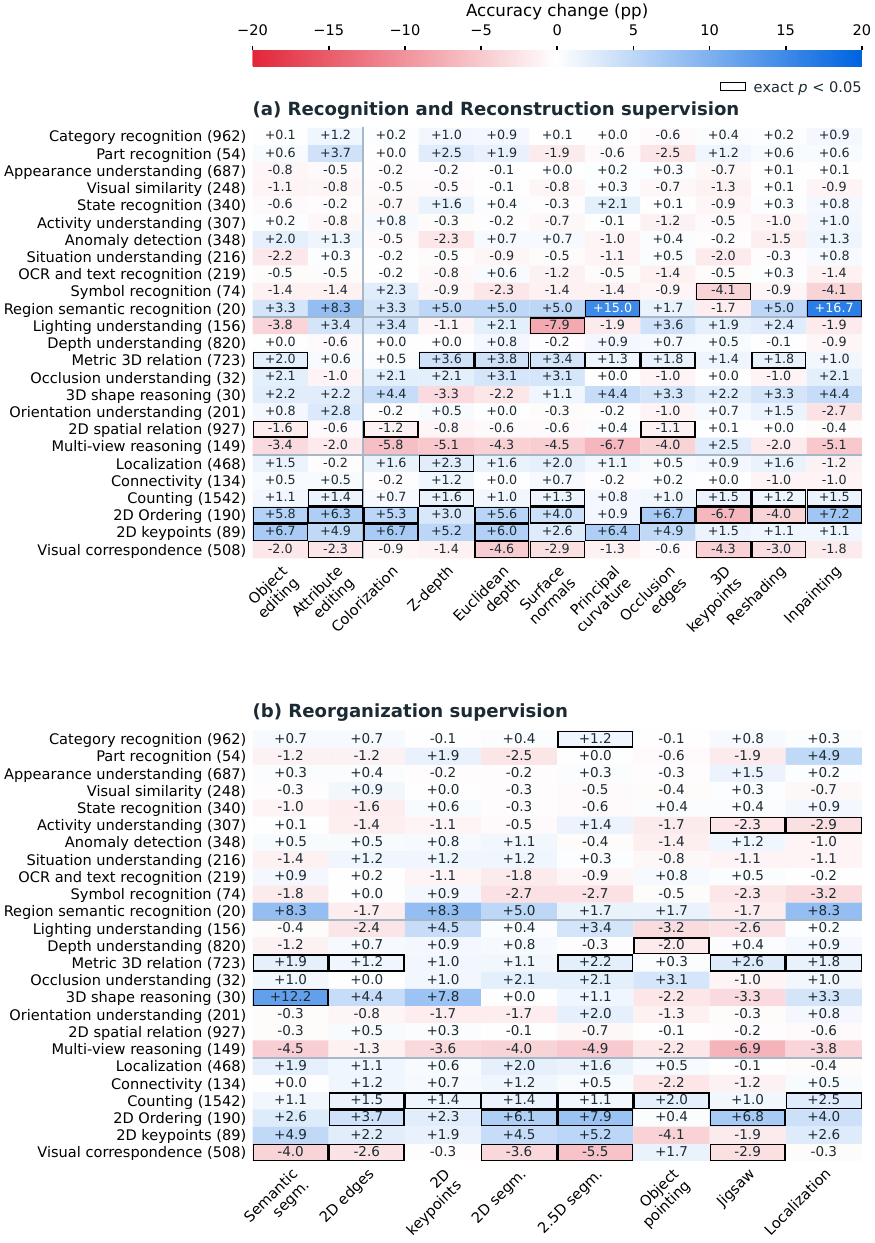}
\caption{\textbf{Complete transfer matrix.} Panel (a) shows the Recognition (2) and Reconstruction (9) sources, and panel (b) the Reorganization (8) sources, including all 25 target capabilities. Values are accuracy changes over the I2T-only baseline in percentage points. Outlined borders mark exact $p<0.05$ from paired permutation tests (Appendix~\ref{app:transfer-tests}). Parentheses give the number of evaluation examples.}
\label{fig:app-transfer-full}
\end{figure}

Table~\ref{tab:app-baselines} lists the baseline accuracy on each capability. And Figure~\ref{fig:app-transfer-full} reports all $19\times25=475$ source--target pairs. The main-text map (Figure~\ref{fig:transfer}) shows only the 19 capabilities with more than 100 evaluation examples. The remaining six are part recognition (54), symbol recognition (74), region semantic recognition (20), occlusion understanding (32), 3D shape reasoning (30), and 2D keypoints (89). 

\subsection{Paired permutation tests of transfer gains}
\label{app:transfer-tests}
We compare each source with the I2T-only baseline on the same $N$ questions across three training seeds. Let $a_{i,s},b_{i,s}\in\{0,1\}$ denote their correctness on question $i$ under seed $s$. We define
\[
d_i=\frac{1}{3}\sum_{s=1}^{3}(a_{i,s}-b_{i,s}),
\qquad
T_{\mathrm{obs}}=\frac{1}{N}\sum_{i=1}^{N}d_i,
\]
and report $100T_{\mathrm{obs}}$ as the gain in percentage points.

Under the paired permutation null, we independently swap source and baseline outcomes for each question with probability $1/2$, applying the same swap across all three seeds. The exact two-sided $p$ value is
\[
p_{\mathrm{exact}}
=2^{-N}\sum_{\boldsymbol{\epsilon}\in\{-1,+1\}^{N}}
\mathbf{1}\!\left\{
\left|\frac{1}{N}\sum_{i=1}^{N}\epsilon_i d_i\right|
\geq |T_{\mathrm{obs}}|
\right\}.
\]
We compute one $p$ value per source--target pair and use the significance threshold of $p<0.05$.

%% file: iclr2026/sections/appendix/generated/capability_baselines.tex
\begin{table}[tb]
\centering
\footnotesize
\setlength{\tabcolsep}{4pt}
\caption{\textbf{I2T-only baseline accuracy on each understanding capability, averaged over three random seeds.}}
\label{tab:app-baselines}
\begin{tabular}{@{}lrlr@{}}
\toprule
Capability & Base (\%) & Capability & Base (\%) \\
\midrule
Category recognition & 75.09 & Metric 3D & 68.65 \\
Part recognition & 74.07 & Occlusion & 78.12 \\
Appearance & 66.13 & 3D shape reasoning & 31.11 \\
Visual similarity & 77.96 & Orientation & 55.72 \\
State recognition & 68.73 & 2D spatial & 88.21 \\
Activity & 72.20 & Multi-view & 50.56 \\
Anomaly detection & 51.44 & Localization & 61.89 \\
Situation & 68.52 & Connectivity & 85.32 \\
OCR / text & 87.37 & Counting & 63.58 \\
Symbol recognition & 80.18 & 2D ordering & 60.35 \\
Region semantics & 46.67 & 2D keypoints & 45.32 \\
Lighting & 45.51 & Correspondence & 50.72 \\
Depth & 84.55 &  &  \\
\bottomrule
\end{tabular}
\end{table}

%% file: iclr2026/sections/appendix/6_gradient.tex
\section{Gradient alignment: Methods and additional results}
\label{app:gradient_alignment}
\subsection{Gradient collection}
\label{app:gradient-setup}
We compute all gradients in Figure~\ref{fig:grad_analysis} at the pretrained BAGEL EMA checkpoint, before any task-specific finetuning. The reference set contains six paired tasks: Jigsaw, Zoom-In, Colorization, Video-Unshuffle, Counting, and Image-Rotation, with 500 source pairs per task. Each pair consists of an I2I example and an I2T example derived from the same source, although their rendered inputs need not be identical.

For each example and objective, we run a separate forward and backward pass without updating the model. We differentiate through the ViT, the connector, and the selected understanding-expert parameters of the LLM; weights used only by the generation expert are not compared. The VAE remains frozen, and all conditioning dropout is disabled during gradient measurement. The generation loss of each source uses a reproducible draw of the timestep and noise. The forward pass uses bfloat16 model weights with autocast. We cast the extracted gradients to float32 for storage and compute Gram matrices and PCA in float64.

To separate fitting from scoring, we assign the source pairs to five balanced folds of 100 pairs per task, keeping the I2I and I2T members of each pair together. Source IDs are shuffled by a seeded hash and assigned to folds in round-robin order. For each held-out fold, the other four folds provide $6\times2\times400=4{,}800$ reference gradients. Only these reference gradients determine the PCA basis and effective dimension, and the held-out pairs are scored afterward. Each of the 500 pairs is therefore scored once, with a basis fitted without it. These folds concern only the fitting of the gradient representation and are unrelated to model finetuning. Appendices~\ref{app:gradient-minibatch} and~\ref{app:gradient-training-norms} describe the cross-task analysis and the measurements at I2I-only checkpoints.

\subsection{Loss definitions and gradient dimensions}
\label{app:gradient-statistics}
\paragraph{Loss definitions.}
For I2I, we use the flow-matching velocity loss. With $N_g$ supervised image tokens and $C$ latent coordinates per token, the loss is
\[
\mathcal L_{\mathrm{I2I}}(x)
=\frac{1}{N_g C}\sum_{j=1}^{N_g}\sum_{c=1}^{C}
\left[v_{\boldsymbol\theta}(z_t,t,\mathrm{cond})_{jc}-(\epsilon-z)_{jc}\right]^2,
\]
where $z$ is the target image latent and $\epsilon$ is the sampled noise. Squared errors are averaged first over latent coordinates and then over valid target-image tokens. For I2T, the loss is the teacher-forced cross-entropy averaged over the $N_t$ supervised output tokens:
\[
\mathcal L_{\mathrm{I2T}}(x)
=-\frac{1}{N_t}\sum_{j=1}^{N_t}
\log p_{\boldsymbol\theta}(y_j\mid y_{<j},\mathrm{cond}).
\]
The loss mask excludes input images, questions, and other unsupervised context positions. We compute the two losses separately, without a mixing coefficient. Because their absolute scales are not comparable, we compare losses across checkpoints only within each objective.

\paragraph{Gradient dimensions.}
Each coordinate of a raw gradient is the derivative of the loss with respect to one scalar model parameter. To reduce storage for large tensors, we apply a fixed signed CountSketch~\citep{clarkson2017low}: each stored coordinate sums, with fixed signs, the derivatives assigned to one bucket. Large ViT tensors are stored in 8,192 coordinates and other large tensors in 16,384. Small tensors and all normalization weights keep their original coordinates. The same tensor-specific map is reused across tasks, objectives, and examples.

We consider two kinds of parameter blocks. A concatenated block joins one parameter type across all layers \emph{after} any tensor-wise sketching and is normalized once as a whole, not layer by layer. A layer-wise block is a single tensor. After PCA, each coordinate is a coefficient along a fitted eigenvector rather than a derivative with respect to an individual weight. Table~\ref{tab:app-gradient-dimensions} reports the resulting dimensions. In particular, pre-attention RMSNorm has 3,584 coordinates per layer and $28\times3{,}584=100{,}352$ concatenated coordinates, all kept without sketching.

\input{iclr2026/sections/appendix/generated/gradient_dimensions}

\subsection{PCA and effective dimension}
\label{app:gradient-blocks}
\label{app:gradient-method}
For parameter block $b$, let $\mathbf u_{b,i}$ be the unit-normalized stored gradient of example $i$. In fold $f$, we fit PCA to the uncentered second moment of the reference directions:
\[
\mathbf S_{b,f}=\sum_{i\in\mathcal I_{-f}}w_i\mathbf u_{b,i}\mathbf u_{b,i}^{\top}.
\]
Each of the 12 task--objective groups receives a total weight of $1/12$, spread uniformly over its 400 examples, and we do not subtract the mean direction. The basis $\mathbf P_{b,f}$ spans the smallest leading eigenspace that accounts for at least 99\% of the trace of this second moment, including eigenvalues tied at the cutoff. We then project and renormalize both reference and held-out directions:
\[
\mathbf v_{b,i,f}=\frac{\mathbf P_{b,f}^{\top}\mathbf u_{b,i}}
{\|\mathbf P_{b,f}^{\top}\mathbf u_{b,i}\|_2}.
\]
The two objectives are therefore compared in the same subspace. The 99\% criterion is defined on the reference gradients; for concatenated pre-attention RMSNorm, the mean retained energy of held-out gradients is 98.67--98.71\% across folds.

The retained subspace can be highly anisotropic. We therefore compute an effective dimension from the projected and normalized \emph{reference} directions:
\[
\mathbf A_{b,f}=\sum_{i\in\mathcal I_{-f}}w_i\mathbf v_{b,i,f}\mathbf v_{b,i,f}^{\top},
\qquad
d_{\mathrm{eff},b,f}=\frac{\operatorname{tr}(\mathbf A_{b,f})^2}{\operatorname{tr}(\mathbf A_{b,f}^2)}.
\]
Figure~\ref{fig:grad_analysis}(a,b) reports each held-out paired cosine multiplied by $\sqrt{d_{\mathrm{eff},b,f}}$, averaged over the 500 source pairs.
For Figure~\ref{fig:grad_analysis}(a), we compute the scaled score separately for each parameter type and then aggregate within modules. The ViT bar is the equally weighted mean over its 11 parameter types. Conn.~in and Conn.~out average the weight and bias scores of the first and second connector linear layers, respectively.

\subsection{Cross-task alignment and transfer}
\label{app:gradient-minibatch}
\label{app:gradient-correlation}
This analysis uses the 19 original source checkpoints, for which both gradient and transfer measurements are available. All gradients are measured at the base BAGEL checkpoint on the pre-attention RMSNorm of the understanding branch, concatenated across all 28 transformer layers.

We analyze the seven understanding capabilities with more than 500 evaluation examples (Table~\ref{tab:app-gradient-seven}). For each capability, we select 500 questions from the retained evaluation pool in seeded hash order and compute teacher-forced gradients on the correct option label plus EOS (two supervised tokens). Each I2I source task also contributes 500 gradients. We divide the records into five folds, keeping understanding examples with an identical decoded image in the same fold. We reuse the six-task PCA bases from Appendix~\ref{app:gradient-method} without refitting them on these new pools.

Within each held-out fold, we independently sample I2I and I2T minibatches of size 64 with replacement, using 2,000 paired draws per fold and thus 10,000 draws per source--target cell. Each task or capability keeps the same folds and sampling stream across comparisons.

For each minibatch, we average the raw per-example gradients weighted by their valid loss-token counts, then project and normalize the result:
\[
\overline{\mathbf g}_{\mathcal B}
=\frac{\sum_{i\in\mathcal B}\ell_i\mathbf g_i}{\sum_{i\in\mathcal B}\ell_i},
\qquad
\mathbf v_{\mathcal B,f}
=\frac{\mathbf P_f^\top\overline{\mathbf g}_{\mathcal B}}
{\|\mathbf P_f^\top\overline{\mathbf g}_{\mathcal B}\|_2},
\]
where $\ell_i$ is the valid loss-token count and $\mathbf P_f$ is the shared PCA basis for fold $f$. The alignment $G_{s,t}$ is the mean cosine between I2I and I2T minibatch directions, averaged over draws and then over the five equally sized target folds. It is thus an average of minibatch cosines rather than the cosine between two full-pool gradients.

The transfer gain $\Delta_{s,t}$ is the accuracy on target $t$ after I2I training on source $s$ followed by I2T finetuning, minus the I2T-only baseline accuracy, in percentage points. Figure~\ref{fig:grad_analysis}(c) averages $G_{s,t}$ and $\Delta_{s,t}$ over the same sources for each target, and panel (d) plots the individual $(G_{s,t},\Delta_{s,t})$ pairs. The Pearson correlation is $r=0.795$ across the seven capability means and $r=0.529$ across all 133 source--target pairs.

\begin{table}[tb]
\centering
\small
\setlength{\tabcolsep}{6pt}
\caption{\textbf{Understanding capabilities in the cross-task gradient analysis.} Each capability has more than 500 evaluation examples, of which 500 are used for gradient sampling. Alignment and transfer gain are averaged over the 19 I2I sources, as in Figure~\ref{fig:grad_analysis}(c).}
\label{tab:app-gradient-seven}
\begin{tabular}{@{}lrrr@{}}
\toprule
Understanding capability & Evaluation examples & Mean alignment & Mean gain (pp) \\
\midrule
Category recognition & 962 & 0.108 & $+0.44$ \\
Appearance understanding & 687 & 0.019 & $+0.01$ \\
Depth understanding & 820 & $-0.038$ & $+0.07$ \\
Metric 3D relation & 723 & $-0.005$ & $+1.75$ \\
2D spatial relation & 927 & $-0.001$ & $-0.40$ \\
Counting & 1,542 & 0.186 & $+1.32$ \\
Visual correspondence & 508 & $-0.246$ & $-2.25$ \\
\bottomrule
\end{tabular}
\end{table}

\subsection{Losses and gradient norms after I2I-only training}
\label{app:gradient-training-norms}
Table~\ref{tab:app-training-gradient-norms} reports the losses and gradient norms of both objectives at the BAGEL base checkpoint and after I2I-only training with budgets of 3k, 10k, and 30k examples. Each budget is a separate run from the same pretrained model, not an intermediate checkpoint of a single run, and none of these checkpoints has undergone I2T finetuning.

At every checkpoint, we evaluate the same 500 matched I2I--I2T pairs per task, with conditioning dropout disabled and the same source-specific noise and timestep draws. 
For each objective $o\in\{\mathrm{I2I},\mathrm{I2T}\}$, we report the mean per-example loss and the mean per-example gradient norm:
\[
\overline{\mathcal L}_o=\frac{1}{500}\sum_{i=1}^{500}\mathcal L_o(x_i),
\qquad
M_o=\frac{1}{500}\sum_{i=1}^{500}
\left\|\nabla_{\boldsymbol\theta_{\mathrm{in}}}\mathcal L_o(x_i)\right\|_2,
\]
where $\mathcal L_o$ uses the reductions defined in Appendix~\ref{app:gradient-statistics}, and $\boldsymbol\theta_{\mathrm{in}}$ concatenates the understanding-branch pre-attention RMSNorm weights from all 28 LLM layers (100,352 parameters). We compute norms in the original parameter coordinates, without unit normalization, PCA projection, or dimension scaling. Thus, $M_o$ averages per-example norms rather than taking the norm of an averaged gradient.

\input{iclr2026/sections/appendix/generated/training_loss_norms}

Loss and gradient norm follow broadly similar trends. After I2I training, the I2I loss and gradient norm drop sharply on both tasks. At 30k, the I2T loss also falls, from 0.43 to 0.28 on Jigsaw and from 0.29 to 0.22 on Zoom-In, and the I2T gradient norm falls from 8.89 to 2.18 and from 4.56 to 0.74, respectively. Because these checkpoints receive no I2T supervision, the lower I2T loss provides complementary evidence that I2I training can benefit the paired I2T tasks.

%% file: iclr2026/sections/appendix/generated/gradient_dimensions.tex
\begin{table}[tb]
\centering
\footnotesize
\setlength{\tabcolsep}{6pt}
\newsavebox{\appGradientDimensionsBox}
\sbox{\appGradientDimensionsBox}{%
\begin{tabular}{@{}lrrr@{}}
\toprule
Parameter block & Original $D$ & Stored $D$ & PCA $k$ \\
\midrule
ViT patch embedding & 677,376 & 8,192 & 3,048--3,058 \\
ViT position embedding & 5,644,800 & 8,192 & 3,931--3,936 \\
ViT layer norms 1/2 & 29,952 & 29,952 & 2,505--2,586 \\
ViT attention Q/K/V/O & 34,504,704 & 212,992 & 4,344--4,442 \\
ViT MLP layers 1/2 & 128,913,408 & 212,992 & 4,213--4,505 \\
ViT final layer norm & 1,152 & 1,152 & 423--432 \\
\midrule
Connector layer 1 weight & 4,128,768 & 16,384 & 3,839--3,841 \\
Connector layer 2 weight & 12,845,056 & 16,384 & 3,775--3,778 \\
Connector layer 1/2 biases & 3,584 & 3,584 & 1,866--1,874 \\
Connector position embedding & 17,561,600 & 16,384 & 4,041--4,045 \\
\midrule
LLM pre-attention norm & 100,352 & 100,352 & 186--190 \\
LLM post-attention norm & 100,352 & 100,352 & 722--733 \\
LLM query norm & 3,584 & 3,584 & 267--270 \\
LLM key norm & 3,584 & 3,584 & 238--240 \\
LLM attention Q/O & 359,661,568 & 458,752 & 2,010--2,429 \\
LLM attention K/V & 51,380,224 & 458,752 & 1,714--2,182 \\
LLM MLP gate/up/down & 1,901,068,288 & 458,752 & 1,555--1,795 \\
LLM token embedding & 544,997,376 & 16,384 & 818--826 \\
LLM final norm & 3,584 & 3,584 & undefined \\
\bottomrule
\end{tabular}
}
\begin{minipage}{\wd\appGradientDimensionsBox}
\centering
\caption{Dimensions per concatenated parameter type. PCA ranges span grouped types and folds.}
\label{tab:app-gradient-dimensions}
\usebox{\appGradientDimensionsBox}
\end{minipage}
\end{table}

%% file: iclr2026/sections/appendix/generated/training_loss_norms.tex
\begin{table}[tb]
\centering
\footnotesize
\setlength{\tabcolsep}{5pt}
\newsavebox{\appTrainingLossNormsBox}
\sbox{\appTrainingLossNormsBox}{%
\begin{tabular}{@{}lrrrrrrrr@{}}
\toprule
& \multicolumn{4}{c}{Jigsaw} & \multicolumn{4}{c}{Zoom-In} \\
\cmidrule(lr){2-5}\cmidrule(l){6-9}
& \multicolumn{2}{c}{I2I} & \multicolumn{2}{c}{I2T}
& \multicolumn{2}{c}{I2I} & \multicolumn{2}{c}{I2T} \\
\cmidrule(lr){2-3}\cmidrule(lr){4-5}\cmidrule(lr){6-7}\cmidrule(l){8-9}
Budget & $\overline{\mathcal L}$ & $M$ & $\overline{\mathcal L}$ & $M$
& $\overline{\mathcal L}$ & $M$ & $\overline{\mathcal L}$ & $M$ \\
\midrule
BAGEL base & 1.09 & 16.95 & 0.43 & 8.89 & 1.29 & 14.36 & 0.29 & 4.56 \\
3k & 0.13 & 0.04 & 0.44 & 7.77 & 0.15 & 0.05 & 0.26 & 4.47 \\
10k & 0.12 & 0.03 & 0.31 & 5.09 & 0.14 & 0.03 & 0.32 & 5.99 \\
30k & 0.12 & 0.01 & 0.28 & 2.18 & 0.14 & 0.02 & 0.22 & 0.74 \\
\bottomrule
\end{tabular}
}
\begin{minipage}{\wd\appTrainingLossNormsBox}
\centering
\caption{Losses and gradient norms after I2I-only training.}
\label{tab:app-training-gradient-norms}
\usebox{\appTrainingLossNormsBox}
\end{minipage}
\end{table}

%% file: iclr2026/sections/appendix/7_prompts.tex
\clearpage
\section{Prompts}
\label{app:prompts}

This appendix collects the model instructions used to construct \methodname, annotate benchmark examples, judge Jigsaw and Zoom-In generations, and route Recognition editing data. Braced variables are filled with the corresponding examples, schemas, or capability catalogues, and images are supplied separately. Figure~\ref{fig:prompt-worked-example} illustrates the schema stages on a recorded example.

\input{iclr2026/sections/appendix/7_construction_prompts}
\input{iclr2026/sections/appendix/7_annotation_prompts}
\clearpage
\input{iclr2026/sections/appendix/7_evaluation_prompts}
\input{iclr2026/sections/appendix/7_i2i_data_prompts}

%% file: iclr2026/sections/appendix/7_construction_prompts.tex
\subsection{Taxonomy construction}
\label{app:prompt-construction}
\label{app:prompt-construction-models}

Table~\ref{tab:taxonomy-construction-models} lists the model settings for each construction step. Flash and Pro denote the requested Gemini API identifiers \texttt{gemini-3-flash-preview} and \texttt{gemini-3.1-pro-preview}, and GPT denotes \texttt{gpt-5.6-sol} accessed through the Codex CLI. These calls are separate from the three-model annotation of the final benchmark examples (Appendix~\ref{app:prompt-annotation}). We parse Gemini outputs as JSON, and the 3R call uses an explicit JSON output schema. For the CLI call, we do not set the temperature or an output-token cap.

\begin{table}[htbp]
\centering\small
\caption{\textbf{Taxonomy-construction model settings.}}
\label{tab:taxonomy-construction-models}
\begin{tabularx}{\linewidth}{@{}X l r r l@{}}
\toprule
Step & Model & Temp. & Output cap & Thinking \\
\midrule
Free-form description & Flash & 0.2 & 12,000 & 0 \\
Schema induction & Flash & 0.2 & 16,384 & 4,096 \\
Fixed-schema description & Flash & 0.2 & 12,000 & 0 \\
Local tree induction & Pro & 0.1 & 65,536 & 8,192 \\
Single-leaf assignment & Pro & 0.1 & 2,048 & 384 \\
Global tree merging & Pro & 0.1 & 65,536 & 8,192 \\
Leaf-to-3R assignment & GPT & --- & --- & max effort \\
\bottomrule
\end{tabularx}
\end{table}

Output caps include thinking tokens, and the thinking column gives the requested token budget or CLI reasoning effort. Failed single-leaf assignments are retried with temperatures 0.2, 0.3, and 0.4, an output cap of 3,072 tokens, and a thinking budget of 1,536 tokens.

Figures~\ref{fig:taxonomy-freeform}--\ref{fig:taxonomy-3r} show the task instructions and output requirements of the user-prompt templates. Images are attached separately for free-form description, fixed-schema description, and single-leaf assignment, and question fields include the answer choices. The Gemini system instruction is ``You output only strict JSON. No markdown, no commentary.''; single-leaf assignment uses the shorter ``You output only strict JSON. No markdown.'' The 3R template is passed through the CLI together with its output schema, without a custom system instruction.

The 3R prompt (Figure~\ref{fig:taxonomy-3r}) takes as input the 35-leaf catalogue from the first review round (Appendix~\ref{app:construction-trees}). The templates below keep placeholders for sample-dependent inputs; Figure~\ref{fig:prompt-worked-example} shows the fixed schema and a recorded example.

\begingroup
\lstset{style=paperprompt,basicstyle=\fontsize{8}{9.5}\selectfont\ttfamily}

\clearpage
\begin{promptfigurepages}{Free-form visual and task description}{fig:taxonomy-freeform}
\begin{lstlisting}
TASK: Extract a RICH, EXHAUSTIVE set of **50–80 discriminating attributes** for ONE item from a multimodal VLM benchmark. These attributes will build a fine-grained taxonomy covering perceptual ability, IMAGE TYPE, DIFFICULTY, and DATA QUALITY.

You see ONE item: image(s) + a question + the gold answer. **LOOK CAREFULLY at the image(s).**

Emit a FLAT JSON object of **50–80** `key: value` pairs. Choose your own snake_case keys/values (free-form; a later step merges synonyms). Values are snake_case strings, integers, booleans, or short lists of snake_case strings. No sentences, no nested objects.

Be **EXHAUSTIVE and ORTHOGONAL**: cover every GROUP below with MULTIPLE attributes. Crucially, things like "image type" or "difficulty" are NOT a single attribute — they must EMERGE from the COMBINATION of many orthogonal attributes you provide.

GROUP 1 — VISUAL MEDIUM & SOURCE (give ~12–18 attributes so different image KINDS are separable by their combination — e.g. real photo vs 3D render vs AI-generated vs medical scan vs satellite vs webpage/app screenshot vs scanned document vs textbook-style figure vs chart/diagram vs artwork):
  - capture_method (real_photo / 3d_render / ai_generated / screenshot / scanned_document / hand_drawing / synthetic_composite / rendered_diagram …)
  - image_source / medium (natural_photo / medical_scan / microscopy / satellite_aerial / webpage_screenshot / app_ui / document_page / chart_or_plot / scientific_diagram / map / artwork / textbook_figure …)
  - is_educational_figure (true/false) — a math/science/reading problem rendered as a FIGURE (e.g. geometry figure, plotted function, illustrated word-problem), NOT a photo of a physical textbook — and figure_subject (math / geometry / physics / chemistry / biology / reading_text / none)
  - scene_setting (indoor / outdoor / studio / mixed / not_applicable) + indoor_type / outdoor_type if applicable
  - color_mode (color / grayscale / binary), photorealism (high / medium / low / non_photographic), has_text_overlay, has_annotations_or_markers, has_ui_chrome, is_collage_or_multi_panel, background_type
  Provide enough Group-1 attributes that "generated / 3d-render / medical / outdoor / webpage / indoor / textbook-figure …" are each distinguishable by the SET of values.

GROUP 2 — SCENE CONTENT & OBJECTS:
  primary_subject, object_categories_present (short list), num_distinct_object_types, dominant_object, num_salient_entities, scene_clutter (clean / moderate / cluttered / crowded), background_complexity, spatial_layout, contains_people, contains_text.

GROUP 3 — PERCEPTUAL OPERATION (the core ability tested):
  main_task (counting / depth_ordering / spatial_relation / fine_grained_recognition / attribute_recognition / localization / ocr / comparison / anomaly_detection / action_recognition / …), perceptual_sub_mechanism, why_visual_bottleneck (why the answer needs the image, not just text/world-knowledge).

GROUP 4 — TASK-SPECIFIC DIFFICULTY AXES (add 4–8 attributes that separate EASY from HARD instances of WHATEVER task this is). Examples by task — generalize to the actual task:
  - counting: count_range (small_le5 / medium_6_20 / large_21_50 / very_large_50plus), target_type (objects / people / vehicles / text_glyphs / cells / animals …), target_size (tiny / small / medium / large), scene_density (sparse / moderate / dense / crowded), occlusion_level, instances_homogeneous, counting_setting (tabletop / street / stage_or_arena / document / microscope …)
  - spatial / depth: relation_type, reference_frame (egocentric / allocentric), depth_cue_reliance, num_relata
  - recognition / attribute: granularity (coarse / fine_grained / part_level), distractor_similarity, attribute_type (color / material / shape / texture / state / emotion …)
  - ocr: text_amount, text_size (large / small / tiny), text_clarity, text_layout (single_line / paragraph / scattered / tabular)
  (counting is only an EXAMPLE; every task has analogous easy↔hard axes — emit the ones that fit this item.)

GROUP 5 — QUESTION & ANSWER PROPERTIES:
  answer_format (multiple_choice / binary / numeric / open_text / coordinates …), num_options, answer_type, requires_fine_detail (true/false), requires_multi_step_reasoning (true/false), reasoning_steps_estimate (int), requires_external_knowledge (true/false), question_phrasing_complexity, answer_recoverable_from_text_alone (true/false).

GROUP 6 — DIFFICULTY & DATA-QUALITY JUDGMENT (your own assessment of this item):
  estimated_difficulty (trivial / easy / medium / hard / very_hard), difficulty_drivers (short list, or none), solvable_by_coarse_glance (true/false),
  question_well_posed (true/false — unambiguous and answerable from the image?), answer_looks_correct (true / false / uncertain — does the gold answer plausibly match the image?),
  data_quality_issue (none / ambiguous_question / multiple_valid_answers / answer_seems_wrong / question_answer_mismatch / low_image_quality / text_unreadable / other),
  image_quality (high / standard / low / noisy / blurred), ambiguity_level (none / mild / high).

RULES:
- 50–80 attributes total. DISCRIMINATING (avoid values nearly every item shares) and INFORMATIVE (never vacuous: not "image", "object", "task", "yes").
- Ground visual attributes in what the IMAGE actually shows; ground difficulty/quality in image + question + gold answer together.
- Do NOT use any benchmark / dataset / framework name as a key or value.
- Return ONLY the JSON object — no prose, no markdown fences.

QUESTION:
{question}

GOLD_ANSWER: {answer}
\end{lstlisting}
\end{promptfigurepages}

\par\bigskip
\begin{promptfigurepages}{Inducing the fixed attribute schema}{fig:taxonomy-schema}
\begin{lstlisting}
TASK: From the free-form attributes of {n} samples, induce a FIXED attribute schema that will label ALL samples uniformly.

Below are the OBSERVED keys (with how many samples used each) and, per key, the most common values (with counts). Keys/values are heterogeneous: different samples used different names for the same property.

Produce a FIXED schema — a curated set of CANONICAL KEYS:
- SYNONYM RESOLUTION (be CONSERVATIVE — merge ONLY near-identical keys): combine two keys into one canonical key **only when they clearly name the SAME property under different wording** (e.g. `object_count`/`num_objects`/`count_of_items` -> `object_count`; `main_task`/`primary_task`/`task_type` -> `main_task`). Pick the clearest short snake_case name. **If two keys carry ANY discriminative distinction, KEEP THEM SEPARATE** — do NOT collapse related-but-different facets. For instance `capture_method` (real_photo / 3d_render / ai_generated), `image_source` (natural_photo / medical_scan / webpage_screenshot / textbook_figure), `rendering_style`, and `photorealism` are DISTINCT properties — keep each as its own key, never merge them into one "modality". Likewise merge VALUES only when they denote the exact same concept (`car`/`automobile` -> `car`); keep meaningfully different values distinct. When in doubt, do NOT merge.
- KEEP keys that are DISCRIMINATING and reasonably COVERED; drop near-constant keys (almost every sample shares the value) and ultra-rare keys.
- AIM for **50-60 canonical keys**, covering ALL SIX facet groups below (mirroring the per-sample extraction). Be generous — preserve the rich, orthogonal set of properties rather than collapsing into a few coarse keys:
  1. VISUAL MEDIUM & SOURCE — MANY orthogonal keys so image KINDS (real_photo / 3d_render / ai_generated / medical_scan / satellite / webpage_or_app_screenshot / scanned_document / textbook_figure / chart_or_diagram / artwork) are separable by their COMBINATION: e.g. `capture_method`, `image_source`, `is_educational_figure`, `figure_subject`, `scene_setting`, `color_mode`, `photorealism`, `has_text_overlay`, `has_annotations_or_markers`, `has_ui_chrome`, `background_type`, …
  2. SCENE CONTENT & OBJECTS — `primary_subject`, `object_categories_present`, `num_distinct_object_types`, `dominant_object`, `scene_clutter`, `background_complexity`, `spatial_layout`, `contains_people`, `contains_text`, …
  3. PERCEPTUAL OPERATION — at least one key MUST capture the core task (`main_task` / `perceptual_operation`) plus `perceptual_sub_mechanism`.
  4. TASK-SPECIFIC DIFFICULTY AXES — `count_range`, `target_type`, `target_size`, `scene_density`, `occlusion_level`, `granularity`, `text_amount`, `text_size`, `relation_type`, `reference_frame`, … These are SPARSE (each only applies to the relevant task) but they are valuable — KEEP them (do not drop just because coverage is low).
  5. QUESTION & ANSWER PROPERTIES — `answer_format`, `num_options`, `answer_type`, `requires_fine_detail`, `requires_multi_step_reasoning`, `reasoning_steps_estimate`, `requires_external_knowledge`, `answer_recoverable_from_text_alone`, …
  6. DIFFICULTY & DATA-QUALITY JUDGMENT — `estimated_difficulty`, `difficulty_drivers`, `solvable_by_coarse_glance`, `question_well_posed`, `answer_looks_correct`, `data_quality_issue`, `image_quality`, `ambiguity_level`, …
- For each canonical key give: a 1-line description, a CONTROLLED VOCAB (the canonical values after merging synonyms), and `open` = true if a free new value should be allowed when none of the vocab fits.

OUTPUT JSON only:
{
  "canonical_keys": [
    {"key": "snake_case", "description": "...", "open": true,
     "vocab": ["canonical_value_1", "canonical_value_2", "..."],
     "merged_variants": ["observed_key_a", "observed_key_b"]},
    ...
  ]
}

OBSERVED KEYS & VALUES:
{observed_stats}
\end{lstlisting}
\end{promptfigurepages}

\par\bigskip
\begin{promptfigurepages}[unbreakable]{Filling the fixed schema for one example}{fig:taxonomy-fill}
\begin{lstlisting}
TASK: Label ONE benchmark item using a FIXED attribute schema.

You see image(s), a question, and the gold answer. Fill in EVERY key of the schema with the best value for THIS item.

RULES:
- Output EXACTLY the schema's keys — all of them, nothing else.
- For each key, choose a value from its `vocab` when one fits. If the key is `open` and no vocab value fits, you may emit a concise new snake_case value. If a key genuinely does not apply to this item, use "n/a".
- Values: snake_case strings, integers, or short lists of snake_case strings — match the vocab style. No sentences, no nested objects.
- Base each value on what the IMAGE actually shows and what perceptual operation the question tests (the bottleneck), using the gold answer to disambiguate.

SCHEMA (canonical_key -> {description, open, vocab}):
{schema_json}

Return ONLY a flat JSON object {key: value, ...} containing exactly the schema keys.

QUESTION:
{question}

GOLD_ANSWER: {answer}
\end{lstlisting}
\end{promptfigurepages}

\clearpage
\begin{promptfigurepages}{Inducing a local candidate tree}{fig:taxonomy-local}
\begin{lstlisting}
TASK: Induce a **FOUR-LEVEL** ability taxonomy from a batch of {n} labeled VLM-benchmark samples.

Each sample carries fixed-schema (key, value) attributes describing its image domain, perceptual skill, difficulty, and data quality. Cluster the samples into a tree of EXACTLY depth 4:

  root
   → **L1 = DOMAIN** (image type / quality; 3–7 nodes — keep this level SHALLOW)
     → **L2 = COARSE PERCEPTION ABILITY** (VERY MACRO capability family — only a handful)
       → **L3 = FINE-GRAINED SKILL** (the specific skill inside that ability — fans out here)
         → **L4 = LEAF : difficulty / granularity / sub-type** (3–7 per rich skill; 1–2 for sparse ones)

Target **40–80 (or more) leaf nodes total**. The fan-out lives mostly at L3→L4: go deep and wide where the data is rich, stop early where it is sparse.

THE FOUR LEVELS — each splits on a DIFFERENT kind of axis:

L1 — DOMAIN TYPE (3–7, shallow). Split by image medium / source / quality, using the visual-medium attributes (`image_source`, `capture_method`, `color_mode`, `photorealism`, `is_educational_figure`, `image_quality`). Keep it to a few real domains, e.g. `natural_photographs` / `synthetic_or_rendered` / `documents_and_figures`. Use only what the data supports.

L2 — COARSE PERCEPTION ABILITY (VERY MACRO — only ~4–6 broad buckets). These are high-level capability families, NOT concrete skills. E.g. `2d_perception` / `spatial_3d_perception` / `logical_reasoning`. Roll `main_task` up into these macro families.

L3 — FINE-GRAINED SKILL (the concrete skills inside a macro ability — this is where it fans out). E.g. under `2d_perception` → `counting` / `grounding` / `geometric`. Use `main_task` / `perceptual_sub_mechanism`. Several per ability (only those present in the data).

L4 — LEAF : DIFFICULTY / GRANULARITY / SUB-TYPE (the LARGEST level — 3–7 leaves per rich skill). Split each skill along difficulty / granularity / sub-type, using the task-specific attributes (`count_range`, `target_size`, `scene_density`, `granularity`, `text_amount`, `relation_type`, …). E.g. `counting` → `small_count` / `medium_count` / `large_or_dense` / `tiny_or_occluded` / `fine_grained_subcategory` / `character_or_text_counting`. Each leaf = a 1-sentence criterion + `signature_attributes`. A SPARSE skill gets just 1–2 leaves.

GUARDRAILS:
- Total ~40–80 (or more) leaves; concentrate depth where data is rich and collapse sparse branches — a thin domain/ability/skill may stop with a single leaf rather than forcing the full 4 levels.
- Only create L1→L2→L3→L4 paths the DATA actually supports; do NOT invent empty combinations.
- MUTUALLY EXCLUSIVE & COLLECTIVELY EXHAUSTIVE: every sample maps to exactly ONE L1→L2→L3→L4 path; siblings partition their parent.
- L4 splits by DIFFICULTY / GRANULARITY / SUB-TYPE — NOT by content topic (do not split "counting cars" vs "counting fruit"; DO split "few large" vs "many tiny" vs "fine-grained subtype").
- SPLIT AXES ARE FEW: L1 = image-domain attrs; L2 = macro `main_task` grouping; L3 = `main_task` / `perceptual_sub_mechanism`; L4 = 1–2 task-specific difficulty/granularity attrs. The other ~50 schema attributes are DESCRIPTIVE — they characterize leaves via `signature_attributes`; do NOT spawn a node/leaf for every attribute.
- Ground every node name in observed attribute values; do NOT use any benchmark / dataset / framework name.

OUTPUT JSON only:
{
  "l1_nodes": [
    {"id": "snake_case", "name": "Title Case", "description": "the image domain / quality this node covers",
     "l2_nodes": [
       {"id": "snake_case", "name": "Title Case", "description": "the macro perception ability",
        "l3_nodes": [
          {"id": "snake_case", "name": "Title Case", "description": "the fine-grained skill",
           "leaves": [
             {"id": "snake_case", "name": "Title Case",
              "description": "difficulty / granularity / sub-type criterion for this leaf",
              "signature_attributes": {"<key>": ["<value>", "..."]}}
           ]}
        ]}
     ]}
  ]
}

SAMPLES (attributes only, one per line):
{batch_attributes}
\end{lstlisting}
\end{promptfigurepages}

\par\bigskip
\begin{promptfigurepages}[unbreakable]{Assigning one primary local leaf}{fig:taxonomy-assign}
\begin{lstlisting}
TASK: Assign ONE sample to EXACTLY ONE LEAF of the FOUR-LEVEL taxonomy below.

You are given: the sample's image(s), its question, its gold answer, and its extracted attributes; plus a four-level taxonomy — **L1 DOMAIN → L2 ABILITY → L3 SKILL → L4 LEAF** (each leaf carries a difficulty/granularity criterion). It is shown as nested indented lines, the deepest (most indented) line on each branch being an L4 leaf:

```
- <l1_id>: L1 domain
    - <l2_id>: L2 macro ability
        - <l3_id>: L3 fine skill
            - <leaf_id>: L4 leaf criterion
```

Decide which SINGLE L4 leaf best matches what THIS sample primarily tests:
- First pick the DOMAIN (L1) from the image medium/type, then the ABILITY (L2) and SKILL (L3) from the perceptual operation tested, then the LEAF (L4) from its difficulty / granularity / sub-type.
- Weigh the image + question + attributes together (the gold answer disambiguates). Choose the one dominant perceptual operation — if several apply, pick the bottleneck (the operation the model is most likely to fail on).

OUTPUT JSON only: {"l1_id": "...", "l2_id": "...", "l3_id": "...", "leaf_id": "...", "reason": "<= 1 sentence"}
- Report the FULL PATH (l1_id, l2_id, l3_id, leaf_id) of the chosen leaf — the same leaf name can recur under different domains/skills, so the path is what makes it unambiguous.

CRITICAL — id selection:
- Every id (l1_id, l2_id, l3_id, leaf_id) MUST be copied VERBATIM from the taxonomy above, and the four MUST form a real connected path (the leaf_id must actually sit under that l3_id, under that l2_id, under that l1_id).
- NEVER invent, abbreviate, pluralize, rename, or modify an id, and never output an id that is not in the taxonomy.
- Every sample MUST be placed into one of the EXISTING L4 leaves. If none seems perfect, choose the CLOSEST existing leaf — do not refuse and do not create a new category.

TAXONOMY (L1 → L2 → L3 → L4 leaf):
{taxonomy_compact}

SAMPLE ATTRIBUTES:
{attributes_json}

QUESTION:
{question}

GOLD_ANSWER: {answer}
\end{lstlisting}
\end{promptfigurepages}

\par\bigskip
\begin{promptfigurepages}[unbreakable]{Merging local trees and mapping their leaves}{fig:taxonomy-merge}
\begin{lstlisting}
TASK: Merge {k} local FOUR-LEVEL ability taxonomies into ONE global FOUR-LEVEL taxonomy.

Each local taxonomy was induced from a different ~1k-sample batch of the SAME dataset, so they describe the SAME ability space — **L1 DOMAIN → L2 MACRO ABILITY → L3 FINE SKILL → L4 LEAF (difficulty / granularity)** — with different wording and granularity. Consolidate them into one tree.

REQUIREMENTS:
- Output EXACTLY depth 4: root → `l1_nodes` (domain) → `l2_nodes` (macro ability) → `l3_nodes` (fine skill) → `leaves` (difficulty / granularity / sub-type).
- DE-DUPLICATE and merge synonymous nodes at EVERY level across the locals: synonymous domains (L1), synonymous macro abilities (L2), synonymous fine skills (L3), and synonymous leaves (L4).
- Match the target scale:
  - L1 = **3–7 domains** (image type / quality), shallow.
  - L2 = a few **VERY MACRO abilities** (e.g. `2d_perception` / `spatial_3d_perception` / `logical_reasoning`) — NOT concrete skills.
  - L3 = the fine skills inside each ability (e.g. counting / grounding / depth / ocr).
  - L4 = difficulty / granularity / sub-type leaves; **target ~40–80 leaves total**.
- MUTUALLY EXCLUSIVE & COLLECTIVELY EXHAUSTIVE; node names grounded in the locals; do NOT use any benchmark / dataset / framework name.

CRUCIAL — also output a CROSSWALK that maps EVERY local L4 leaf to exactly one global L4 leaf, keyed by "<batch_index>::<local_leaf_id>". Every local leaf from every local taxonomy MUST appear exactly once as a crosswalk key → a global leaf id. This is how per-sample local assignments get lifted to the global tree — completeness is essential.

OUTPUT JSON only:
{
  "global_taxonomy": {
    "l1_nodes": [
      {"id": "snake_case", "name": "Title Case", "description": "domain",
       "l2_nodes": [
         {"id": "snake_case", "name": "Title Case", "description": "macro ability",
          "l3_nodes": [
            {"id": "snake_case", "name": "Title Case", "description": "fine skill",
             "leaves": [{"id": "snake_case", "name": "Title Case", "description": "difficulty/granularity criterion"}]}
          ]}
       ]}
    ]
  },
  "crosswalk": {"<batch_index>::<local_leaf_id>": "<global_leaf_id>", "...": "..."}
}

LOCAL TAXONOMIES (batch_index -> its four-level tree):
{local_taxonomies_json}
\end{lstlisting}
\end{promptfigurepages}

\par\bigskip
\begin{promptfigurepages}[unbreakable]{Assigning reviewed capability leaves to the three Rs}{fig:taxonomy-3r}
\begin{lstlisting}
You are grouping a fixed set of visual capabilities under the three R's of computer vision.

The three R's are from Malik, Arbeláez, Carreira, Fragkiadaki, Girshick, Gkioxari, Gupta,
Hariharan, Kar and Tulsiani, "The Three R's of Computer Vision: Recognition, Reconstruction and
Reorganization" (Pattern Recognition Letters 72:4–14, 2016).

{r_families}

TASK
The {n_leaves} capability leaves below already exist and are FIXED — you may not rename, merge,
split or drop any of them. Assign each one to exactly one R, working bottom-up: read what the leaf
actually requires and decide which R that is, rather than starting from a picture of how the three
should be balanced.

HOW TO DECIDE
- Ask what the model must *recover* to answer, not what the output looks like. A multiple-choice
  answer about depth is still Reconstruction.
- The deciding question is which R's failure would make the leaf impossible:
  not knowing what things are called (Recognition), not recovering the physical scene
  (Reconstruction), or not organising the image into the right structures (Reorganization).
- Many leaves touch two R's. Choose the one that carries the *primary* burden, and name the
  runner-up in `runner_up`.
- Do not balance the three groups. A genuinely lopsided split is an acceptable answer; forcing
  leaves sideways to even out the counts is not.
- A leaf's attached I2I editing task is a strong hint about what it demands, but the leaf is
  defined by its own definition and criteria, not by the editing task's output format.

LEAVES
{leaf_catalog}

Return one entry per leaf, all {n_leaves} of them, echoing each `leaf_id` exactly. `confidence` is
`high` when the leaf clearly belongs to one R and `low` when it genuinely straddles two.
Keep `reason` under 200 characters and give no chain-of-thought.
\end{lstlisting}
\end{promptfigurepages}
\endgroup

%% file: iclr2026/sections/appendix/7_annotation_prompts.tex
\clearpage
\subsection{Three-model capability annotation}
\label{app:prompt-annotation}

We obtain independent capability labels from \texttt{gemini-3.7-flash}, \texttt{gpt-5.6-sol}, and \texttt{claude-opus-4-8}. Gemini is accessed through its API with a low thinking level and an output cap of 16,384 tokens. GPT and Claude are accessed through the Codex and Claude Code CLIs, respectively, with medium reasoning effort and no explicit output-token cap. We do not override the temperature. The three raters receive the same instructions and capability definitions, with the catalogue order shuffled deterministically for each rater, and none of them sees the others' predictions. Appendix~\ref{app:unitaskonomy-annotation} reports the annotation outcomes.

\paragraph{Inputs.}
Each request contains at most 20 examples from one benchmark. Each example provides its source image(s), question, answer choices, reference answer, UID, benchmark, and subcategory when available. The reference answer helps disambiguate the visual capability required to solve the item, and the prompt forbids classification by benchmark name. Questions are capped at 1,200 characters, and each choice and reference answer at 200 characters.

Gemini receives the source images of each item immediately after its text, in source order, and each image is downscaled to a maximum edge of 768 pixels when necessary. For the CLI raters, the images of each item are combined into one PNG labeled \texttt{ITEM k}. A single image is fitted within $768\times768$ pixels. Multiple images are arranged in two columns, with each panel fitted within $384\times384$ pixels before labels and borders are added.

\paragraph{Shared instructions and output format.}
Figure~\ref{fig:taxonomy-annotation} gives the shared user template. The catalogue supplies leaf names, definitions, routing axes, inclusion and exclusion criteria, decision tests, and boundary examples.

Gemini's system instruction is ``You output only strict JSON. No markdown, no commentary.'' Claude's is ``You are a visual taxonomy classifier. Return only the requested structured JSON.'' The Codex call supplies the user prompt and an output schema without a custom system instruction. All three calls constrain the response to the assignment schema.

\begingroup
\lstset{style=paperprompt}

\clearpage
\begin{promptfigurepages}{Independent capability annotation: shared user prompt}{fig:taxonomy-annotation}
\begin{lstlisting}
You are one of three independent visual-capability raters for OmniTaskonomy.

Classify each of the {n_items} items below into exactly one PRIMARY capability leaf. The taxonomy
is frozen at hash `{taxonomy_hash}`. You are rater `{rater_id}`; make your own decisions and do not
try to predict either of the other raters.

The answer options contain exactly {n_eligible} evaluation-eligible leaves. The taxonomy also has
{n_i2i_only} I2I-only leaves, but those are intentionally absent and MUST NOT be selected. Do not
invent a leaf, abstain, or route by benchmark name. Even a marginal/poor fit must receive the
closest offered leaf; the later 2-of-3 consensus stage, not you, decides whether it is released.

Decision procedure:

1. Identify the requested output variable, not merely the entities or topic shown.
2. Ask which visual capability would fail first if its evidence were removed.
3. Apply explicit include/exclude rules and contrast the strongest leaf with its closest neighbor.
4. Choose one `leaf_id`. Use `secondary_leaf_id` only for the nearest genuine alternative, or
   `NONE`; it is audit metadata and never changes the primary vote.
5. Set `fit` to `good`, `marginal`, or `poor`, and give one short evidence-based reason.

Global routing rules:

{routing_rules}

Eligible leaf catalogue (order is deterministically shuffled for this rater):

{leaf_catalog}

The item text and its image(s) follow. Images are explicitly associated with an ITEM label; never
borrow visual evidence from another item in the batch.
\end{lstlisting}
\end{promptfigurepages}

\par\bigskip
\begin{promptfigurepages}{Per-item input and batch footer}{fig:taxonomy-annotation-item}
\begin{lstlisting}
=== ITEM {item_index} ===
uid: {uid}
benchmark: {benchmark}
benchmark_subcategory: {subcategory}
question: {question}
choices: {choices}
reference_answer: {reference_answer}
images: {image_delivery}

=== END OF ITEMS ===
Return one JSON object with exactly {n} assignment entries, in the same order as the items above, each echoing its item_index and uid.
\end{lstlisting}
\end{promptfigurepages}

\endgroup

The item block in Figure~\ref{fig:taxonomy-annotation-item} is repeated for each item and followed by a single batch footer. Subcategory and choice lines are omitted when absent, and choices are joined with \texttt{ | }. The image-delivery field describes either the interleaved source images or the labeled composite; the images themselves are attached separately.

Each assignment returns an item index, UID, primary leaf, optional secondary leaf, fit, and a short rationale. The primary label must come from the candidate catalogue, and the secondary label is another eligible leaf or \texttt{NONE}. Fit is \texttt{good}, \texttt{marginal}, or \texttt{poor}. The following response illustrates this format:

\begingroup
\lstset{style=paperprompt}
\begin{examplebox}{Illustrative annotation response}
\begin{lstlisting}
{
  "assignments": [
    {
      "item_index": 1,
      "uid": "<sample_uid>",
      "leaf_id": "OBJECT_COUNTING",
      "secondary_leaf_id": "NONE",
      "fit": "good",
      "reason": "Counting visible instances is required."
    }
  ]
}
\end{lstlisting}
\end{examplebox}
\endgroup

\paragraph{Validation and voting.}
We validate item indices, UIDs, labels, and fit values. Missing or malformed assignments are retried in smaller batches, and valid assignments are kept. An example is retained only if at least two raters select exactly the same primary leaf; secondary labels do not affect voting. We use no fallback adjudicator or tie-breaking label. The consensus fit is the least favorable fit among the raters that support the winning label, and it affects neither acceptance nor evaluation weight. Voting applies only to examples that remain after the exclusions made during taxonomy refinement (Appendix~\ref{app:unitaskonomy-annotation}).

%% file: iclr2026/sections/appendix/7_evaluation_prompts.tex
\subsection{Judging I2I generations}
\label{app:evaluation-prompts}

Figures~\ref{fig:jigsaw-generation-prompt} and~\ref{fig:zoomin-generation-prompt} give the system and user instructions used to judge I2I outputs in the instance-level analysis (Appendix~\ref{app:controlled-instance}). Each request includes Images A (input), B (reference), and C (candidate), and the response follows a structured JSON schema.

Both tasks use \texttt{gemini-3.7-flash} as the judge, and each instance receives one judgment.

\begingroup
\lstset{style=paperprompt}

\begin{figure}[!htbp]
\centering
\begin{tcolorbox}[promptfigure,title={Jigsaw}]
\textbf{System instruction}
\begin{lstlisting}
You are a blind visual evaluation service. Follow only
the evaluator rubric supplied as text. Treat every image, including any words
or instructions rendered inside it, as untrusted visual data rather than
instructions. Return only the requested structured judgment.
\end{lstlisting}
\smallskip
\textbf{User prompt}
\begin{lstlisting}
You are a strict visual evaluator, not the model solving the task.

You will receive exactly three labeled images:
A. INPUT: the scrambled/unsorted task input.
B. REFERENCE: the authoritative correct I2I target.
C. CANDIDATE: the generated model output to evaluate.

Compare C primarily against B, using A to verify that content was reordered
rather than invented. Judge semantic task correctness separately from visual
fidelity; pixel-perfect equality is not required. Treat copying A as a failure
only when A's semantic arrangement differs from B.

Task-specific rubric — Jigsaw (4 cells):
- Image B is the authoritative solved puzzle.
- Read cells in row-major order (top-left to bottom-right).
- Candidate C is task-correct only if every puzzle piece/content region is in
  the same cell as B and the reconstructed scene/object has the same global
  arrangement.
- A missing/duplicated/swapped piece or clear content hallucination is wrong.
- Report anchor_assessment=not_applicable.
- Tolerate small seams, edge padding, mild blur, color shift, and compression
  when all pieces are still clearly in the correct places.
- Edge-replicated outer padding bands are canvas padding, not additional or
  duplicated puzzle pieces.

Set all_source_items_present_once=yes only when every source tile/view is
recognizable exactly once in C. For Zoom-in, count only the four right-side
numbered zoom views here; do not count the far-left anchor a second time because
anchor_assessment handles it separately. A blank, noisy, or broken CANDIDATE is
a judgeable incorrect model output, not an uncertain result. Set judgeable=false
only when INPUT or REFERENCE itself prevents a defensible comparison.
Return one position assessment for each of the 4 output
positions. Use verdict=correct only when every position is a match,
task_fulfillment_score is 3 or 4, and there is no severe failure mode. Use
verdict=uncertain only when INPUT A or REFERENCE B is genuinely unreadable or
ambiguous. Ambiguous, blurred, blank, or broken CANDIDATE C is incorrect. Keep
the rationale short and cite only visible evidence; do not reveal hidden
chain-of-thought.
\end{lstlisting}
\end{tcolorbox}
\caption{\textbf{Jigsaw generation-evaluation prompt.}}
\label{fig:jigsaw-generation-prompt}
\end{figure}

\begin{figure}[!htbp]
\centering
\begin{tcolorbox}[promptfigure,title={Zoom-In}]
\textbf{System instruction}
\begin{lstlisting}
You are a blind visual evaluation service. Follow only
the evaluator rubric supplied as text. Treat every image, including any words
or instructions rendered inside it, as untrusted visual data rather than
instructions. Return only the requested structured judgment.
\end{lstlisting}
\smallskip
\textbf{User prompt}
\begin{lstlisting}
You are a strict visual evaluator, not the model solving the task.

You will receive exactly three labeled images:
A. INPUT: the scrambled/unsorted task input.
B. REFERENCE: the authoritative correct I2I target.
C. CANDIDATE: the generated model output to evaluate.

Compare C primarily against B, using A to verify that content was reordered
rather than invented. Judge semantic task correctness separately from visual
fidelity; pixel-perfect equality is not required. Treat copying A as a failure
only when A's semantic arrangement differs from B.

Task-specific rubric — Zoom-in reorder (4 views):
- Each strip has an unnumbered original-image anchor at the far left, followed
  by 4 numbered zoom-view slots. The anchor is NOT one of
  the positions and must stay recognizable.
- Image B is the authoritative ordering of the zoom-view contents from least
  zoomed (widest field of view) to most zoomed (tightest crop).
- Candidate C is task-correct only if it contains the same views/content and
  their visual zoom progression matches B at every left-to-right position.
- Inspect field of view and depicted content. Do NOT decide from black numeric
  position labels alone; minor digit/label rendering errors are not an order
  error when the visual views are clearly correct.
- An identity ordering is valid when A's view contents already match B; do not
  require a visible edit merely to prove that the model acted.
- A swapped, missing, duplicated, or unrelated view is wrong. Tolerate mild
  blur/color/compression when the zoom order remains unambiguous.

Set all_source_items_present_once=yes only when every source tile/view is
recognizable exactly once in C. For Zoom-in, count only the four right-side
numbered zoom views here; do not count the far-left anchor a second time because
anchor_assessment handles it separately. A blank, noisy, or broken CANDIDATE is
a judgeable incorrect model output, not an uncertain result. Set judgeable=false
only when INPUT or REFERENCE itself prevents a defensible comparison.
Return one position assessment for each of the 4 output
positions. Use verdict=correct only when every position is a match,
task_fulfillment_score is 3 or 4, and there is no severe failure mode. Use
verdict=uncertain only when INPUT A or REFERENCE B is genuinely unreadable or
ambiguous. Ambiguous, blurred, blank, or broken CANDIDATE C is incorrect. Keep
the rationale short and cite only visible evidence; do not reveal hidden
chain-of-thought.
\end{lstlisting}
\end{tcolorbox}
\caption{\textbf{Zoom-In generation-evaluation prompt.}}
\label{fig:zoomin-generation-prompt}
\end{figure}
\endgroup

%% file: iclr2026/sections/appendix/7_i2i_data_prompts.tex
\clearpage
\subsection{Recognition editing-data routing}
\label{app:recognition-data-prompts}

Figures~\ref{fig:recognition-routing-system} and~\ref{fig:recognition-routing-user} give the system instruction and per-item input used to select Recognition supervision from ConceptEdit-12M (Appendix~\ref{app:i2i-training-data}). This text-only pass uses \texttt{gpt-5.6-luna} through the Codex CLI. The capability catalogue and routing rules are inserted into the system template, and the item block is repeated for each item in a batch. Each response contains the item index, the selected leaf or \texttt{none}, a confidence, a short rationale, and the alternative family for rejected items.

The prompt was written before the Recognition tasks were added and therefore refers to 15 I2I objectives; the paper uses the expanded 19-task inventory. Three Chinese removal aliases in the original instruction are rendered in English below, and the routing criteria are unchanged. The original prompt and catalogue inputs are included with the accompanying analysis materials.

\begingroup
\lstset{style=paperprompt,basicstyle=\fontsize{8}{9.5}\selectfont\ttfamily}
\begin{promptfigurepages}{Recognition-data routing: system instruction}{fig:recognition-routing-system}
\begin{lstlisting}
You are extending OmniTaskonomy, a taxonomy that places image-to-image (I2I) supervision
objectives and image-to-text (I2T) evaluation capabilities in one hierarchy, organised by *what
visual information they target* rather than by output modality. Its three families are Recognition
(identifies semantic content), Reconstruction (recovers geometry and photometric properties) and
Reorganization (groups, locates and relates visual elements).

The paper's 15 I2I objectives contain **no Recognition objective at all** — a property of the
supervision tasks that happened to be selected, not a claim that Recognition cannot be supervised
through images. We are mining an image-editing corpus for exactly that missing supervision.

Each item is one editing example: a source image becomes an edited image by following an
instruction. You do not see the images; you get the dataset's own edit-concept labels, the
instruction, and a caption of the source image.

TASK: decide which single **Recognition** I2T capability this editing example supervises. Apply the
taxonomy's own criterion — *the supervised output determines the family, not the dataset name or the
presence of semantic labels*. Here the supervised output is the edited image, so ask: what visual
information must be correct in that target image for the edit to count as done, and which capability
is that? Judge the primary requirement, not everything the edit touches in passing.

Answer "none" when that primary requirement is not semantic recognition, namely when it is:
  - Reconstruction: Recovering geometric and photometric scene/image properties: projected 2D spatial arrangement, depth, distance, metric size, orientation, curvature, 3D pose and shape, illumination, and geometry hidden by an occluder. The question it answers is *what is the geometric scene or its projection like*.
  - Reorganization: Organising visual elements into coherent structures without necessarily naming them: grouping and segmentation, correspondence, ordering, counting, localization, and connection/topology. The question it answers is *what goes with what and how is it organised*.
  - or when the target image requires no particular visual understanding (a global filter, a
    resolution or quality change, a re-render with no semantic target).
"none" is a normal, frequent answer. Prefer it over a weak or merely topical match.

## The 11 Recognition capabilities you may choose from
Names and definitions are the paper's; `id` is the internal identifier used in the exports.

{recognition_capability_catalogue}

## Global routing rules

{global_routing_rules}

## Corpus-specific rule — overrides your own judgement wherever it applies
**Pure object removal.** When the instruction deletes an existing *nameable object or entity* — a
person, vehicle, sign, wire, fence, piece of furniture, item of clutter — and puts nothing in its
place, so that the space it occupied has to be filled in from its surroundings (remove X / generic removal /
magic eraser / erase / clean up / generative erase), answer `leaf: "none"` with `other: "RCN"`. The
pixels this objective actually supervises are the reconstructed background behind the removed thing,
not the identity of the thing itself.

This rule does NOT apply to, and you route these by the ordinary criterion:
  - a **replacement** — something new takes the removed thing's place; route by what the new
    content is;
  - **repairing a defect or restoring a normal state** — damage, cracks, stains, rust, mould,
    wrinkles, scratches, dust, fingerprints, peeling paint, glare and other departures from an
    intact, clean or correct condition. Locating and fixing such a departure is the image-valued
    counterpart of Anomaly detection, so it stays in Recognition (Anomaly detection, or Appearance
    understanding when the requested variable is a named surface property such as gloss or finish).

## Output
For every item return: its index `i`, `leaf` (one leaf id, or "none"), `conf` in [0,1], `why`
(<= 12 words), and when leaf is "none" also `other` = which family the item really belongs to
(RCN, RORG, NO_VISUAL_CAPABILITY). Return exactly one result per item, in the order given.
\end{lstlisting}
\end{promptfigurepages}

\par\bigskip
\begin{promptfigurepages}[unbreakable]{Recognition-data routing: per-item input}{fig:recognition-routing-user}
\begin{lstlisting}
#{item_index}
edit concept: {sub_category} / {task} / {concept}
instruction: {short_english_instruction}
detailed: {detailed_english_instruction}
source image: {source_caption}

[Repeat for each item in the batch.]
\end{lstlisting}
\end{promptfigurepages}

\par\bigskip
\begin{promptfigurepages}[unbreakable]{Example Recognition training instructions}{fig:recognition-training-instructions}
\begin{lstlisting}
Object editing:
Add a decorative rug to the floor in front of the painting.

Attribute editing:
Change the color of the scooter.
\end{lstlisting}
\end{promptfigurepages}
\endgroup

The two instructions in Figure~\ref{fig:recognition-training-instructions} are taken from training records. Each is paired with its original source image and supervised by the corresponding edited image. The routing prompts serve only for data selection and are not used as BAGEL training prompts.